\documentclass[11pt]{article}

\usepackage[preprint]{acl}

\usepackage{times}
\usepackage{latexsym}

\usepackage[T1]{fontenc}

\usepackage[utf8]{inputenc}

\usepackage{microtype}

\usepackage{inconsolata}

\usepackage{graphicx}
\usepackage{amsmath}
\usepackage{amssymb}
\usepackage{booktabs}
\usepackage{subcaption}
\usepackage{colortbl}
\usepackage{comment}
\usepackage{algorithm}
\usepackage[noend]{algpseudocode}
\usepackage{multirow}
\usepackage{arydshln}

\title{Beyond the Covariance Trap: Unlocking Generalization in Same-Subject Knowledge Editing for Large Language Models}

\author{%
  Xiyu Liu\textsuperscript{\rm 1,2}, Qingyi Si\textsuperscript{\rm 3}, Zhengxiao Liu\textsuperscript{\rm 1,2}\thanks{Zhengxiao Liu is the corresponding author.}, Chenxu Yang\textsuperscript{\rm 1,2}, Naibin Gu\textsuperscript{\rm 1,2}, Zheng Lin\textsuperscript{\rm 1,2}\\
  \textsuperscript{\rm 1}Institute of Information Engineering, Chinese Academy of Sciences, Beijing, China\\
  \textsuperscript{\rm 2}School of Cyber Security, University of Chinese Academy of Sciences, Beijing, China\\
    \textsuperscript{\rm 3}JD.com, Beijing, China\\
  \{liuxiyu, liuzhengxiao, gunaibin, linzheng\}@iie.ac.cn, siqingyi.phoebus@jd.com \\
}

\begin{document}
\definecolor{darkred}{RGB}{220,0,0}

\maketitle

\begin{abstract}
While locate-then-edit knowledge editing efficiently updates knowledge encoded within Large Language Models (LLMs), a critical generalization failure mode emerges in the practical same-subject knowledge editing scenario: models fail to recall the updated knowledge when following user instructions, despite successfully recalling it in the original edited form.
This paper identifies the geometric root of this generalization collapse as a fundamental conflict where the inner activation drifts induced by prompt variations exceed the model's geometric tolerance for generalization after editing. We attribute this instability to a dual pathology: (1) The joint optimization with orthogonal gradients collapses solutions into sharp minima with narrow stability, and (2) the standard covariance constraint paradoxically acts as a \textbf{Covariance Trap} that amplifies input perturbations. To resolve this, we introduce RoSE (Robust Same-subject Editing), which employs \textit{Isotropic Geometric Alignment} to minimize representational deviation and \textit{Hierarchical Knowledge Integration} to smooth the optimization landscape. Extensive experiments demonstrate that RoSE significantly improves instruction-following capabilities, laying the foundation for robust interactive parametric memory of LLM agents.
\end{abstract}

\section{Introduction}
Large Language Models (LLMs)~\cite{minaee2024largelanguagemodelssurvey, zhao2024surveylargelanguagemodels} have become foundational in numerous applications, yet their knowledge is inherently static~\cite{heinzerling-inui-2021-language, wang-etal-2021-generative, roberts-etal-2020-much}, becoming outdated or incorrect as the world changes. Knowledge Editing (KE) has emerged as a crucial field to address this limitation, offering a direct and computationally efficient way to inject or amend factual knowledge within the parameters of LLMs without the need for costly full-scale retraining~\cite{yao-etal-2023-editing, wang2024knowledgeeditinglargelanguage, mitchell2022fast}. This process is essential for maintaining the reliability, accuracy, and safety of LLM agents over time, by enabling deliberate and auditable revisions of their parametric memory~\cite{hu2025memoryageaiagents}.

\begin{figure}[t]
    \centering
    \includegraphics[width=1.0\linewidth]{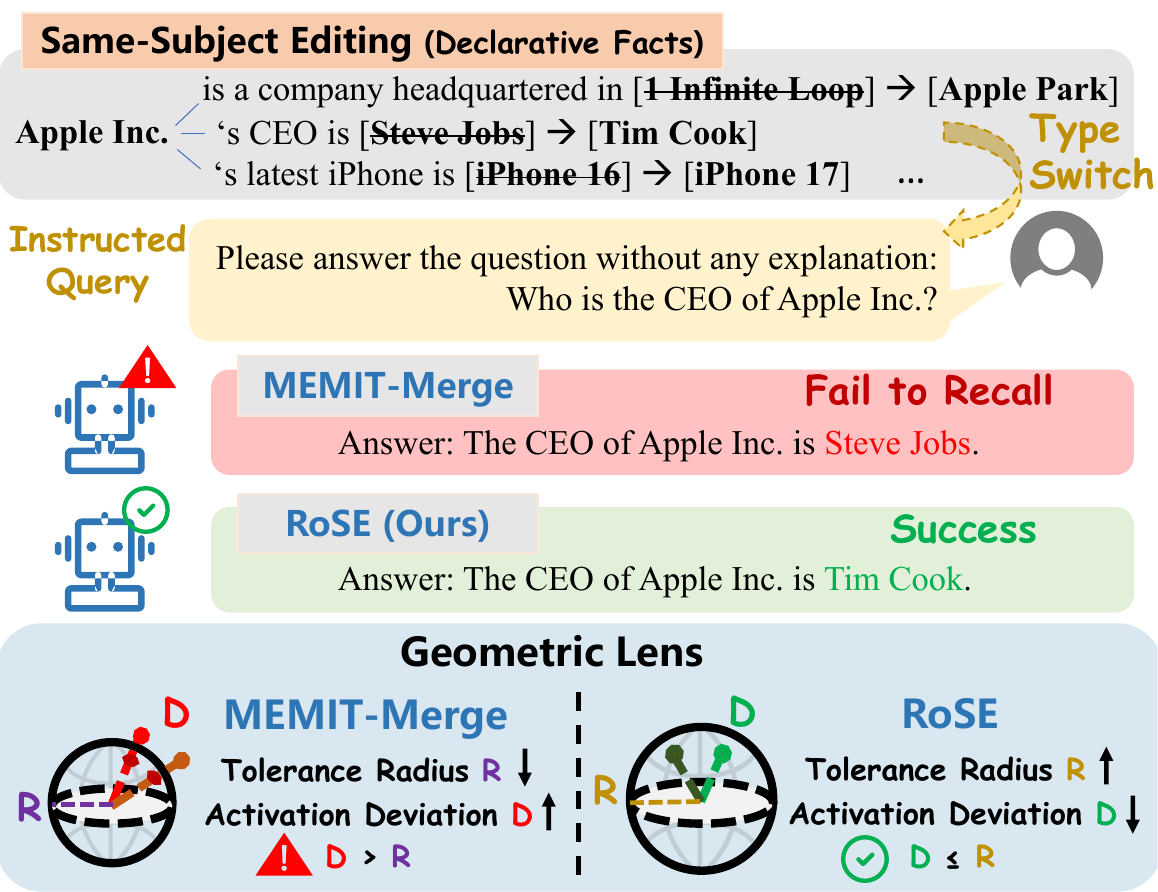}
    \caption{Our work reveals that the current same-subject knowledge editing method MEMIT-Merge fails to generalize to instructed queries because Activation Deviation exceeds the edited model's Tolerance Radius ($D > R$) in the activation space. We unlock robustness by reshaping the geometry towards the safe condition $D \le R$.}
    \label{fig:motivation}
\end{figure}

A dominant paradigm for KE is the locate-then-edit approach~\cite{zhang2024comprehensivestudyknowledgeediting,meng2022locating,Meng2022MassEditingMI,fang2024alphaeditnullspaceconstrainedknowledge, gupta2025efficientknowledgeeditingminimal}, which has demonstrated remarkable success in editing isolated facts by identifying and updating specific key-value associations within MLP sublayers of the model~\cite{geva-etal-2022-transformer, geva-etal-2023-dissecting}. However, the complexity of real-world information demands more than isolated facts, as knowledge is inherently interconnected. This paradigm faces a significant challenge in a common and critical use case: editing multiple, distinct relations for the same subject~\cite{duan2025relatedknowledgeperturbationmatters} (e.g., updating both the \textit{CEO} and the \textit{headquarter} of \textit{Apple Inc.}). 

While the recent work~\cite{dong2025memitmergeaddressingmemitskeyvalue} effectively resolves the conflict of editing multiple facts for the same subject via joint optimization, we identify a critical, previously overlooked failure mode: \textbf{generalization collapse}. We observe that although models can perfectly recall edited facts when prompted with the exact linguistic form used during editing (i.e., declarative sentences), they fail catastrophically when queried with unseen forms like user instructions or questions (Fig.~\ref{fig:motivation}). Since real-world interactions with LLMs are overwhelmingly instruction-driven, this generalization gap between declarative fact storage and instructed recall represents a major barrier to deploying knowledge editing systems in the wild.

What drives this generalization failure? We diagnose the problem of locate-then-edit same-subject editing through a geometric lens, identifying a fundamental conflict where the prompt-induced \textbf{Activation Deviation ($D$)} exceeds the model's \textbf{Tolerance Radius ($R$)} for generalization (formally defined in Section~\ref{sec:geo_formulation}).
First, we find that the joint optimization for same-subject facts forces the model to satisfy orthogonal gradients. This collapses the solution space into a sharp minimum, drastically shrinking the Tolerance Radius ($R \downarrow$).
Second, we uncover a \textbf{Covariance Trap}. The standard covariance constraint ($C$), intended for locality, acts as an anisotropic distortion lens. It amplifies subtle vector differences between prompt formats, causing the Activation Deviation to surge ($D \uparrow$). More importantly, our mathematical and empirical analysis reveals that the orthogonality property of different subjects inherent in the LLM activation space makes this covariance constraint redundant.
Ultimately, generalization fails because the amplified deviation overshoots the diminished tolerance radius ($D > R$), rendering the model incapable of following instructions.

To resolve this geometric deadlock, we propose \textbf{RoSE} (\textbf{Ro}bust \textbf{S}ame-subject \textbf{E}diting), a principled framework designed to ensure $D \le R$ via a two-pronged strategy.
To minimize deviation ($D \downarrow$), we introduce \textit{Isotropic Geometric Alignment} (IGA). By exploiting our finding that subject keys are naturally orthogonal, we replace the distortion-inducing covariance constraint with an isotropic identity constraint. This rectification ensures that the update direction aligns with the intrinsic subspace of the subject, making the edit robust to prompt variations.
To maximize tolerance ($R \uparrow$), we propose \textit{Hierarchical Knowledge Integration} (HKI). Instead of point-wise optimization, HKI aggregates gradients from diverse linguistic formats in a tree-structured manner. This flattens the sharp minima, effectively expanding the solution space's radius to accommodate potential shifts in activation.

Our main contributions are as follows:
\begin{itemize}
\item We discover the instruction-following failure in locate-then-edit same-subject editing and provide an in-depth geometric explanation as the root cause of this failure.
\item We identify the Covariance Trap, proving that the standard covariance constraint amplifies input noise for edited subjects, and reveal the sharp minima phenomenon caused by orthogonal value gradients.
\item We propose RoSE, which restores the safe generalization condition via a dual-pronged strategy of Isotropic Geometric Alignment and Hierarchical Knowledge Integration.
\item Extensive experiments on diverse benchmarks demonstrate that RoSE achieves the optimal performance, effectively unlocking robust instruction-following capabilities.
\end{itemize}

\section{Related Works}
Our work primarily engages with the mainstream \emph{locate-then-edit} knowledge editing paradigm and addresses the emergent challenges in same-subject fact updates.

\paragraph{The Locate-then-Edit Paradigm and The Covariance Constraint.}
Pioneered by approaches like ROME~\cite{meng2022locating}, the locate-then-edit paradigm modifies specific MLP sublayers identified via causal mediation analysis. This method treats the sublayer weight as a linear associative memory, solving a least-squares problem to map the \textit{key} of a subject ($k$) to a new target \textit{value} ($v$). To prevent catastrophic forgetting of unrelated knowledge, these methods typically introduce a regularization term weighted by a pre-computed covariance matrix ($C$)  of \textit{key} activations. MEMIT~\cite{Meng2022MassEditingMI} scales this to batch editing by distributing updates across multiple layers. While effective for isolated facts, most subsequent works~\cite{ma2024perturbationrestrainedsequentialmodelediting, cai2024oeditorthogonalsubspaceediting, fang2024alphaeditnullspaceconstrainedknowledge, li-chu-2025-adaedit} continue to rely on this covariance-based constraint, among other variants, to safeguard locality. In contrast, \textit{our work re-examines this foundational assumption, identifying the covariance matrix not as a safeguard, but as a primary source of geometric distortion that hampers generalization for same-subject knowledge editing}.

\paragraph{The Challenge of Same-Subject Multi-Relation Editing.}
A significant challenge arises in the practical scenario of editing multiple relations for the \textit{same subject}, where applications of the above methods lead to severe conflicts. The growing importance of this problem is highlighted by the introduction of dedicated benchmarks such as S2RKE~\cite{duan2025relatedknowledgeperturbationmatters}. To address this, the state-of-the-art same-subject editing work like MEMIT-Merge~\cite{dong2025memitmergeaddressingmemitskeyvalue} formulates the task as a joint optimization problem, solving for a single update matrix that satisfies all new factual constraints simultaneously.
\textit{Our work reveals that such joint optimization collapses the solution space into a sharp minimum, rendering the edited knowledge brittle to prompt variations}.

\section{Analysis of Generalization Collapse}
\label{sec:analysis}

While the recent work successfully prevents catastrophic conflicts when editing same-subject facts, the failure to generalize to instruction-following scenarios points to a deep mechanistic flaw. In this section, we investigate this phenomenon through a geometric lens. We propose that this generalization collapse is not merely a stochastic error but a deterministic consequence of a geometric pathology: the \textbf{Activation Deviation ($D$)} induced by prompt variations exceeds the model's \textbf{Tolerance Radius ($R$)}, formally denoted as the boundary violation $D > R$. We rigorously unveil this pathology from the dual perspectives of value optimization and key projection, both supported by empirical evidence on 200 samples from S2RKE~\cite{duan2025relatedknowledgeperturbationmatters} using Qwen2.5-7B-Instruct~\cite{qwen2025qwen25technicalreport}. Empirical evidence from Llama-3.1-8B-Instruct~\cite{grattafiori2024llama3herdmodels} shows the similar phenomenon, as detailed in Appendix~\ref{app:llama_pilot}.

\subsection{Preliminaries}

Our analysis builds upon the locate-then-edit editing paradigm, specifically focusing on the challenges of same-subject editing~\cite{duan2025relatedknowledgeperturbationmatters}.

\paragraph{Locate-then-Edit Paradigm.}
The locate-then-edit approach (e.g., MEMIT~\cite{Meng2022MassEditingMI}) posits that a sequence of mid-early MLP layers acts as key-value stores mediating factual recall. It treats the MLP output weights $W^l_{\text{out}}$ as linear associative memories. For a specific request $(s, r, o^*)$, the \emph{key} ($k$) is the input vector at the last subject token, and the output \emph{value} vector ($v^*$) is optimized to maximize the probability of the target object $o^*$:
\begin{equation}
 v^* = \arg\min_v \left( -\log P_v[o^* \mid (s, r)] \right).
\end{equation}
For a batch of edits with keys $K \in \mathbb{R}^{d_k \times B}$ and optimized values $V \in \mathbb{R}^{d_v \times B}$, the weights are updated via a closed-form solution derived from the least-squares objective with a constraint to preserve existing knowledge:
\begin{equation}
\label{eqn:update_rule}
W^l_{\text{out}} = W^l_0 + (V - W^l_0 K) K^\top (C + K K^\top)^{-1},
\end{equation}
where $W^l_0$ represents the original weights and $C = \mathbb{E}[kk^\top]$ is the pre-computed covariance matrix of keys, intended to approximate the distribution of unedited knowledge.

\paragraph{Same-Subject Conflicts.}
Standard editors assume keys are distinct. However, editing multiple relations for the same subject results in key collisions (identical $s$ yields identical $k$ but different $v$ for different relations). MEMIT-Merge~\cite{dong2025memitmergeaddressingmemitskeyvalue} resolves this by grouping edits by subject $S = \{(s, r_j, o^*_j)\}_j$ and jointly optimizing a single unified value vector $v^*$ that satisfies all relations simultaneously:
\begin{equation}
    v^* = \arg\min_v \sum_{(s, r_j, o_j) \in S} \left( -\log P_v[o^*_j \mid (s, r_j)] \right).
\label{eq:joint_v}
\end{equation}

\subsection{Geometric Formulations: The \texorpdfstring{$D > R$}{D > R} Pathology}
\label{sec:geo_formulation}

To mathematically formalize the generalization collapse, we define two critical metrics characterizing the geometry of the model's activation space.

\paragraph{Definition 1 (Tolerance Radius $R$).}
The Tolerance Radius quantifies the robustness of the solution basin in the \textit{value} space. It represents the maximum magnitude of Gaussian noise $\|\xi\|_2$ that can be added to the optimal \textit{value} vector $v^*$ while maintaining the prediction accuracy for the target object above a threshold $\tau$ (e.g., 0.9):
\begin{equation}
 R = \max_{\rho} \left\{ \rho \mid \mathbb{E}_{\|\xi\|=\rho} \left[ \mathbb{I}(\text{M}(v^* + \xi) = o^*) \right] > \tau \right\}
\end{equation}
Here, $\text{M}(v^* + \xi)$ denotes the model's prediction when the \textit{value} vector is set to $v^*+\xi$, given input prompts of the original edited form.
A larger $R$ indicates a wide, flat optimum, while a small $R$ indicates a sharp, fragile minimum. 

\paragraph{Definition 2 (Activation Deviation $D$).}
The Activation Deviation measures the shift in the output activation caused by prompt variations. Let $k_{o}$ be the \textit{key} vector from the original declarative prompt used during editing, and $\tilde{k}$ be the \textit{key} from another linguistic form (e.g., $k_{q}$ for the natural question form and $k_{inst}$ for the instructed query form). The update matrix $\Delta W = (V - W^l_0 K) K^\top (C + K K^\top)^{-1}$ transforms the input difference $\delta = \tilde{k} - k_{o}$ into an output deviation:
\begin{equation}
 D = \| \Delta W \tilde{k} - \Delta W k_{o} \|_2 = \| \Delta W \delta \|_2.
 \label{eq:act_dev}
\end{equation}

\paragraph{Hypothesis.}
Generalization succeeds only if the deviation stays within the tolerance bounds ($D \le R$). We hypothesize that same-subject editing fails because it simultaneously shrinks $R$ and amplifies $D$, leading to the condition $D > R$.

\begin{figure}[t]
   \centering
   \subcaptionbox{Tolerance Radius $R$.\label{fig:tol_radius}}[.5\linewidth][c]{%
      \includegraphics[width=1.\linewidth]{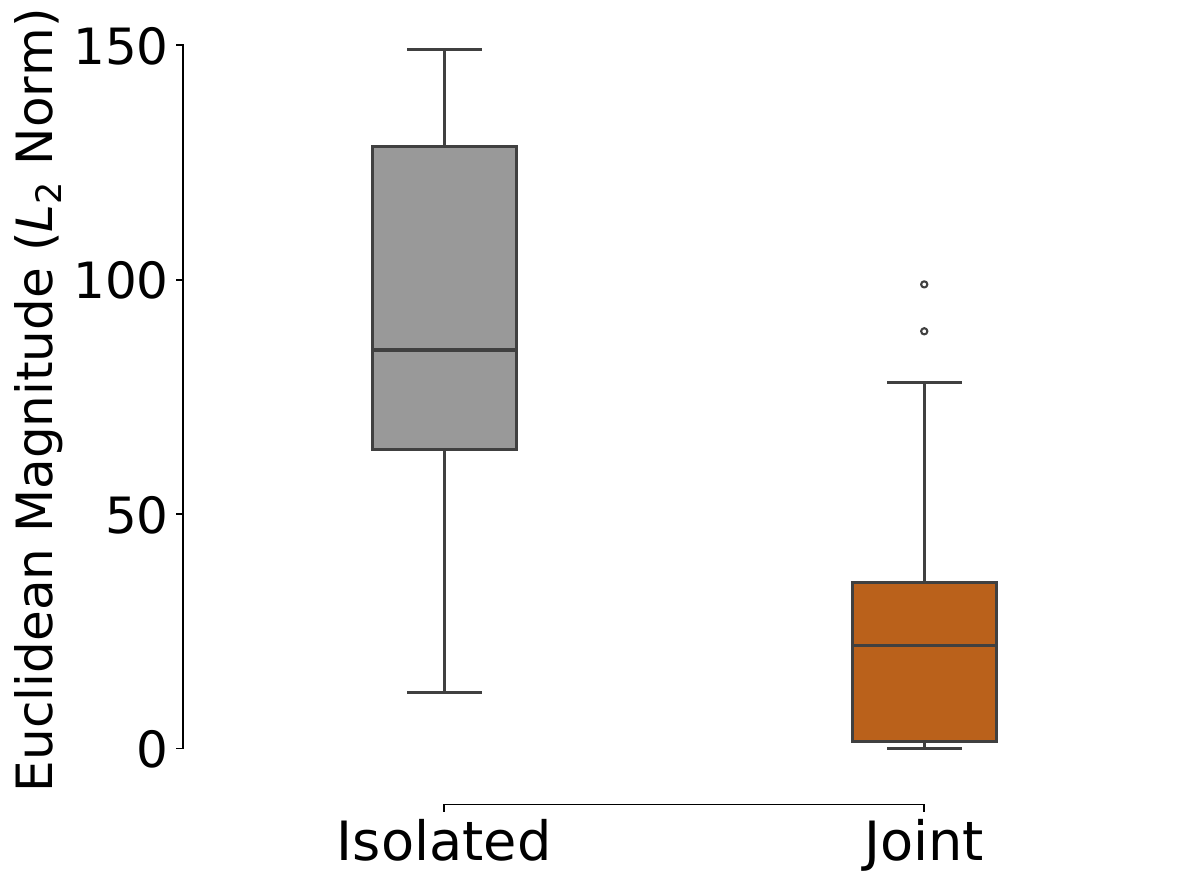}
   }
   \hspace{-0.3cm}
   \subcaptionbox{Activation Deviation $D$.\label{fig:act_dev}}[.5\linewidth][c]{%
      \includegraphics[width=1.\linewidth]{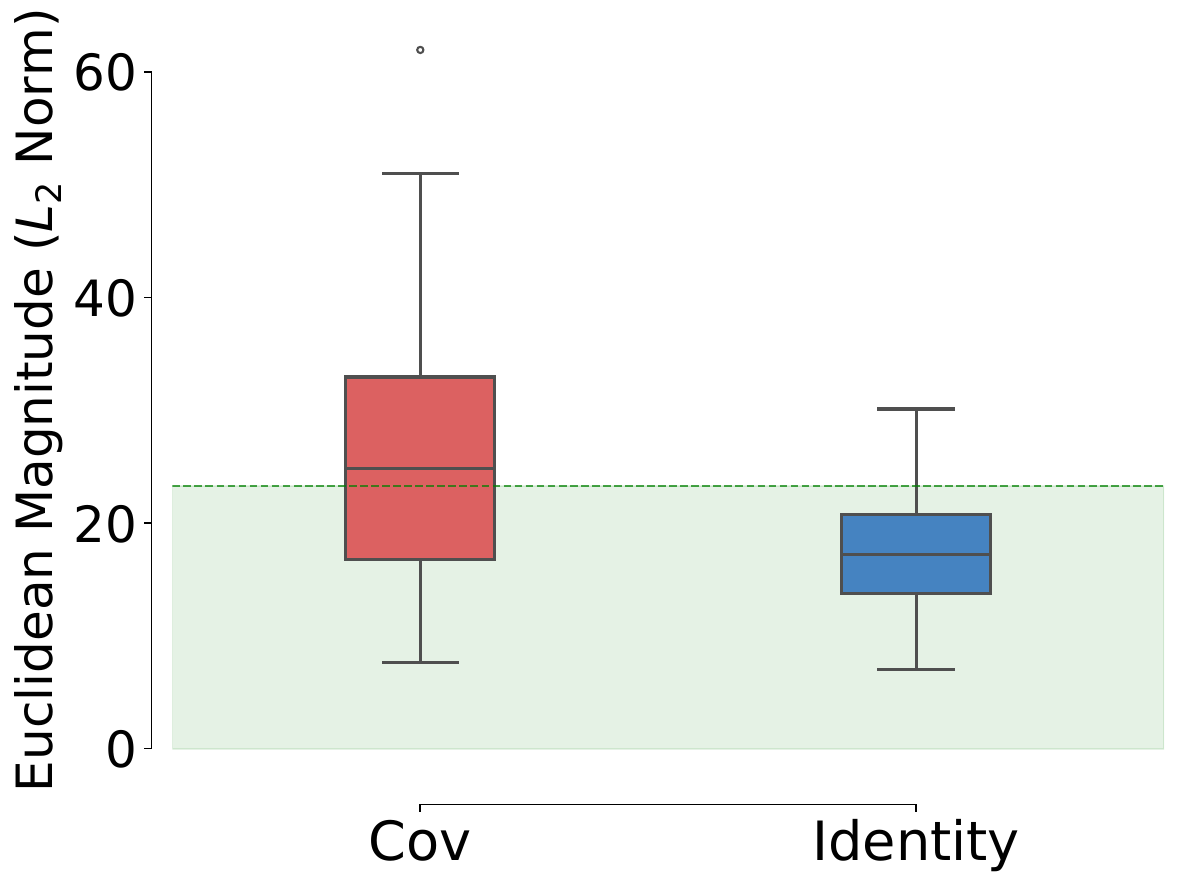}
   }
   \caption{Geometric Pathology: $D > R$. (a) Orthogonal gradients in joint same-subject editing cause tolerance radius $R$ to collapse. (b) The covariance matrix $C$ serves as an amplification trap, leading to a deviation $D$ of approximately 26.1 beyond $R$. The area below the green line (average $R$) is the safe zone. Replacing $C$ with identity matrix can suppress $D$ to around $17.4$.}
\end{figure}

\subsection{Value Perspective: Collapse of Tolerance Radius (\texorpdfstring{$R \downarrow$)}{R downarrow} }
\label{sec:value_analysis}

We first investigate the stability of the edited region ($R$) from the \textit{value} perspective (Eq.~\eqref{eq:joint_v}).

\paragraph{Empirical Observation: Radius Collapse.}
We empirically measure $R$ for both isolated fact edits (standard MEMIT) and joint same-subject multi-relation edits (MEMIT-Merge). The results reveal a catastrophic collapse. As shown in Fig.~\ref{fig:tol_radius}, while isolated edits enjoy a broad solution basin with average $R_{isolated} \approx 92.9$, the Tolerance Radius for joint same-subject editing shrinks drastically to average $R_{joint} \approx 23.3$. This suggests that the model has converged to a sharp minimum for same-subject multi-relation editing.

\begin{figure}[t]
    \centering
    \includegraphics[width=0.8\columnwidth]{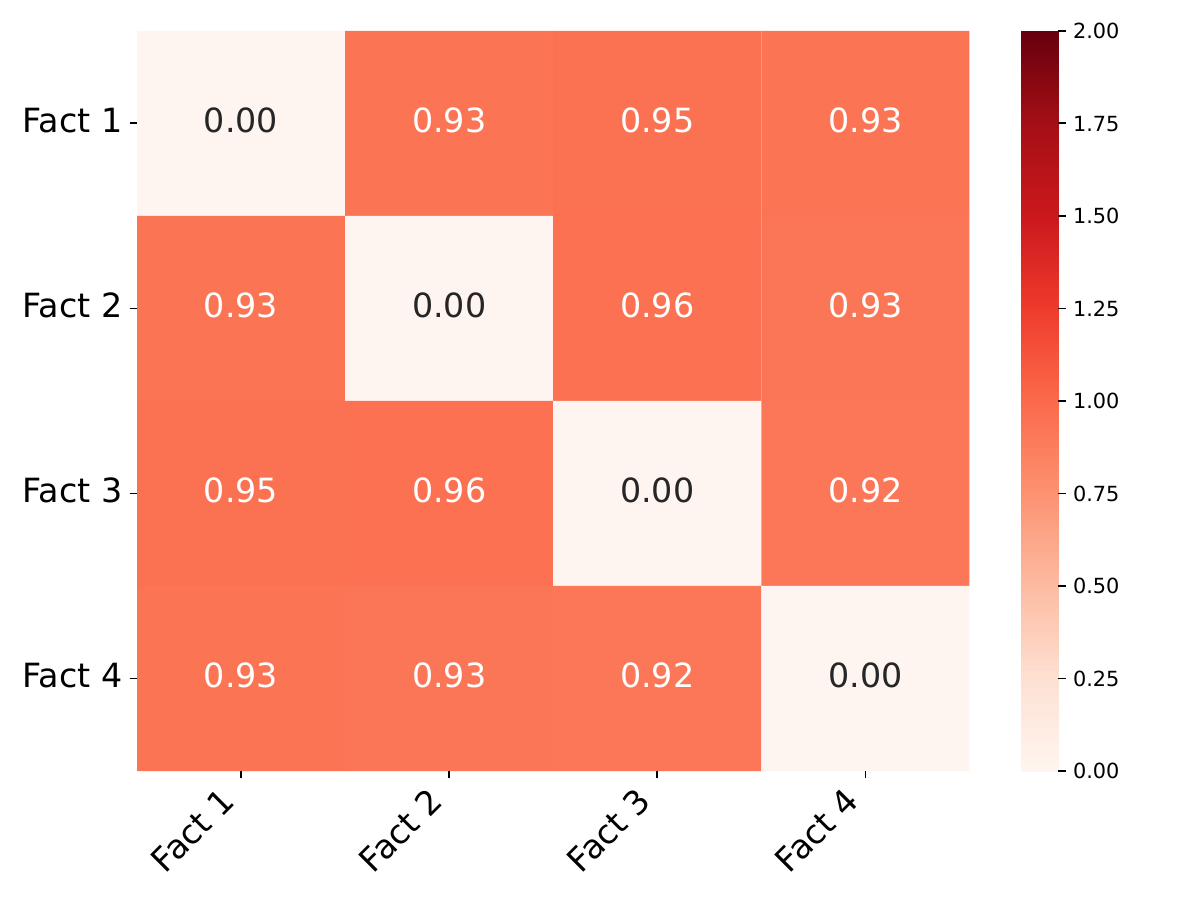}
    \caption{Distribution of gradient conflict scores for pairs of edits concerning the same subject but different relations. The conflict scores of near 1 demonstrate that their update gradients are near-orthogonal.}
    \label{fig:gradient_conflict}
\end{figure}

\paragraph{Mechanism: Gradient Orthogonality.}
Why does the solution space shrink? We attribute this to the geometric conflict between relations during optimization. We compute the pairwise gradient conflict scores of the update gradients for different relations (e.g., $g_{r_1}$ and $g_{r_2}$) of the same subject.
\begin{equation}
\label{eq:grad_conflict}
\textbf{Conflict\_Score}(g_{r_1}, g_{r_2}) = 1 - \frac{g_{r_1} \cdot g_{r_2}}{\|g_{r_1}\| \|g_{r_2}\|}
\end{equation}
As shown in Fig.~\ref{fig:gradient_conflict}, the gradient vectors are nearly orthogonal (average gradient conflict scores from $0.92$ to $0.96$).
Geometrically, optimization for each relation defines a solution subspace $S_i$. The joint optimization seeks the intersection of these subspaces: $\mathcal{S}_{joint} = \cap_i \mathcal{S}_i$. When the normal vectors (gradients) of these subspaces are orthogonal, the volume of their intersection is extremely constrained. Unlike isolated fact editing which requires satisfying only one direction for the subject, multi-fact editing for the subject forces the parameter update to walk a tightrope between multiple orthogonal constraints, resulting in a fragile solution with a collapsed Tolerance Radius ($R \downarrow$).

\subsection{Key Perspective: The Covariance Trap (\texorpdfstring{$D \uparrow$)}{D uparrow}}
\label{sec:key_analysis}

With $R$ critically diminished (from $\approx 92.9$ to $\approx 23.3$), we examine why the Activation Deviation $D$ becomes large enough to breach this boundary.

\paragraph{The Breach: $D > R$.}
We measure the actual deviation $D$ for queries of other linguistic forms using the standard update rule for locate-then-edit editing (Eq.~\eqref{eqn:update_rule}). The average result is $D_{cov} \approx 26.1$. Comparing this to the radius $R_{joint} \approx 23.3$, we observe a clear violation: $\mathbf{D_{cov} > R_{joint}}$ (Fig.~\ref{fig:act_dev}). The comparative analysis of $D$ and $R$ is mathematically sound because both are quantified through L2 normalization in the same value vector space, which guarantees metric consistency. This result confirms that the generalization failure is a deterministic geometric boundary violation.

\begin{figure}[t]
    \centering
    \includegraphics[width=1.0\columnwidth]{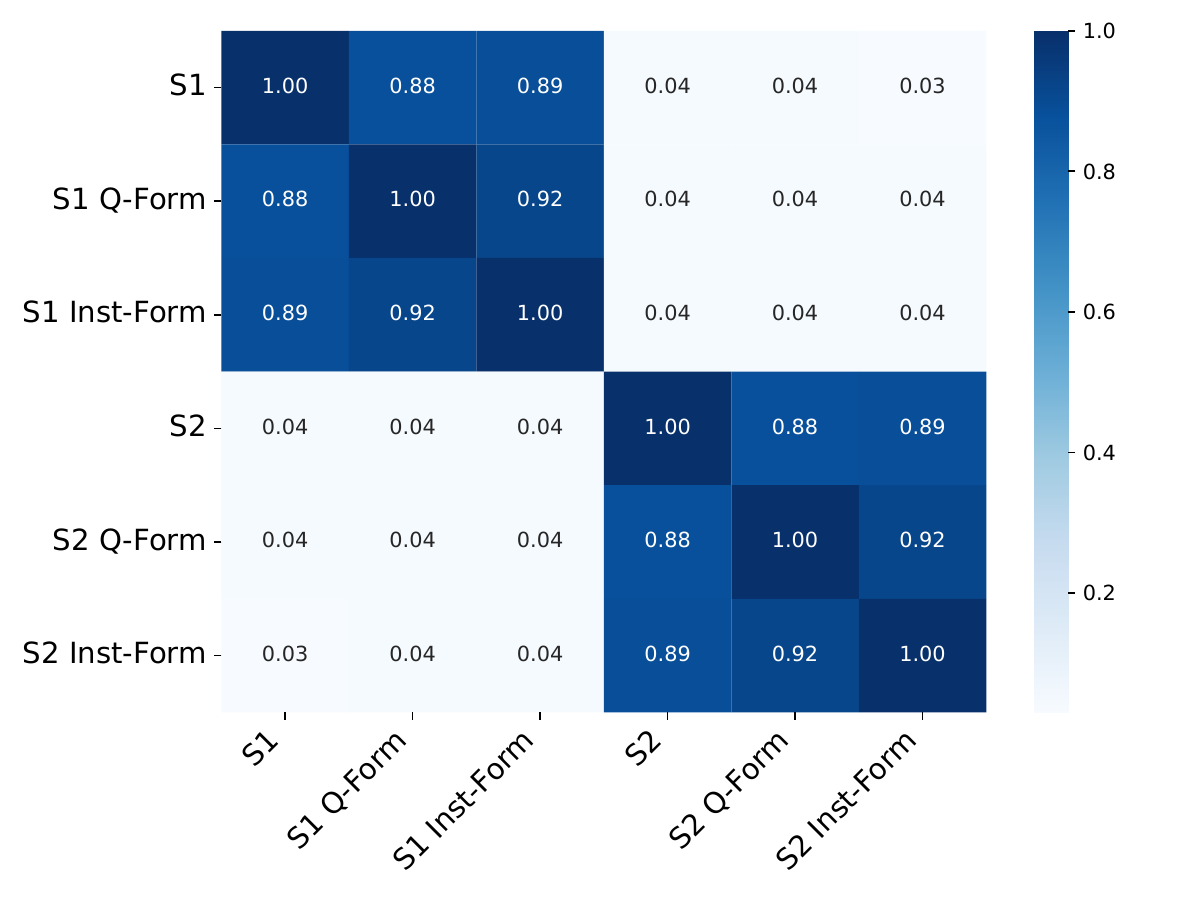}
    \caption{Average pairwise cosine similarity of $k$. The high similarity in the diagonal blocks (S1 vs. S1 Q-Form, S1 vs. S1 Instruction-Form) shows that subject representation is stable across prompt formats. The near-zero similarity in the off-diagonal blocks (S1 vs. S2) reveals that keys of distinct subjects are orthogonal.}
    \label{fig:key_heatmap}
\end{figure}

\paragraph{Mechanism: The Covariance Trap.}
We first investigate the origin of the massive activation deviation ($D$). Our geometric probing rules out retrieval failure: the average pairwise cosine similarity between \textit{key} vectors of varying forms of the same fact (declarative form: \textit{S1}, natural question form: \textit{S1 Q-Form}, instructional form: \textit{S1 Inst-Form}) is remarkably high (from 0.88 to 0.92), implying the intrinsic input perturbation $\delta = \tilde{k} - k_{o}$ is minimal. Consequently, the explosion of $D$ must be an artifact of the projection metric in the update rule (Eq.~\eqref{eqn:update_rule}).
Specifically, the standard update involves the inverse covariance matrix $C^{-1}$, which acts as an anisotropic whitening filter. Since $C$ captures enormous corpus statistics, its eigenspectrum is highly skewed. Its inverse matrix $C^{-1}$ therefore possesses an extremely large eigenvalue $\lambda_{max}$ in the direction of low data variance. We formally show that $C$ creates a \textbf{Covariance Trap} that amplifies the small prompt perturbation $\delta$ (details in App.~\ref{app:cov_trap_analysis}):
\begin{align}
    D_{cov} &\approx \| C^{-1} \delta\|_2 \\ 
    &\geq \lambda_{max}(C^{-1}) \| \delta_{proj} \|_2 > \|\delta\|_2
\end{align}
where $\delta_{proj}$ is the projection of $\delta$ onto the principal component of $C^{-1}$. This amplification explains why $D_{cov}$ is large ($\approx 26.1$) while $\delta$ is small. Empirical validation (Fig.~\ref{fig:act_dev}) confirms that replacing the anisotropic $C$ with an isotropic identity matrix $I$ can effectively close the trap, suppressing the deviation to $D_{identity} \approx 17.4$.

\paragraph{Redundancy of Covariance.}
The removal of $C$ raises a concern about locality: does using $I$ cause catastrophic forgetting of unrelated knowledge? \textbf{Surprisingly, our analysis reveals that $C$ is empirically redundant due to the inherent orthogonality of the \textit{key} space}. We find that key vectors for distinct subjects are nearly orthogonal ($k_{old} \perp k_{new}$) in Fig.~\ref{fig:key_heatmap} (e.g., cosine similarity of 0.04 for S1 vs. S2). Thus, \textbf{$K^\top k_{old} \approx \mathbf{0}$ essentially holds naturally}, where $K$ is the batched keys for the edited subject (Eq.~\eqref{eqn:update_rule}). 
Mathematically, the locality constraint for an unrelated key $k_{old}$ requires the update to have no effect, i.e., $\Delta W k_{old} \approx \mathbf{0}$. While traditional methods rely on $C^{-1}$ to orthogonalize the keys ($K^\top C^{-1} k_{old} \approx \mathbf{0}$), the intrinsic geometry already satisfies this condition under an Identity update according to the Sherman-Morrison-Woodbury formula:
\begin{align}
\Delta &W_{I} k_{old} = R K^\top ( I + K K^\top)^{-1} k_{\text{old}} \nonumber \\
&= R K^\top \left( I - K( I + K^\top K)^{-1}K^\top \right) k_{\text{old}} \nonumber\\
&=  R \left( K^\top k_{\text{old}} - K^\top K(I + K^\top K)^{-1}K^\top k_{\text{old}} \right) \nonumber\\
&\approx  R \left( \mathbf{0} - K^\top K( I + K^\top K)^{-1}\mathbf{0} \right) \nonumber\\
&= \mathbf{0},
\label{eq:cov_trap}
\end{align}
where we define $R = V - W^l_0 K$ for simplicity of representation. This proves that the computationally expensive and numerically unstable covariance matrix is unnecessary. The model naturally partitions knowledge into orthogonal subspaces, allowing us to safely adopt isotropic geometric alignment without compromising locality.

\begin{figure*}[t]
    \centering
    \includegraphics[width=2.0\columnwidth]{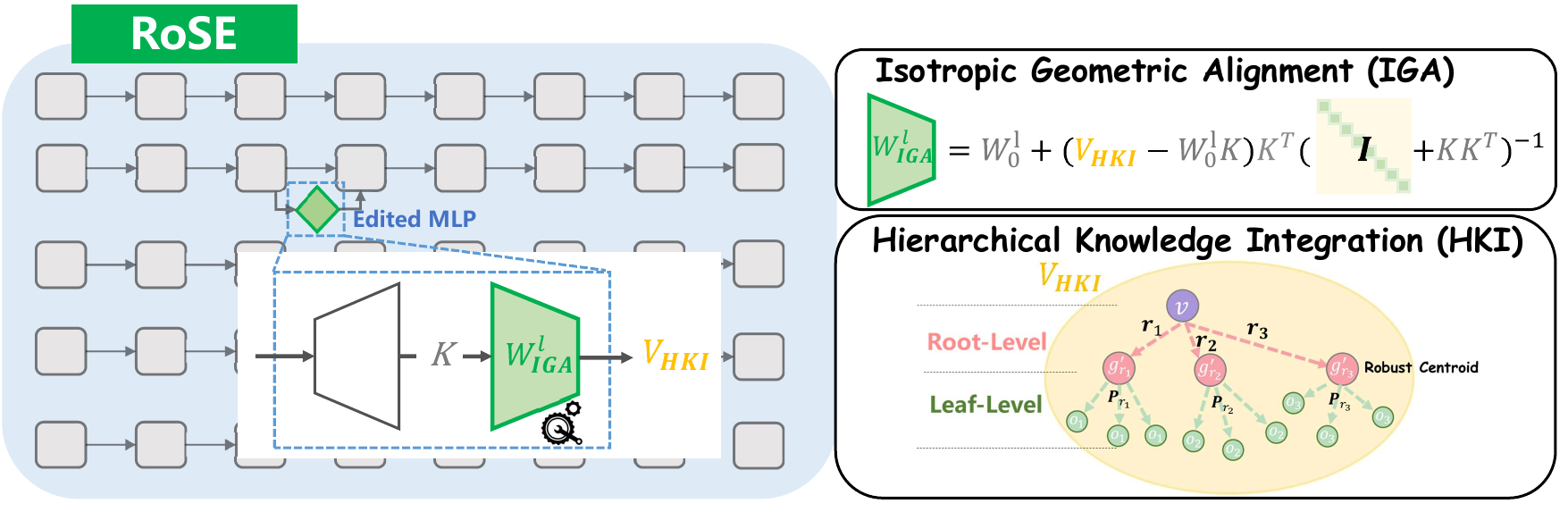}
    \caption{The RoSE framework. We expand the tolerance radius $R$ via Hierarchical Knowledge Integration (HKI) and shrink the activation deviation $D$ via Isotropic Geometric Alignment (IGA) towards the ideal condition $D \le R$.}
    \label{fig:rose}
\end{figure*}

\section{Mitigation Strategy: RoSE}
\label{sec:methodology}

Our analysis in Sec.~\ref{sec:analysis} establishes that the failure of same-subject instruction following is a deterministic consequence of the geometric condition $D > R$. To unlock robustness, we introduce \textbf{RoSE} (\textbf{Ro}bust \textbf{S}ame-subject \textbf{E}diting), a principled framework designed to reverse this inequality. RoSE achieves the target state $D \le R$ through a dual-strategy approach: (1) \textbf{Isotropic Geometric Alignment (IGA)}, which reduces the activation deviation $D$ by removing the covariance trap; and (2) \textbf{Hierarchical Knowledge Integration (HKI)}, which expands the tolerance radius $R$ via tree-like gradient aggregation.

\subsection{Isotropic Geometric Alignment}
\label{sec:alignment}

Standard editing methods rely on the update rule $\Delta W \propto (C + KK^\top)^{-1}$. As proven in Sec.~\ref{sec:key_analysis}, the inverse covariance matrix $C^{-1}$ acts as an anisotropic filter that amplifies minimal prompt variations into massive deviations. To eliminate this pathology, RoSE proposes \textbf{Isotropic Geometric Alignment (IGA)}, substituting $C$ with an identity matrix $I$. The reformulated update rule is:
\begin{equation}
    \Delta W_{\text{IGA}} = (V - W_0^l K) K^\top (I + K K^\top)^{-1}
    \label{eqn:rose_update}
\end{equation}
This modification serves two critical functions, one geometric and one mathematical:

\noindent\textbf{Closing the Covariance Trap.} Geometrically, replacing $C$ with $I$  enforces an isotropic response to input perturbations, ensuring that the projection of the key difference vector $\delta$ is uniform across all directions. Consequently, the noise inherent in instructional prompts is no longer selectively amplified, minimizing $D$.

\noindent\textbf{Regularization for Invertibility.} Mathematically, the term $I$ is strictly necessary for numerical stability. In practical editing scenarios, the Gram matrix $K K^\top$ is typically rank-deficient and singular due to the similarity of $k$ vectors for the same-subject facts. Directly removing $C$ would hinder the inverse calculation. By adding $I$, we effectively apply the Tikhonov Regularization to the update rule. This ensures that the matrix $(I + K K^\top)$ is full-rank, guaranteeing a unique and stable solution for the update $\Delta W_{RoSE}$.

\begin{table*}[ht]
\centering
\resizebox{\textwidth}{!}{%
\begin{tabular}{llcccc}
\hline \toprule
\textbf{Model} & \textbf{Method} & \textbf{Efficacy $\uparrow$} & \textbf{Paraphrase $\uparrow$} & \textbf{Locality $\uparrow$} & \textbf{Overall $\uparrow$} \\
\midrule
\multirow{6}{*}{Qwen2.5-7B-Instruct}
& Vanilla & 26.0 ($\pm$0.0) & 38.7 ($\pm$0.0) & \textbf{64.3} ($\pm$0.0) & 43.0 ($\pm$0.0) \\
& MEMIT & 50.5 ($\pm$0.3) & 44.7 ($\pm$0.1) & 63.6 ($\pm$0.1) & 51.8 ($\pm$0.1) \\
& AlphaEdit & 63.8 ($\pm$0.1) & 49.6 ($\pm$0.3) & 63.5 ($\pm$0.1) & 58.5 ($\pm$0.8) \\
& MEMIT-Merge & 75.5 ($\pm$0.5) & 52.8 ($\pm$0.7) & 63.8 ($\pm$0.1) & 64.0 ($\pm$0.4) \\
& \textbf{RoSE (Ours)} & \textbf{86.7} ($\pm$0.2) & \textbf{63.8} ($\pm$0.3) & 63.3 ($\pm$0.0) & \textbf{72.2} ($\pm$0.2) \\
\midrule
\multirow{6}{*}{Llama3.1-8B-Instruct}
& Vanilla & 22.4 ($\pm$0.0) & 41.2 ($\pm$0.0) & \textbf{65.3} ($\pm$0.0) & 43.0 ($\pm$0.0) \\
& MEMIT & 35.9($\pm$ 0.3) & 44.4($\pm$ 0.2) & 65.2($\pm$ 0.1) & 48.5($\pm$ 0.1) \\
& AlphaEdit & 34.3($\pm$ 0.5) & 44.1($\pm$ 0.3) & 65.2($\pm$ 0.1) & 47.9($\pm$ 0.2) \\
& MEMIT-Merge & 75.1($\pm$ 0.6) & 58.5($\pm$ 0.1) & 65.2($\pm$ 0.1) & 66.3($\pm$ 0.2) \\
& \textbf{RoSE (Ours)} & \textbf{87.3}($\pm$ 0.4) & \textbf{66.9}($\pm$ 0.2) & 64.8($\pm$ 0.0) & \textbf{73.0}($\pm$ 0.2) \\
\bottomrule \hline 
\end{tabular}%
}
\caption{Experimental instruction-following QA results on S2RKE across two models. Our method (\textbf{RoSE}) achieves the optimal performance across most key metrics. Standard deviations are shown in parentheses.}
\label{tab:s2rke_combined_results}
\end{table*}

\subsection{Hierarchical Knowledge Integration}
\label{sec:hki}

To address the Sharp Minimum caused by gradient orthogonality (Sec.~\ref{sec:value_analysis}), RoSE abandons flat joint optimization in favor of a structured, two-level \textbf{Hierarchical Knowledge Integration (HKI)}.

\noindent\textbf{Leaf-Level: Robust Centroid Estimation.}
For a specific relation $r_i$ of a subject, we aggregate gradients from a diverse prompt set $\mathcal{P}_{r_i}$ of different linguistic forms to compute a \textbf{Robust Centroid} $g'_{r_i}$, rather than optimizing on raw noisy gradients:
\begin{equation}
    g'_{r_i} = \frac{1}{|\mathcal{P}_{r_i}|} \sum_{p \in \mathcal{P}_{r_i}} \nabla_{\theta} \mathcal{L}(f_\theta(p), o_{r_i})
\end{equation}
Geometrically, $g'_{r_i}$ points towards the center of the feasible region for relation $r_i$, expanding the solution space before intersections with others. This aggregation smooths out the high-frequency noise of specific phrasings, extracting the semantic essence of the relation.

\noindent\textbf{Root-Level: Intersection Expansion.}
At the root level, we optimize the model using these centroids $\{g'_{r_1}, g'_{r_2}, \dots\}$ for each relation $r_i$. Unlike standard methods that try to intersect conflicting hyperplanes (which results in a tiny, sharp solution space), HKI intersects a small number of stabilized semantic subspaces defined by the centroids: $\mathcal{S}_{\text{HKI}} = \bigcap_{i} \text{Neighborhood}(g'_{r_i})$. This structured integration prevents the solution space from collapsing, flattening the loss landscape and increasing the Tolerance Radius.

\section{Experiments}
\label{sec:experiments}

We conduct a series of experiments to validate the effectiveness of RoSE. Our evaluation is designed to answer three primary questions: (\textbf{Q1}) Does RoSE improve instruction-following performance on the standard same-subject editing benchmark? (\textbf{Q2}) How does RoSE perform in more practical challenging scenarios involving character-oriented conversational knowledge? (\textbf{Q3}) Are both core components of RoSE, Isotropic Geometric Alignment and Hierarchical Knowledge Integration, essential to its performance?

\subsection{Experimental Setup}

\noindent\textbf{Datasets.}
To provide a comprehensive assessment, we use two distinct datasets. The first is \textbf{S2RKE}~\cite{duan2025relatedknowledgeperturbationmatters}, a standard benchmark for evaluating multi-relation edits on the same subject entity. To specifically test the central claims of this paper, we evaluate the edits on all original metrics using a held-out set of 2,000 generated instructional-style queries.
The second dataset, which we introduce as a more challenging practical test, is \textbf{LoCoMo-Edit}. Derived from the long-form conversational dataset LoCoMo~\cite{maharana2024evaluatinglongtermconversationalmemory}, LoCoMo-Edit simulates real-world demands by featuring complex, long-term targets of the same character from conversations.
LoCoMo-Edit serves to evaluate an editor's ability to internalize and generalize complex, contextual information, a critical step towards creating personalized AI with parametric conversational memory. Details can be referred to App.~\ref{app:datasets}.

\noindent\textbf{Models.}
Our experiments are conducted on two widely used open-source large language models: Llama-3.1-8B-Instruct~\cite{grattafiori2024llama3herdmodels} and Qwen2.5-7B-Instruct~\cite{qwen2025qwen25technicalreport}.

\noindent\textbf{Baselines.}
We compare RoSE against a suite of strong locate-then-edit methods, including the foundational single-fact editor \textbf{ROME}~\cite{meng2022locating}, its batch-editing extension \textbf{MEMIT}~\cite{Meng2022MassEditingMI}, and another recent approach, \textbf{AlphaEdit}~\citep{fang2024alphaeditnullspaceconstrainedknowledge}. Our primary and most competitive baseline is \textbf{MEMIT-Merge}~\cite{dong2025memitmergeaddressingmemitskeyvalue}, an adaptation of MEMIT designed for same-subject joint optimization.

\noindent\textbf{Evaluation Metrics.}
We employ a tailored evaluation protocol for each dataset. On S2RKE, we report the standard metrics of Efficacy, Paraphrase, and Locality. For the conversational knowledge in LoCoMo-Edit, we adopt a nuanced question-answering (QA) evaluation, including three knowledge types: Single-Hop, Multi-Hop, and Temporal. F1 exact match score is reported for QA on LoCoMo-Edit. Detailed definitions and the full experimental protocol are provided in App.~\ref{app:metrics}.





\begin{table*}[t]
\centering
\resizebox{\textwidth}{!}{%
\begin{tabular}{llcccc}
\hline \toprule
\textbf{Model} & \textbf{Method} & \textbf{Single Hop $\uparrow$} & \textbf{Multi Hop $\uparrow$} & \textbf{Temporal $\uparrow$} & \textbf{Overall $\uparrow$} \\
\midrule
\multirow{4}{*}{Qwen2.5-7B-Instruct}
& Vanilla & 12.9 ($\pm$0.1) & 13.7 ($\pm$0.2) & 5.1 ($\pm$0.5) & 11.3 ($\pm$0.2) \\
& MEMIT & 15.1 ($\pm$0.6) & 15.7 ($\pm$1.4) & 8.7 ($\pm$1.2) & 13.8 ($\pm$0.5) \\
& AlphaEdit & 7.5 ($\pm$1.4) & 12.9($\pm$2.8) & 8.5 ($\pm$2.1)& 8.8($\pm$1.8) \\
& MEMIT-Merge & 17.2 ($\pm$0.4) & 17.3 ($\pm$1.5) & 11.2 ($\pm$0.8) & 15.9 ($\pm$0.7) \\
& \textbf{RoSE (Ours)} & \textbf{22.3} ($\pm$0.3) & \textbf{23.5} ($\pm$1.6) & \textbf{19.1} ($\pm$0.9) & \textbf{21.9} ($\pm$0.3) \\
\midrule
\multirow{4}{*}{Llama3.1-8B-Instruct}
& Vanilla & 11.0($\pm$ 0.2) & 12.8($\pm$ 0.1) & 1.6($\pm$ 0.2) & 9.4($\pm$ 0.1) \\
& MEMIT & 12.4($\pm$ 0.0) & 14.0($\pm$ 0.3) & 1.5($\pm$ 0.3) & 10.4($\pm$ 0.0) \\
& AlphaEdit & 11.4($\pm$ 0.3) & 12.7($\pm$ 0.7) & 1.5($\pm$ 0.3) & 9.5($\pm$ 0.3) \\
& MEMIT-Merge & 16.6($\pm$ 0.8) & 19.2($\pm$ 0.4) & 4.0($\pm$ 0.9) & 14.4($\pm$ 0.6) \\
& \textbf{RoSE (Ours)} & \textbf{26.9}($\pm$ 0.7) & \textbf{28.7}($\pm$ 0.2) & \textbf{14.7}($\pm$ 1.3) & \textbf{24.5}($\pm$ 0.5) \\
\bottomrule \hline
\end{tabular}%
}
\caption{Experimental instruction-following QA performance on LoCoMo-Edit across two models. Our method (\textbf{RoSE}) achieves the best performance across all crucial metrics. Standard deviations are shown in parentheses.}
\label{tab:locomo_combined_results}
\end{table*}

\subsection{Main Results}

Our main results, presented in Tab.~\ref{tab:s2rke_combined_results} and Tab.~\ref{tab:locomo_combined_results}, demonstrate RoSE's consistent and significant superiority across both benchmarks and models.

(\textbf{Q1}) On the standard S2RKE benchmark, RoSE achieves a decisive advantage in Efficacy and Paraphrase in instruction following. Compared to the strongest baseline MEMIT-Merge, RoSE achieves substantial gains of over 11 points in both Efficacy (86.7\% vs. 75.5\%) and Paraphrase (63.8\% vs. 52.8\%). Crucially, these significant gains come at almost no cost to model stability, as RoSE's Locality score (63.3\%) remains on par with all baselines. 

(\textbf{Q2}) The advantage of our approach is magnified on the more challenging LoCoMo-Edit stress test. Baseline methods show only marginal gains, with MEMIT-Merge achieving an overall F1 score of just 15.9\%/14.4\%. In contrast, RoSE is the only method to successfully recall complex conversational knowledge. It significantly outperforms all baselines across every metric, achieving a notably higher overall F1 score of 21.9\%/24.5\%. 

\begin{figure}[t]
   \centering
   \subcaptionbox{Qwen2.5-7B-Instruct.\label{fig:abla_qwen}}[.5\linewidth][c]{%
      \includegraphics[width=1.\linewidth]{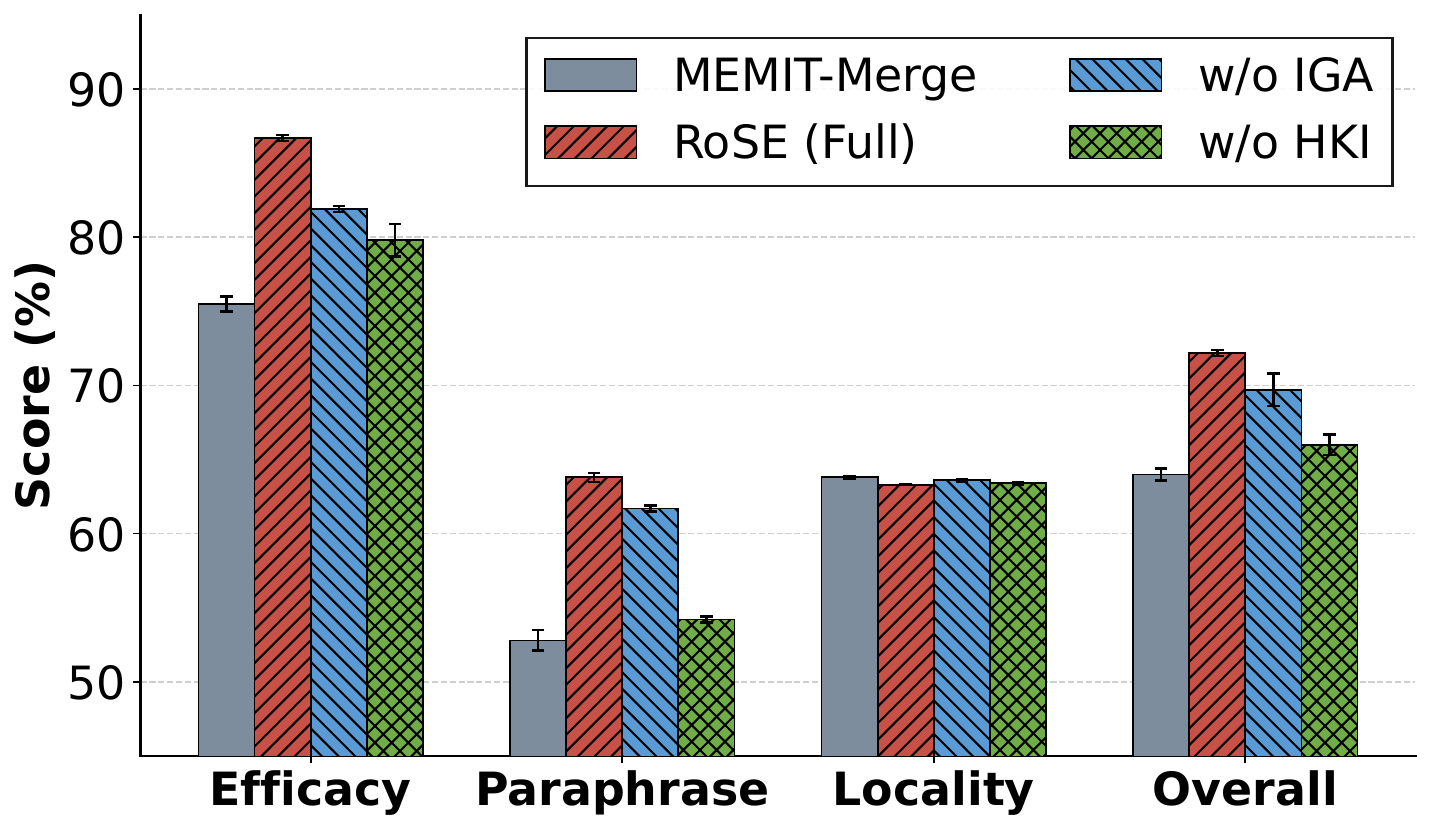}
   }
   \hspace{-0.2cm}
   \subcaptionbox{Llama-3.1-8B-Instruct.\label{fig:abla_llama}}[.5\linewidth][c]{%
      \includegraphics[width=1.\linewidth]{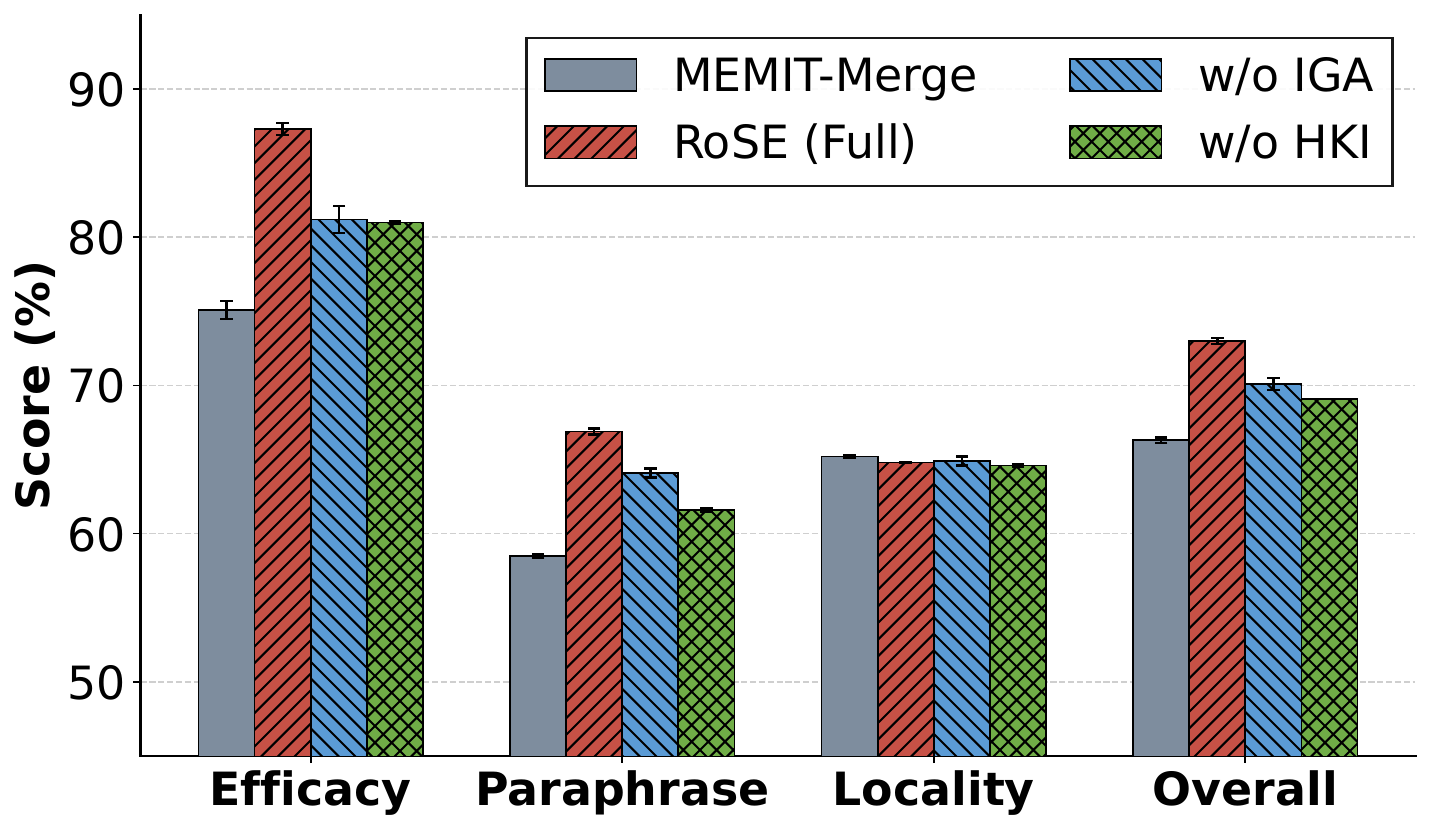}
   }
   \caption{Ablation study on the core components of RoSE on S2RKE. Both components are shown to be crucial for achieving maximum robustness.}
   \label{tab:ablation}
\end{figure}

\begin{figure}[t]
   \centering
   \subcaptionbox{Qwen2.5-7B-Instruct.\label{fig:geo_val_qwen}}[.5\linewidth][c]{%
      \includegraphics[width=1.\linewidth]{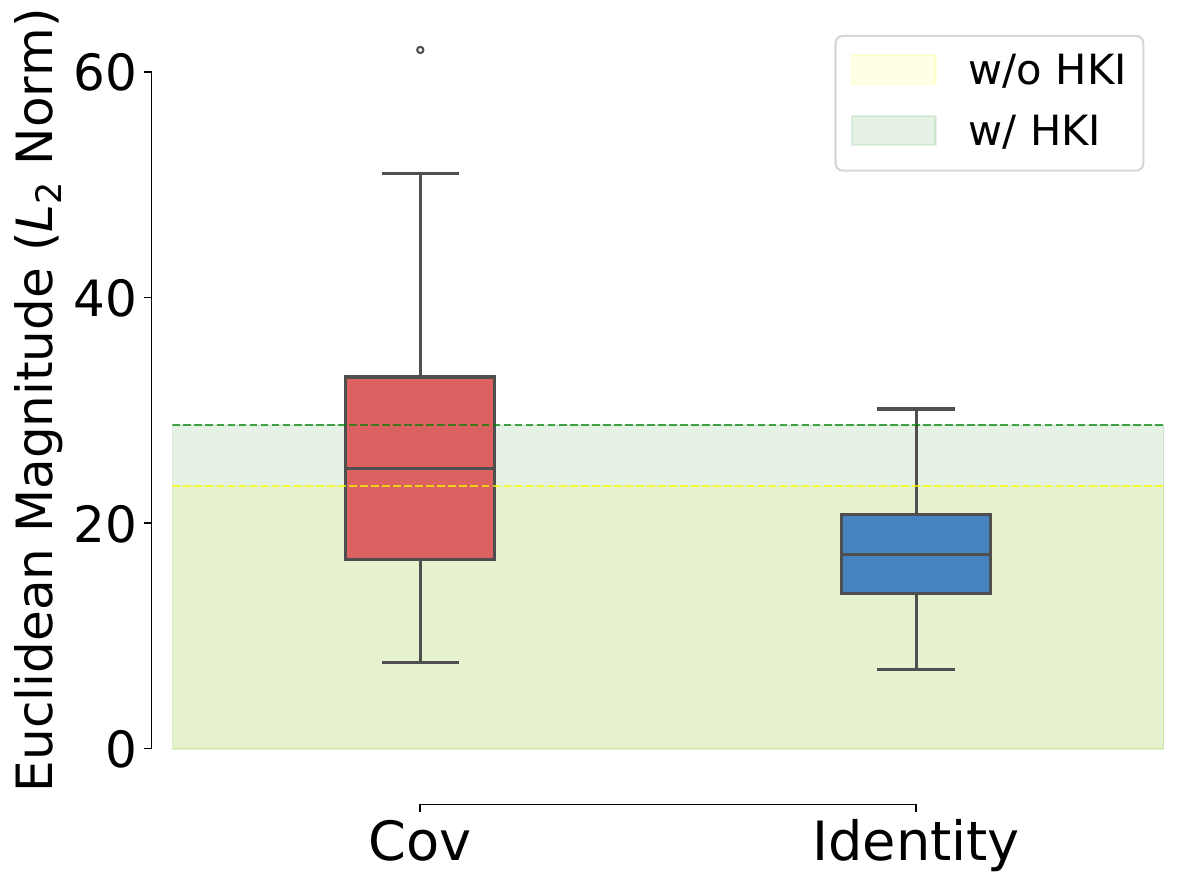}
   }
   \hspace{-0.2cm}
   \subcaptionbox{Llama-3.1-8B-Instruct.\label{fig:geo_val_llama}}[.5\linewidth][c]{%
      \includegraphics[width=1.\linewidth]{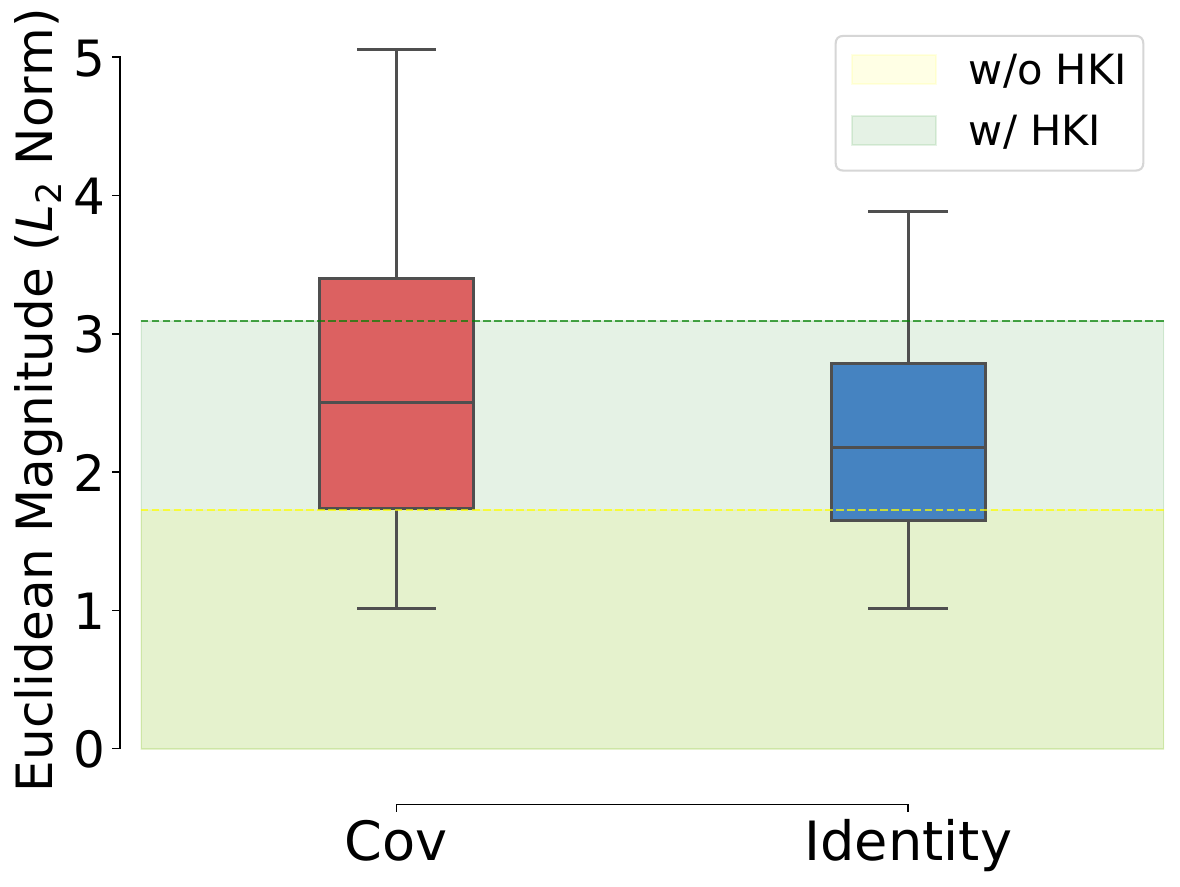}
   }
   \caption{Geometric validation of RoSE: IGA suppresses deviation (blue box), while HKI expands the tolerance basin (green zone). }
   \label{fig:geometric_validation}
\end{figure}

\noindent\textbf{Validating the Geometric Restoration ($D \le R$).}
Fig.~\ref{fig:geometric_validation} visualizes the geometric restoration achieved by RoSE. IGA suppresses the input noise, reducing Activation Deviation $D$ (blue box), and HKI expands the solution space, raising the average Tolerance Radius $R$ (green zone). Crucially, the blue box falls almost entirely within the Green zone, confirming that RoSE successfully encapsulates instruction perturbations within the model's valid editing boundaries ($D \le R$).

In summary, these results offer compelling evidence that RoSE pioneers a path toward truly robust and practical knowledge editing.

\subsection{Ablation Study}
\label{sec:ablation}

(\textbf{Q3}) Our ablation study, presented in Fig.~\ref{tab:ablation}, validates the crucial role of both RoSE components. Removing the Isotropic Geometric Alignment (- w/o IGA) causes a clear degradation in both Efficacy and Paraphrase, confirming that removing the Covariance Trap is vital to unlock generalization. Likewise, removing Hierarchical Knowledge Integration (- w/o HKI) also leads to a significant drop in both metrics, demonstrating its necessity for overcoming overfitting caused by gradient orthogonality to ensure knowledge is deeply and generally learned. Notably, while both ablated variants outperform MEMIT-Merge, their synergistic combination (full RoSE) is essential for achieving optimal instruction-following capabilities.

\section{Conclusion}
\label{sec:conclusion}
This work deconstructs instruction-following failures in same-subject knowledge editing as a geometric pathology where the activation deviation exceeds the model's tolerance radius. We identify the Covariance Trap and gradient orthogonality as root causes driving this imbalance. Our proposed RoSE strategy systematically resolves this by enforcing isotropic alignment to suppress deviation and integrating hierarchical knowledge to expand tolerance. By restoring the stable condition, RoSE achieves the optimal instruction-following capability. This geometric perspective not only solves immediate editing challenges but also lays a theoretical foundation for future research in lifelong learning of same-subject facts, aspiring towards self-evolving LLM agents that are plastic to new knowledge and structurally robust to diverse queries.

\section*{Limitations}
Despite its success, RoSE has limitations. Our HKI component slightly increases computational overhead due to its tree-like knowledge integration per edit. Furthermore, our current approach is focused on editing simple structured knowledge and has not been adapted to more complex or unstructured knowledge. Finally, a key limitation of our framework is its reliance on batch editing, as all facts concerning a single entity must be updated concurrently, rendering it unsuitable for true lifelong learning scenarios involving sequential information acquisition. These aspects present clear directions for future research and refinement.

\bibliography{custom}

@misc{minaee2024largelanguagemodelssurvey,
      title={Large Language Models: A Survey}, 
      author={Shervin Minaee and Tomas Mikolov and Narjes Nikzad and Meysam Chenaghlu and Richard Socher and Xavier Amatriain and Jianfeng Gao},
      year={2024},
      eprint={2402.06196},
      archivePrefix={arXiv},
      primaryClass={cs.CL},
      url={https://arxiv.org/abs/2402.06196}, 
}

@misc{zhao2024surveylargelanguagemodels,
      title={A Survey of Large Language Models}, 
      author={Wayne Xin Zhao and Kun Zhou and Junyi Li and Tianyi Tang and Xiaolei Wang and Yupeng Hou and Yingqian Min and Beichen Zhang and Junjie Zhang and Zican Dong and Yifan Du and Chen Yang and Yushuo Chen and Zhipeng Chen and Jinhao Jiang and Ruiyang Ren and Yifan Li and Xinyu Tang and Zikang Liu and Peiyu Liu and Jian-Yun Nie and Ji-Rong Wen},
      year={2024},
      eprint={2303.18223},
      archivePrefix={arXiv},
      primaryClass={cs.CL},
      url={https://arxiv.org/abs/2303.18223}, 
}

@inproceedings{heinzerling-inui-2021-language,
    title = "Language Models as Knowledge Bases: On Entity Representations, Storage Capacity, and Paraphrased Queries",
    author = "Heinzerling, Benjamin  and
      Inui, Kentaro",
    editor = "Merlo, Paola  and
      Tiedemann, Jorg  and
      Tsarfaty, Reut",
    booktitle = "Proceedings of the 16th Conference of the European Chapter of the Association for Computational Linguistics: Main Volume",
    month = apr,
    year = "2021",
    address = "Online",
    publisher = "Association for Computational Linguistics",
    url = "https://aclanthology.org/2021.eacl-main.153",
    doi = "10.18653/v1/2021.eacl-main.153",
    pages = "1772--1791",
}

@inproceedings{wang-etal-2021-generative,
    title = "Can Generative Pre-trained Language Models Serve As Knowledge Bases for Closed-book {QA}?",
    author = "Wang, Cunxiang  and
      Liu, Pai  and
      Zhang, Yue",
    editor = "Zong, Chengqing  and
      Xia, Fei  and
      Li, Wenjie  and
      Navigli, Roberto",
    booktitle = "Proceedings of the 59th Annual Meeting of the Association for Computational Linguistics and the 11th International Joint Conference on Natural Language Processing (Volume 1: Long Papers)",
    month = aug,
    year = "2021",
    address = "Online",
    publisher = "Association for Computational Linguistics",
    url = "https://aclanthology.org/2021.acl-long.251",
    doi = "10.18653/v1/2021.acl-long.251",
    pages = "3241--3251",
}

@misc{liu2024relationknowsrethinkingrecall,
      title={Relation Also Knows: Rethinking the Recall and Editing of Factual Associations in Auto-Regressive Transformer Language Models}, 
      author={Xiyu Liu and Zhengxiao Liu and Naibin Gu and Zheng Lin and Wanli Ma and Ji Xiang and Weiping Wang},
      year={2024},
      eprint={2408.15091},
      archivePrefix={arXiv},
      primaryClass={cs.CL},
      url={https://arxiv.org/abs/2408.15091}, 
}

@article{meng2022locating,
  title={Locating and Editing Factual Associations in {GPT}},
  author={Kevin Meng and David Bau and Alex Andonian and Yonatan Belinkov},
  journal={Advances in Neural Information Processing Systems},
  volume={35},
  year={2022}
}

@article{Meng2022MassEditingMI,
  title={Mass-Editing Memory in a Transformer},
  author={Kevin Meng and Arnab Sharma and Alex Andonian and Yonatan Belinkov and David Bau},
  journal={ArXiv},
  year={2022},
  volume={abs/2210.07229},
  url={https://api.semanticscholar.org/CorpusID:252873467}
}

@misc{fang2024alphaeditnullspaceconstrainedknowledge,
      title={AlphaEdit: Null-Space Constrained Knowledge Editing for Language Models}, 
      author={Junfeng Fang and Houcheng Jiang and Kun Wang and Yunshan Ma and Xiang Wang and Xiangnan He and Tat-seng Chua},
      year={2024},
      eprint={2410.02355},
      archivePrefix={arXiv},
      primaryClass={cs.CL},
      url={https://arxiv.org/abs/2410.02355}, 
}

@misc{zhang2024comprehensivestudyknowledgeediting,
      title={A Comprehensive Study of Knowledge Editing for Large Language Models}, 
      author={Ningyu Zhang and Yunzhi Yao and Bozhong Tian and Peng Wang and Shumin Deng and Mengru Wang and Zekun Xi and Shengyu Mao and Jintian Zhang and Yuansheng Ni and Siyuan Cheng and Ziwen Xu and Xin Xu and Jia-Chen Gu and Yong Jiang and Pengjun Xie and Fei Huang and Lei Liang and Zhiqiang Zhang and Xiaowei Zhu and Jun Zhou and Huajun Chen},
      year={2024},
      eprint={2401.01286},
      archivePrefix={arXiv},
      primaryClass={cs.CL},
      url={https://arxiv.org/abs/2401.01286}, 
}

@inproceedings{yao-etal-2023-editing,
    title = "Editing Large Language Models: Problems, Methods, and Opportunities",
    author = "Yao, Yunzhi  and
      Wang, Peng  and
      Tian, Bozhong  and
      Cheng, Siyuan  and
      Li, Zhoubo  and
      Deng, Shumin  and
      Chen, Huajun  and
      Zhang, Ningyu",
    editor = "Bouamor, Houda  and
      Pino, Juan  and
      Bali, Kalika",
    booktitle = "Proceedings of the 2023 Conference on Empirical Methods in Natural Language Processing",
    month = dec,
    year = "2023",
    address = "Singapore",
    publisher = "Association for Computational Linguistics",
    url = "https://aclanthology.org/2023.emnlp-main.632",
    doi = "10.18653/v1/2023.emnlp-main.632",
    pages = "10222--10240",
}

@inproceedings{geva-etal-2023-dissecting,
    title = "Dissecting Recall of Factual Associations in Auto-Regressive Language Models",
    author = "Geva, Mor  and
      Bastings, Jasmijn  and
      Filippova, Katja  and
      Globerson, Amir",
    editor = "Bouamor, Houda  and
      Pino, Juan  and
      Bali, Kalika",
    booktitle = "Proceedings of the 2023 Conference on Empirical Methods in Natural Language Processing",
    month = dec,
    year = "2023",
    address = "Singapore",
    publisher = "Association for Computational Linguistics",
    url = "https://aclanthology.org/2023.emnlp-main.751",
    doi = "10.18653/v1/2023.emnlp-main.751",
    pages = "12216--12235",
}

@misc{ma2024perturbationrestrainedsequentialmodelediting,
      title={Perturbation-Restrained Sequential Model Editing}, 
      author={Jun-Yu Ma and Hong Wang and Hao-Xiang Xu and Zhen-Hua Ling and Jia-Chen Gu},
      year={2024},
      eprint={2405.16821},
      archivePrefix={arXiv},
      primaryClass={cs.CL},
      url={https://arxiv.org/abs/2405.16821}, 
}

@inproceedings{geva-etal-2022-transformer,
    title = "Transformer Feed-Forward Layers Build Predictions by Promoting Concepts in the Vocabulary Space",
    author = "Geva, Mor  and
      Caciularu, Avi  and
      Wang, Kevin  and
      Goldberg, Yoav",
    editor = "Goldberg, Yoav  and
      Kozareva, Zornitsa  and
      Zhang, Yue",
    booktitle = "Proceedings of the 2022 Conference on Empirical Methods in Natural Language Processing",
    month = dec,
    year = "2022",
    address = "Abu Dhabi, United Arab Emirates",
    publisher = "Association for Computational Linguistics",
    url = "https://aclanthology.org/2022.emnlp-main.3",
    doi = "10.18653/v1/2022.emnlp-main.3",
    pages = "30--45",
}

@misc{cai2024oeditorthogonalsubspaceediting,
      title={O-Edit: Orthogonal Subspace Editing for Language Model Sequential Editing}, 
      author={Yuchen Cai and Ding Cao},
      year={2024},
      eprint={2410.11469},
      archivePrefix={arXiv},
      primaryClass={cs.CL},
      url={https://arxiv.org/abs/2410.11469}, 
}

@misc{wang2024knowledgeeditinglargelanguage,
      title={Knowledge Editing for Large Language Models: A Survey}, 
      author={Song Wang and Yaochen Zhu and Haochen Liu and Zaiyi Zheng and Chen Chen and Jundong Li},
      year={2024},
      eprint={2310.16218},
      archivePrefix={arXiv},
      primaryClass={cs.CL},
      url={https://arxiv.org/abs/2310.16218}, 
}

@misc{wang2024wiserethinkingknowledgememory,
      title={WISE: Rethinking the Knowledge Memory for Lifelong Model Editing of Large Language Models}, 
      author={Peng Wang and Zexi Li and Ningyu Zhang and Ziwen Xu and Yunzhi Yao and Yong Jiang and Pengjun Xie and Fei Huang and Huajun Chen},
      year={2024},
      eprint={2405.14768},
      archivePrefix={arXiv},
      primaryClass={cs.CL},
      url={https://arxiv.org/abs/2405.14768}, 
}

@inproceedings{10.5555/3666122.3668201,
author = {Hartvigsen, Thomas and Sankaranarayanan, Swami and Palangi, Hamid and Kim, Yoon and Ghassemi, Marzyeh},
title = {Aging with GRACE: lifelong model editing with discrete key-value adaptors},
year = {2023},
publisher = {Curran Associates Inc.},
address = {Red Hook, NY, USA},
booktitle = {Proceedings of the 37th International Conference on Neural Information Processing Systems},
articleno = {2079},
numpages = {26},
location = {New Orleans, LA, USA},
series = {NIPS '23}
}

@misc{mitchell2022fast,
      title={Fast Model Editing at Scale}, 
      author={Eric Mitchell and Charles Lin and Antoine Bosselut and Chelsea Finn and Christopher D. Manning},
      year={2022},
      eprint={2110.11309},
      archivePrefix={arXiv},
      primaryClass={cs.LG}
}

@inproceedings{roberts-etal-2020-much,
    title = "How Much Knowledge Can You Pack Into the Parameters of a Language Model?",
    author = "Roberts, Adam  and
      Raffel, Colin  and
      Shazeer, Noam",
    editor = "Webber, Bonnie  and
      Cohn, Trevor  and
      He, Yulan  and
      Liu, Yang",
    booktitle = "Proceedings of the 2020 Conference on Empirical Methods in Natural Language Processing (EMNLP)",
    month = nov,
    year = "2020",
    address = "Online",
    publisher = "Association for Computational Linguistics",
    url = "https://aclanthology.org/2020.emnlp-main.437",
    doi = "10.18653/v1/2020.emnlp-main.437",
    pages = "5418--5426",
}

@misc{zhang2025uncoveringoverfittinglargelanguage,
      title={Uncovering Overfitting in Large Language Model Editing}, 
      author={Mengqi Zhang and Xiaotian Ye and Qiang Liu and Pengjie Ren and Shu Wu and Zhumin Chen},
      year={2025},
      eprint={2410.07819},
      archivePrefix={arXiv},
      primaryClass={cs.CL},
      url={https://arxiv.org/abs/2410.07819}, 
}

@misc{duan2025relatedknowledgeperturbationmatters,
      title={Related Knowledge Perturbation Matters: Rethinking Multiple Pieces of Knowledge Editing in Same-Subject}, 
      author={Zenghao Duan and Wenbin Duan and Zhiyi Yin and Yinghan Shen and Shaoling Jing and Jie Zhang and Huawei Shen and Xueqi Cheng},
      year={2025},
      eprint={2502.06868},
      archivePrefix={arXiv},
      primaryClass={cs.CL},
      url={https://arxiv.org/abs/2502.06868}, 
}

@misc{dong2025memitmergeaddressingmemitskeyvalue,
      title={MEMIT-Merge: Addressing MEMIT's Key-Value Conflicts in Same-Subject Batch Editing for LLMs}, 
      author={Zilu Dong and Xiangqing Shen and Rui Xia},
      year={2025},
      eprint={2502.07322},
      archivePrefix={arXiv},
      primaryClass={cs.CL},
      url={https://arxiv.org/abs/2502.07322}, 
}

@misc{qwen2025qwen25technicalreport,
      title={Qwen2.5 Technical Report}, 
      author={Qwen and : and An Yang and Baosong Yang and Beichen Zhang and Binyuan Hui and Bo Zheng and Bowen Yu and Chengyuan Li and Dayiheng Liu and Fei Huang and Haoran Wei and Huan Lin and Jian Yang and Jianhong Tu and Jianwei Zhang and Jianxin Yang and Jiaxi Yang and Jingren Zhou and Junyang Lin and Kai Dang and Keming Lu and Keqin Bao and Kexin Yang and Le Yu and Mei Li and Mingfeng Xue and Pei Zhang and Qin Zhu and Rui Men and Runji Lin and Tianhao Li and Tianyi Tang and Tingyu Xia and Xingzhang Ren and Xuancheng Ren and Yang Fan and Yang Su and Yichang Zhang and Yu Wan and Yuqiong Liu and Zeyu Cui and Zhenru Zhang and Zihan Qiu},
      year={2025},
      eprint={2412.15115},
      archivePrefix={arXiv},
      primaryClass={cs.CL},
      url={https://arxiv.org/abs/2412.15115}, 
}

@misc{maharana2024evaluatinglongtermconversationalmemory,
      title={Evaluating Very Long-Term Conversational Memory of LLM Agents}, 
      author={Adyasha Maharana and Dong-Ho Lee and Sergey Tulyakov and Mohit Bansal and Francesco Barbieri and Yuwei Fang},
      year={2024},
      eprint={2402.17753},
      archivePrefix={arXiv},
      primaryClass={cs.CL},
      url={https://arxiv.org/abs/2402.17753}, 
}

@misc{grattafiori2024llama3herdmodels,
      title={The Llama 3 Herd of Models}, 
      author={Aaron Grattafiori and Abhimanyu Dubey and Abhinav Jauhri and Abhinav Pandey and Abhishek Kadian and Ahmad Al-Dahle and Aiesha Letman and Akhil Mathur and Alan Schelten and Alex Vaughan and Amy Yang and Angela Fan and Anirudh Goyal and Anthony Hartshorn and Aobo Yang and Archi Mitra and Archie Sravankumar and Artem Korenev and Arthur Hinsvark and Arun Rao and Aston Zhang and Aurelien Rodriguez and Austen Gregerson and Ava Spataru and Baptiste Roziere and Bethany Biron and Binh Tang and Bobbie Chern and Charlotte Caucheteux and Chaya Nayak and Chloe Bi and Chris Marra and Chris McConnell and Christian Keller and Christophe Touret and Chunyang Wu and Corinne Wong and Cristian Canton Ferrer and Cyrus Nikolaidis and Damien Allonsius and Daniel Song and Danielle Pintz and Danny Livshits and Danny Wyatt and David Esiobu and Dhruv Choudhary and Dhruv Mahajan and Diego Garcia-Olano and Diego Perino and Dieuwke Hupkes and Egor Lakomkin and Ehab AlBadawy and Elina Lobanova and Emily Dinan and Eric Michael Smith and Filip Radenovic and Francisco Guzmán and Frank Zhang and Gabriel Synnaeve and Gabrielle Lee and Georgia Lewis Anderson and Govind Thattai and Graeme Nail and Gregoire Mialon and Guan Pang and Guillem Cucurell and Hailey Nguyen and Hannah Korevaar and Hu Xu and Hugo Touvron and Iliyan Zarov and Imanol Arrieta Ibarra and Isabel Kloumann and Ishan Misra and Ivan Evtimov and Jack Zhang and Jade Copet and Jaewon Lee and Jan Geffert and Jana Vranes and Jason Park and Jay Mahadeokar and Jeet Shah and Jelmer van der Linde and Jennifer Billock and Jenny Hong and Jenya Lee and Jeremy Fu and Jianfeng Chi and Jianyu Huang and Jiawen Liu and Jie Wang and Jiecao Yu and Joanna Bitton and Joe Spisak and Jongsoo Park and Joseph Rocca and Joshua Johnstun and Joshua Saxe and Junteng Jia and Kalyan Vasuden Alwala and Karthik Prasad and Kartikeya Upasani and Kate Plawiak and Ke Li and Kenneth Heafield and Kevin Stone and Khalid El-Arini and Krithika Iyer and Kshitiz Malik and Kuenley Chiu and Kunal Bhalla and Kushal Lakhotia and Lauren Rantala-Yeary and Laurens van der Maaten and Lawrence Chen and Liang Tan and Liz Jenkins and Louis Martin and Lovish Madaan and Lubo Malo and Lukas Blecher and Lukas Landzaat and Luke de Oliveira and Madeline Muzzi and Mahesh Pasupuleti and Mannat Singh and Manohar Paluri and Marcin Kardas and Maria Tsimpoukelli and Mathew Oldham and Mathieu Rita and Maya Pavlova and Melanie Kambadur and Mike Lewis and Min Si and Mitesh Kumar Singh and Mona Hassan and Naman Goyal and Narjes Torabi and Nikolay Bashlykov and Nikolay Bogoychev and Niladri Chatterji and Ning Zhang and Olivier Duchenne and Onur Çelebi and Patrick Alrassy and Pengchuan Zhang and Pengwei Li and Petar Vasic and Peter Weng and Prajjwal Bhargava and Pratik Dubal and Praveen Krishnan and Punit Singh Koura and Puxin Xu and Qing He and Qingxiao Dong and Ragavan Srinivasan and Raj Ganapathy and Ramon Calderer and Ricardo Silveira Cabral and Robert Stojnic and Roberta Raileanu and Rohan Maheswari and Rohit Girdhar and Rohit Patel and Romain Sauvestre and Ronnie Polidoro and Roshan Sumbaly and Ross Taylor and Ruan Silva and Rui Hou and Rui Wang and Saghar Hosseini and Sahana Chennabasappa and Sanjay Singh and Sean Bell and Seohyun Sonia Kim and Sergey Edunov and Shaoliang Nie and Sharan Narang and Sharath Raparthy and Sheng Shen and Shengye Wan and Shruti Bhosale and Shun Zhang and Simon Vandenhende and Soumya Batra and Spencer Whitman and Sten Sootla and Stephane Collot and Suchin Gururangan and Sydney Borodinsky and Tamar Herman and Tara Fowler and Tarek Sheasha and Thomas Georgiou and Thomas Scialom and Tobias Speckbacher and Todor Mihaylov and Tong Xiao and Ujjwal Karn and Vedanuj Goswami and Vibhor Gupta and Vignesh Ramanathan and Viktor Kerkez and Vincent Gonguet and Virginie Do and Vish Vogeti and Vítor Albiero and Vladan Petrovic and Weiwei Chu and Wenhan Xiong and Wenyin Fu and Whitney Meers and Xavier Martinet and Xiaodong Wang and Xiaofang Wang and Xiaoqing Ellen Tan and Xide Xia and Xinfeng Xie and Xuchao Jia and Xuewei Wang and Yaelle Goldschlag and Yashesh Gaur and Yasmine Babaei and Yi Wen and Yiwen Song and Yuchen Zhang and Yue Li and Yuning Mao and Zacharie Delpierre Coudert and Zheng Yan and Zhengxing Chen and Zoe Papakipos and Aaditya Singh and Aayushi Srivastava and Abha Jain and Adam Kelsey and Adam Shajnfeld and Adithya Gangidi and Adolfo Victoria and Ahuva Goldstand and Ajay Menon and Ajay Sharma and Alex Boesenberg and Alexei Baevski and Allie Feinstein and Amanda Kallet and Amit Sangani and Amos Teo and Anam Yunus and Andrei Lupu and Andres Alvarado and Andrew Caples and Andrew Gu and Andrew Ho and Andrew Poulton and Andrew Ryan and Ankit Ramchandani and Annie Dong and Annie Franco and Anuj Goyal and Aparajita Saraf and Arkabandhu Chowdhury and Ashley Gabriel and Ashwin Bharambe and Assaf Eisenman and Azadeh Yazdan and Beau James and Ben Maurer and Benjamin Leonhardi and Bernie Huang and Beth Loyd and Beto De Paola and Bhargavi Paranjape and Bing Liu and Bo Wu and Boyu Ni and Braden Hancock and Bram Wasti and Brandon Spence and Brani Stojkovic and Brian Gamido and Britt Montalvo and Carl Parker and Carly Burton and Catalina Mejia and Ce Liu and Changhan Wang and Changkyu Kim and Chao Zhou and Chester Hu and Ching-Hsiang Chu and Chris Cai and Chris Tindal and Christoph Feichtenhofer and Cynthia Gao and Damon Civin and Dana Beaty and Daniel Kreymer and Daniel Li and David Adkins and David Xu and Davide Testuggine and Delia David and Devi Parikh and Diana Liskovich and Didem Foss and Dingkang Wang and Duc Le and Dustin Holland and Edward Dowling and Eissa Jamil and Elaine Montgomery and Eleonora Presani and Emily Hahn and Emily Wood and Eric-Tuan Le and Erik Brinkman and Esteban Arcaute and Evan Dunbar and Evan Smothers and Fei Sun and Felix Kreuk and Feng Tian and Filippos Kokkinos and Firat Ozgenel and Francesco Caggioni and Frank Kanayet and Frank Seide and Gabriela Medina Florez and Gabriella Schwarz and Gada Badeer and Georgia Swee and Gil Halpern and Grant Herman and Grigory Sizov and Guangyi and Zhang and Guna Lakshminarayanan and Hakan Inan and Hamid Shojanazeri and Han Zou and Hannah Wang and Hanwen Zha and Haroun Habeeb and Harrison Rudolph and Helen Suk and Henry Aspegren and Hunter Goldman and Hongyuan Zhan and Ibrahim Damlaj and Igor Molybog and Igor Tufanov and Ilias Leontiadis and Irina-Elena Veliche and Itai Gat and Jake Weissman and James Geboski and James Kohli and Janice Lam and Japhet Asher and Jean-Baptiste Gaya and Jeff Marcus and Jeff Tang and Jennifer Chan and Jenny Zhen and Jeremy Reizenstein and Jeremy Teboul and Jessica Zhong and Jian Jin and Jingyi Yang and Joe Cummings and Jon Carvill and Jon Shepard and Jonathan McPhie and Jonathan Torres and Josh Ginsburg and Junjie Wang and Kai Wu and Kam Hou U and Karan Saxena and Kartikay Khandelwal and Katayoun Zand and Kathy Matosich and Kaushik Veeraraghavan and Kelly Michelena and Keqian Li and Kiran Jagadeesh and Kun Huang and Kunal Chawla and Kyle Huang and Lailin Chen and Lakshya Garg and Lavender A and Leandro Silva and Lee Bell and Lei Zhang and Liangpeng Guo and Licheng Yu and Liron Moshkovich and Luca Wehrstedt and Madian Khabsa and Manav Avalani and Manish Bhatt and Martynas Mankus and Matan Hasson and Matthew Lennie and Matthias Reso and Maxim Groshev and Maxim Naumov and Maya Lathi and Meghan Keneally and Miao Liu and Michael L. Seltzer and Michal Valko and Michelle Restrepo and Mihir Patel and Mik Vyatskov and Mikayel Samvelyan and Mike Clark and Mike Macey and Mike Wang and Miquel Jubert Hermoso and Mo Metanat and Mohammad Rastegari and Munish Bansal and Nandhini Santhanam and Natascha Parks and Natasha White and Navyata Bawa and Nayan Singhal and Nick Egebo and Nicolas Usunier and Nikhil Mehta and Nikolay Pavlovich Laptev and Ning Dong and Norman Cheng and Oleg Chernoguz and Olivia Hart and Omkar Salpekar and Ozlem Kalinli and Parkin Kent and Parth Parekh and Paul Saab and Pavan Balaji and Pedro Rittner and Philip Bontrager and Pierre Roux and Piotr Dollar and Polina Zvyagina and Prashant Ratanchandani and Pritish Yuvraj and Qian Liang and Rachad Alao and Rachel Rodriguez and Rafi Ayub and Raghotham Murthy and Raghu Nayani and Rahul Mitra and Rangaprabhu Parthasarathy and Raymond Li and Rebekkah Hogan and Robin Battey and Rocky Wang and Russ Howes and Ruty Rinott and Sachin Mehta and Sachin Siby and Sai Jayesh Bondu and Samyak Datta and Sara Chugh and Sara Hunt and Sargun Dhillon and Sasha Sidorov and Satadru Pan and Saurabh Mahajan and Saurabh Verma and Seiji Yamamoto and Sharadh Ramaswamy and Shaun Lindsay and Shaun Lindsay and Sheng Feng and Shenghao Lin and Shengxin Cindy Zha and Shishir Patil and Shiva Shankar and Shuqiang Zhang and Shuqiang Zhang and Sinong Wang and Sneha Agarwal and Soji Sajuyigbe and Soumith Chintala and Stephanie Max and Stephen Chen and Steve Kehoe and Steve Satterfield and Sudarshan Govindaprasad and Sumit Gupta and Summer Deng and Sungmin Cho and Sunny Virk and Suraj Subramanian and Sy Choudhury and Sydney Goldman and Tal Remez and Tamar Glaser and Tamara Best and Thilo Koehler and Thomas Robinson and Tianhe Li and Tianjun Zhang and Tim Matthews and Timothy Chou and Tzook Shaked and Varun Vontimitta and Victoria Ajayi and Victoria Montanez and Vijai Mohan and Vinay Satish Kumar and Vishal Mangla and Vlad Ionescu and Vlad Poenaru and Vlad Tiberiu Mihailescu and Vladimir Ivanov and Wei Li and Wenchen Wang and Wenwen Jiang and Wes Bouaziz and Will Constable and Xiaocheng Tang and Xiaojian Wu and Xiaolan Wang and Xilun Wu and Xinbo Gao and Yaniv Kleinman and Yanjun Chen and Ye Hu and Ye Jia and Ye Qi and Yenda Li and Yilin Zhang and Ying Zhang and Yossi Adi and Youngjin Nam and Yu and Wang and Yu Zhao and Yuchen Hao and Yundi Qian and Yunlu Li and Yuzi He and Zach Rait and Zachary DeVito and Zef Rosnbrick and Zhaoduo Wen and Zhenyu Yang and Zhiwei Zhao and Zhiyu Ma},
      year={2024},
      eprint={2407.21783},
      archivePrefix={arXiv},
      primaryClass={cs.AI},
      url={https://arxiv.org/abs/2407.21783}, 
}

@inproceedings{li-chu-2025-adaedit,
    title = "{A}da{E}dit: Advancing Continuous Knowledge Editing For Large Language Models",
    author = "Li, Qi  and
      Chu, Xiaowen",
    editor = "Che, Wanxiang  and
      Nabende, Joyce  and
      Shutova, Ekaterina  and
      Pilehvar, Mohammad Taher",
    booktitle = "Proceedings of the 63rd Annual Meeting of the Association for Computational Linguistics (Volume 1: Long Papers)",
    month = jul,
    year = "2025",
    address = "Vienna, Austria",
    publisher = "Association for Computational Linguistics",
    url = "https://aclanthology.org/2025.acl-long.208/",
    doi = "10.18653/v1/2025.acl-long.208",
    pages = "4127--4149",
    ISBN = "979-8-89176-251-0"
}

@misc{gupta2025efficientknowledgeeditingminimal,
      title={Efficient Knowledge Editing via Minimal Precomputation}, 
      author={Akshat Gupta and Maochuan Lu and Thomas Hartvigsen and Gopala Anumanchipalli},
      year={2025},
      eprint={2506.04226},
      archivePrefix={arXiv},
      primaryClass={cs.CL},
      url={https://arxiv.org/abs/2506.04226}, 
}

@misc{deepseekai2025deepseekr1incentivizingreasoningcapability,
      title={DeepSeek-R1: Incentivizing Reasoning Capability in LLMs via Reinforcement Learning}, 
      author={DeepSeek-AI and Daya Guo and Dejian Yang and Haowei Zhang and Junxiao Song and Ruoyu Zhang and Runxin Xu and Qihao Zhu and Shirong Ma and Peiyi Wang and Xiao Bi and Xiaokang Zhang and Xingkai Yu and Yu Wu and Z. F. Wu and Zhibin Gou and Zhihong Shao and Zhuoshu Li and Ziyi Gao and Aixin Liu and Bing Xue and Bingxuan Wang and Bochao Wu and Bei Feng and Chengda Lu and Chenggang Zhao and Chengqi Deng and Chenyu Zhang and Chong Ruan and Damai Dai and Deli Chen and Dongjie Ji and Erhang Li and Fangyun Lin and Fucong Dai and Fuli Luo and Guangbo Hao and Guanting Chen and Guowei Li and H. Zhang and Han Bao and Hanwei Xu and Haocheng Wang and Honghui Ding and Huajian Xin and Huazuo Gao and Hui Qu and Hui Li and Jianzhong Guo and Jiashi Li and Jiawei Wang and Jingchang Chen and Jingyang Yuan and Junjie Qiu and Junlong Li and J. L. Cai and Jiaqi Ni and Jian Liang and Jin Chen and Kai Dong and Kai Hu and Kaige Gao and Kang Guan and Kexin Huang and Kuai Yu and Lean Wang and Lecong Zhang and Liang Zhao and Litong Wang and Liyue Zhang and Lei Xu and Leyi Xia and Mingchuan Zhang and Minghua Zhang and Minghui Tang and Meng Li and Miaojun Wang and Mingming Li and Ning Tian and Panpan Huang and Peng Zhang and Qiancheng Wang and Qinyu Chen and Qiushi Du and Ruiqi Ge and Ruisong Zhang and Ruizhe Pan and Runji Wang and R. J. Chen and R. L. Jin and Ruyi Chen and Shanghao Lu and Shangyan Zhou and Shanhuang Chen and Shengfeng Ye and Shiyu Wang and Shuiping Yu and Shunfeng Zhou and Shuting Pan and S. S. Li and Shuang Zhou and Shaoqing Wu and Shengfeng Ye and Tao Yun and Tian Pei and Tianyu Sun and T. Wang and Wangding Zeng and Wanjia Zhao and Wen Liu and Wenfeng Liang and Wenjun Gao and Wenqin Yu and Wentao Zhang and W. L. Xiao and Wei An and Xiaodong Liu and Xiaohan Wang and Xiaokang Chen and Xiaotao Nie and Xin Cheng and Xin Liu and Xin Xie and Xingchao Liu and Xinyu Yang and Xinyuan Li and Xuecheng Su and Xuheng Lin and X. Q. Li and Xiangyue Jin and Xiaojin Shen and Xiaosha Chen and Xiaowen Sun and Xiaoxiang Wang and Xinnan Song and Xinyi Zhou and Xianzu Wang and Xinxia Shan and Y. K. Li and Y. Q. Wang and Y. X. Wei and Yang Zhang and Yanhong Xu and Yao Li and Yao Zhao and Yaofeng Sun and Yaohui Wang and Yi Yu and Yichao Zhang and Yifan Shi and Yiliang Xiong and Ying He and Yishi Piao and Yisong Wang and Yixuan Tan and Yiyang Ma and Yiyuan Liu and Yongqiang Guo and Yuan Ou and Yuduan Wang and Yue Gong and Yuheng Zou and Yujia He and Yunfan Xiong and Yuxiang Luo and Yuxiang You and Yuxuan Liu and Yuyang Zhou and Y. X. Zhu and Yanhong Xu and Yanping Huang and Yaohui Li and Yi Zheng and Yuchen Zhu and Yunxian Ma and Ying Tang and Yukun Zha and Yuting Yan and Z. Z. Ren and Zehui Ren and Zhangli Sha and Zhe Fu and Zhean Xu and Zhenda Xie and Zhengyan Zhang and Zhewen Hao and Zhicheng Ma and Zhigang Yan and Zhiyu Wu and Zihui Gu and Zijia Zhu and Zijun Liu and Zilin Li and Ziwei Xie and Ziyang Song and Zizheng Pan and Zhen Huang and Zhipeng Xu and Zhongyu Zhang and Zhen Zhang},
      year={2025},
      eprint={2501.12948},
      archivePrefix={arXiv},
      primaryClass={cs.CL},
      url={https://arxiv.org/abs/2501.12948}, 
}

@misc{hu2025memoryageaiagents,
      title={Memory in the Age of AI Agents}, 
      author={Yuyang Hu and Shichun Liu and Yanwei Yue and Guibin Zhang and Boyang Liu and Fangyi Zhu and Jiahang Lin and Honglin Guo and Shihan Dou and Zhiheng Xi and Senjie Jin and Jiejun Tan and Yanbin Yin and Jiongnan Liu and Zeyu Zhang and Zhongxiang Sun and Yutao Zhu and Hao Sun and Boci Peng and Zhenrong Cheng and Xuanbo Fan and Jiaxin Guo and Xinlei Yu and Zhenhong Zhou and Zewen Hu and Jiahao Huo and Junhao Wang and Yuwei Niu and Yu Wang and Zhenfei Yin and Xiaobin Hu and Yue Liao and Qiankun Li and Kun Wang and Wangchunshu Zhou and Yixin Liu and Dawei Cheng and Qi Zhang and Tao Gui and Shirui Pan and Yan Zhang and Philip Torr and Zhicheng Dou and Ji-Rong Wen and Xuanjing Huang and Yu-Gang Jiang and Shuicheng Yan},
      year={2025},
      eprint={2512.13564},
      archivePrefix={arXiv},
      primaryClass={cs.CL},
      url={https://arxiv.org/abs/2512.13564}, 
}

\appendix

\onecolumn
\section{Analysis of Error Amplification via the Covariance Matrix}
\label{app:cov_trap_analysis}

In this section, we provide a theoretical derivation to explain why the inclusion of the pre-statistics covariance matrix $C$ in the update rule leads to an amplification of the error term $D$, compared to an isotropic update (where $C$ is replaced by the identity matrix $I$).

Recall the update rule for the weights:
\begin{equation}
    \Delta W = W^l_{\text{out}} - W^l_0 = (V - W^l_0 K) K^\top (C + K K^\top)^{-1}.
\end{equation}
Let $R = V - W^l_0 K$ denote the residual vector. We represent the difference in input keys as $\delta = \tilde{k} - k_o$. The error norm induced by this difference is defined as:
\begin{equation}
    D = \| \Delta W \delta \|_2 = \| R K^\top (C + K K^\top)^{-1} \delta \|_2.
\end{equation}

To understand the behavior of $D$, we analyze the term $M_C \delta$, where $M_C = (C + K K^\top)^{-1}$. Since $KK^\top$ is a perturbation of extremely small rank relative to C, we can use the Sherman-Morrison-Woodbury formula to expand this inverse matrix as:
\begin{equation}
    (C + K K^\top)^{-1} = C^{-1} - C^{-1} K (I + K^\top C^{-1} K)^{-1} K^\top C^{-1}.
\end{equation}
Multiplying by the perturbation vector $\delta$, we obtain:
\begin{equation}
    \label{eq:woodbury_expansion}
    (C + K K^\top)^{-1} \delta = C^{-1} \delta - C^{-1} K \underbrace{(I + K^\top C^{-1} K)^{-1} (K^\top C^{-1} \delta)}_{\text{Correction Term}}.
\end{equation}

\paragraph{Approximation via Orthogonality in Whitened Space.}
Specifically, $\delta = \tilde{k} - k_o$ denote the deviation vector arising from distinct representations of the same unified knowledge. We analyze the interaction term $K^\top C^{-1} \delta$, which represents the inner product of $K$ and $\delta$ under the metric induced by the inverse covariance matrix $C^{-1}$ (i.e., the Mahalanobis metric). 

Geometrically, the matrix $C^{-1}$ acts as a whitening transformation that normalizes the global feature correlations. In this whitened feature space, the vector $K$ encodes the principal semantic direction of the target knowledge, while $\delta$ captures the nuisance variations (e.g., minor modal or syntactic shifts) that are statistically decoupled from the core semantics. Due to the high dimensionality of the feature space, such independent residual vectors are orthogonal to the semantic directions. Consequently, the projection of the deviation $\delta$ onto the editing keys $K$ vanishes in the whitened metric, yielding $K^\top C^{-1} \delta \approx 0$.

Based on the orthogonality in the whitened space, the correction term is negligible, leading to the approximation:
\begin{equation}
    \label{eq:approx}
    (C + K K^\top)^{-1} \delta \approx C^{-1} \delta.
\end{equation}

\paragraph{Spectral Analysis and The Covariance Trap.}
The fundamental disparity between the standard edit (using $C$) and the isotropic modification (using $I$) stems from the spectral properties of the covariance matrix. As stated in the main text, $C = \mathbb{E}[k k^\top]$ acts as an anisotropic whitening filter. In the high-dimensional feature spaces of Transformers, the eigenspectrum of $C$ is known to be heavy-tailed and ill-conditioned.

Let $\sigma_i$ denote the eigenvalues of $C$. Due to the redundancy in language representations, the trailing eigenvalues corresponding to low-variance directions approach zero ($\sigma_{\min} \to 0$). Consequently, the inverse matrix $C^{-1}$ possesses extremely large eigenvalues $\lambda_i(C^{-1}) = 1/\sigma_i$ in these directions. This spectral structure forms the theoretical basis of the Covariance Trap.

Since the perturbation $\delta$ represents generalized noise or drift between knowledge forms, it is statistically isotropic relative to the principal axes of $C$. Therefore, $\delta$ inevitably contains a non-zero component, denoted as $\delta_{proj}$, lying in the subspace spanned by the eigenvectors of $C^{-1}$ associated with its maximal eigenvalues ($\lambda_{\max}$). The operation $C^{-1} \delta$ drastically scales this component.

\paragraph{Comparative Error Bounds.}
We formally compare the error magnitude $D$ under the two conditions:

\begin{itemize}
    \item \textbf{Isotropic Case ($C=I$):} The operator simplifies to $(I + K K^\top)^{-1}$. Since the eigenvalues of this matrix are strictly bounded within $(0, 1]$, the perturbation is suppressed rather than amplified:
    \begin{equation}
        \| (I + KK^\top)^{-1} \delta \|_2 \le \| \delta \|_2.
    \end{equation}
    
    \item \textbf{Anisotropic Case (Covariance Trap):} Under the approximation $(C + K K^\top)^{-1} \delta \approx C^{-1} \delta$, the error is dictated by the spectrum of $C^{-1}$. The upper bound is determined by the largest eigenvalue of the inverse covariance matrix:
    \begin{equation}
        D_{cov} \approx \| C^{-1} \delta\|_2 \geq \lambda_{\max}(C^{-1}) \| \delta_{proj} \|_2 \gg \|\delta\|_2.
    \end{equation}
\end{itemize}

\noindent \textbf{Conclusion.} The presence of $C$ introduces a large amplification factor $\lambda_{\max}(C^{-1})$ (related to the condition number of $C$) acting on $\delta_{proj}$. This theoretical derivation explains the empirical results in Fig.~\ref{fig:act_dev}, where the standard covariance-based update leads to a significantly larger deviation ($D_{cov} \approx 26.1$) compared to the trap-free isotropic update ($D_{identity} \approx 17.4$).

\section{Locality Preservation in Covariance-Based Methods}
\label{app:cov_locality_analysis}

In this section, we analyze how standard locate-then-edit editing methods (e.g., MEMIT) utilize the covariance matrix $C$ to enforce locality constraints. The goal is to clarify the theoretical role of the term $K^\top C^{-1} k_{old} \approx \mathbf{0}$, which we contrast with our simplified identity-based approach in the main text.

\paragraph{Optimization Objective.}
Standard methods formulate the weight update $\Delta W$ as a constrained optimization problem. They aim to align the new knowledge ($K$) with the target residual ($R$) while minimizing the interference on previously learned knowledge. This general knowledge is modeled as a Gaussian distribution of keys $k \sim \mathcal{N}(0, C)$, where $C = \mathbb{E}[k k^\top]$. The objective is:
\begin{equation}
    \min_{\Delta W} \mathbb{E}_{k \sim C} \| \Delta W k \|_2^2 \quad \text{s.t.} \quad \Delta W K = R.
\end{equation}
The closed-form solution to this problem (derived via the method of Lagrange multipliers) is given by:
\begin{equation}
    \label{eq:cov_update_rule}
    \Delta W = R (K^\top C^{-1} K)^{-1} K^\top C^{-1}.
\end{equation}

\paragraph{Mechanism of Constraint.}
To determine if this update preserves an unrelated existing knowledge represented by a key $k_{old}$, we examine the product $\Delta W k_{old}$:
\begin{align}
    \Delta W k_{old} &= R (K^\top C^{-1} K)^{-1} \underbrace{K^\top C^{-1} k_{old}}_{\text{Interaction Term}}.
\end{align}
For the locality constraint to hold (i.e., $\Delta W k_{old} \approx \mathbf{0}$), the interaction term $K^\top C^{-1} k_{old}$ must vanish.

\paragraph{Interpretation of $K^\top C^{-1} k_{old} \approx \mathbf{0}$.}
This condition represents orthogonality in the \textit{Mahalanobis metric space} (or whitened space).

The matrix $C^{-1}$ effectively acts as a whitening filter. Let $z = C^{-1/2} k$ be the whitened feature vector. In this transformed space, the global correlations are removed, and the feature distribution becomes isotropic. Since $K$ (the specific editing target) and $k_{old}$ (a random unrelated key) represent statistically independent concepts, their corresponding whitened vectors $z_{edit}$ and $z_{old}$ are uncorrelated.

Therefore, the inner product in the whitened space satisfies:
\begin{equation}
    \langle K, k_{old} \rangle_{C^{-1}} = K^\top C^{-1} k_{old} = z_{edit}^\top z_{old} \approx 0.
\end{equation}

\twocolumn
\section{Geometric Investigation Details}
In this section, we provide the detailed experimental setup for the geometric analysis presented in Section 3 and furnish additional empirical evidence using the Llama-3.1-8B-Instruct model to demonstrate the universality of the observed phenomena.
\subsection{Geometric Measurement}
To rigorously quantify the \textbf{generalization collapse} phenomenon, we define and measure two critical geometric indicators: Tolerance Radius ($R$) and Activation Deviation ($D$).
\paragraph{Tolerance Radius ($R$) Measurement.}
The Tolerance Radius $R$ quantifies the robustness of the optimal solution $v^*$ in the value space against perturbations. It represents the maximum radius of a hypersphere within which the model maintains its prediction accuracy. We estimate $R$ via Monte Carlo sampling:
\begin{itemize}
    \item Noise Injection: We sample random noise vectors $\xi$ from a Gaussian distribution $\mathcal{N}(0, I)$ and scale them to varying magnitudes $\rho$.
    \item Validity Check: For a given magnitude $\rho$, we add the noise to the optimized value vector: $v' = v^* + \rho \cdot \frac{\xi}{||\xi||_2}$. We then check if the model still correctly predicts the target object $o^*$ with the probability $P(o^*) \ge \tau $ when given the original input prompt. We set the threshold $\tau = 0.9$. For each sample and each value of $\rho$, we perform 10 trials and ensure a success rate exceeding 90
    \item Estimation: We perform an iterative $\rho$ increasing search to find the maximum radius with a fixed step $\epsilon$ ($1.0$ for Qwen2.5-7B-Instruct and $0.1$ for Llama-3.1-8B-Instruct) over 200 random samples.
\end{itemize}
\paragraph{Activation Deviation ($D$) Measurement.}
The Activation Deviation $D$ measures the displacement of the value vector caused by prompt variations. For a given subject $s$ and relation $r$, let $k_o$ be the key vector activated by the declarative prompt used during editing (e.g., "The CEO of Apple is ..."), and let $\tilde{k}$ be the key vector activated by a natural question or an instructed query (e.g., "Please answer the question without any explanation. Question: Who is the CEO of Apple?"). We calculate the deviation as the $L_2$ norm of the difference between the projected updates as shown in Eq.~\eqref{eq:act_dev}.
\paragraph{Gradient Conflict Score.} To analyze the geometry of the optimization landscape, we compute the cosine similarity between the gradients of different relations ($r_i, r_j$) for the same subject. The Gradient Conflict Score is defined as:
$$Score(g_i, g_j) = 1 - \frac{g_i \cdot g_j}{||g_i||_2 ||g_j||_2},$$
ranging from $0$ to $2$.
A score near 1 indicates that the gradients are orthogonal, implying that the solution subspaces for different relations are perpendicular, which constrains the intersection space.

\subsection{Additional Empirical Evidence on Llama}
\label{app:llama_pilot}
We replicate the geometric analysis on Llama-3.1-8B-Instruct to verify that the pathology $D > R$ is not specific to the Qwen architecture but is a fundamental issue in current editing paradigms. The results on Llama-3.1 strongly corroborate our findings in the main text.

\begin{figure}[t]
    \centering
    \includegraphics[width=0.7\columnwidth]{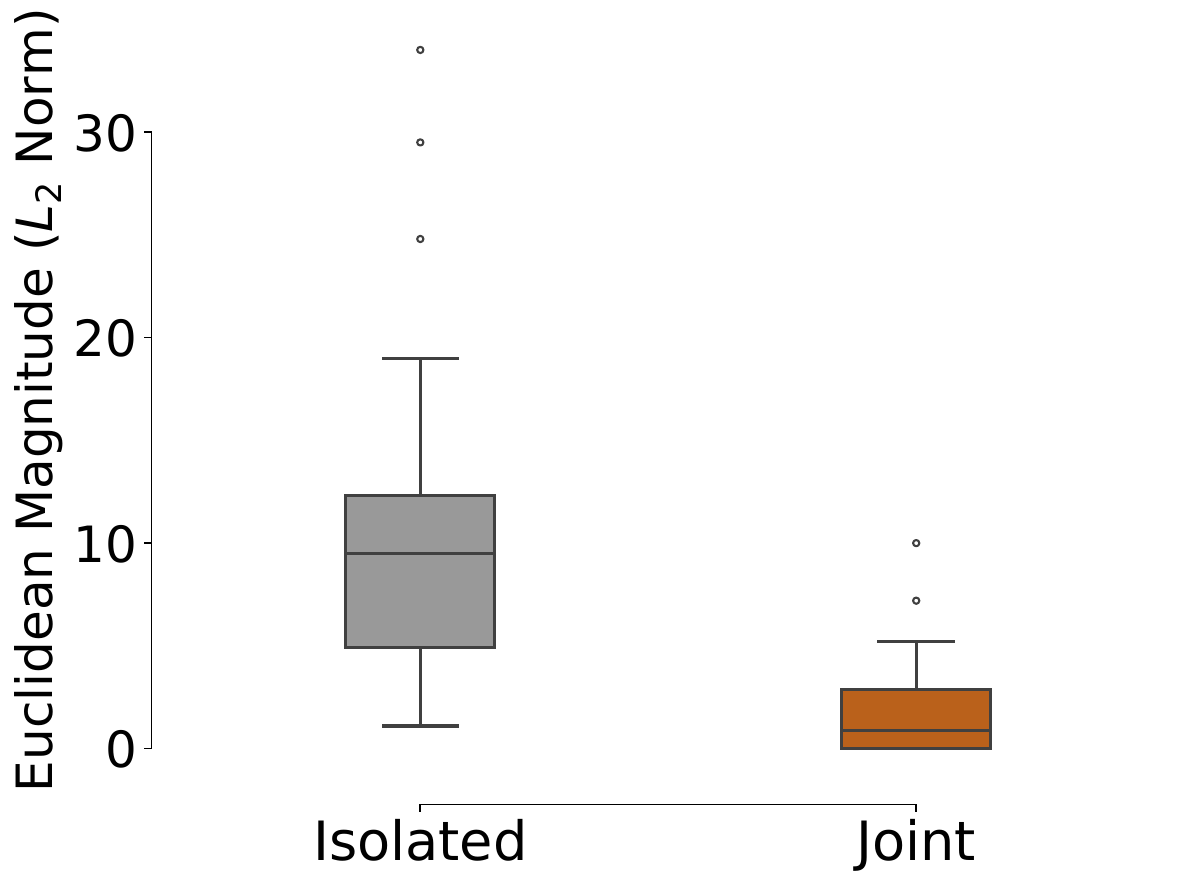}
    \caption{Radius Collapse on Llama-3.1. The Tolerance Radius $R$ shrinks significantly from "Isolated" editing to "Joint" same-subject editing, indicating the formation of Sharp Minima.}
    \label{fig:tol_collapse_llama}
\end{figure}

\paragraph{Collapse of Tolerance Radius ($R \downarrow$).}
We compare the Tolerance Radius $R$ obtained from isolated single-fact editing (Standard MEMIT) versus joint same-subject editing (MEMIT-Merge). As shown in Fig.~\ref{fig:tol_collapse_llama}, Llama-3.1 exhibits a drastic collapse in tolerance. While isolated edits maintain a large solution space (Mean $R \approx 10.2$), joint optimization compresses the radius to a narrow region (Mean $R \approx 1.7$). This confirms that satisfying multiple constraints simultaneously forces the model into a sharp minimum.

The $R$ value for Llama-3.1 is considerably lower than for Qwen2.5. This difference likely stems from distinct token encoding spaces across models. Nevertheless, the core phenomenon remains consistent.

\begin{figure}[t]
    \centering
    \includegraphics[width=0.7\columnwidth]{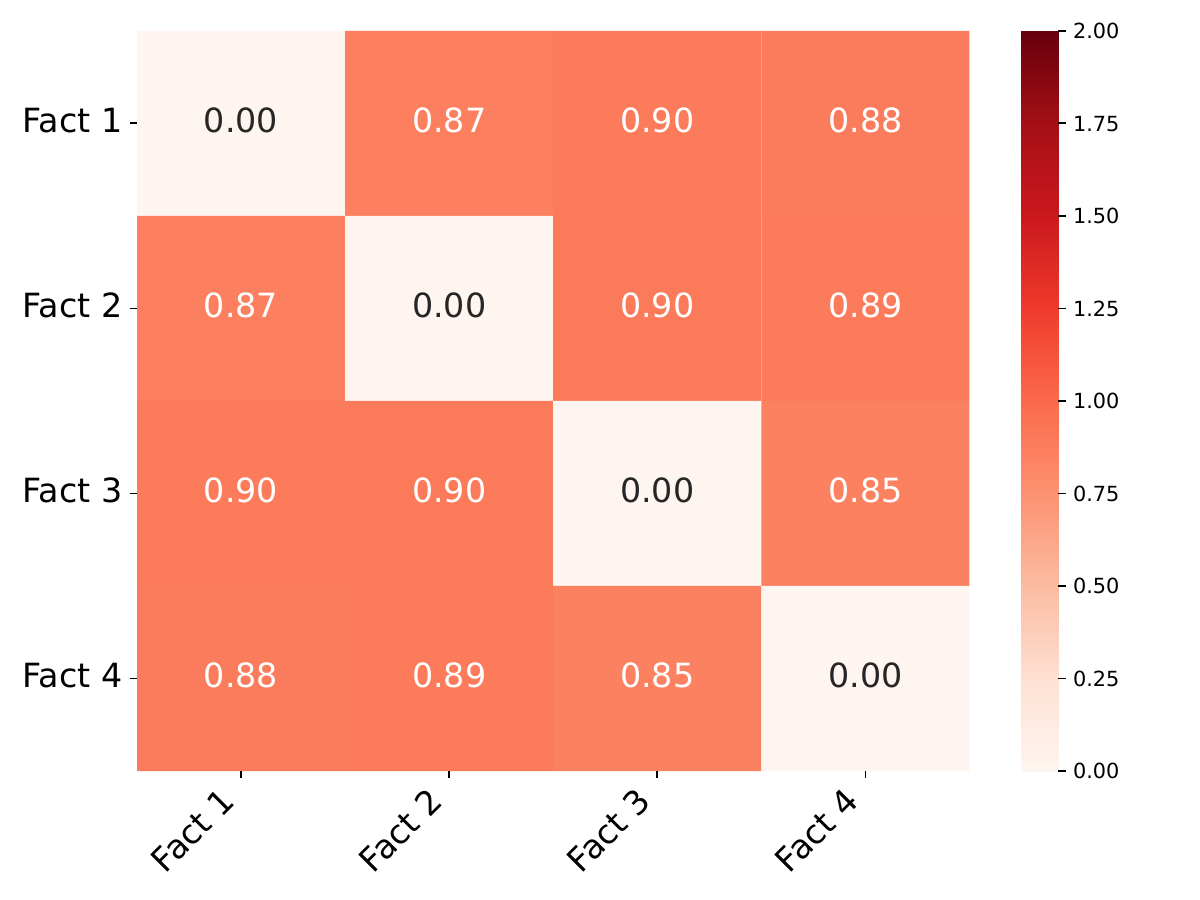}
    \caption{Gradient Orthogonality on Llama-3.1. Pairwise conflict scores between different relations (Facts 1-4) are consistently near 1 ($\approx 0.90$), indicating orthogonal optimization directions.}
    \label{fig:v_gradient_conflict_llama}
\end{figure}

\paragraph{Gradient Orthogonality as the Root Cause.}
To explain the radius collapse, we analyze the gradients of different facts concerning the same subject. Fig.~\ref{fig:v_gradient_conflict_llama} presents the pairwise gradient conflict scores. We observe consistently high conflict scores (ranging from 0.85 to 0.90) between different facts (Fact 1 through Fact 4). This near-orthogonality confirms that the model is trying to find an intersection between mutually perpendicular subspaces, geometrically explaining the sharp minimum observed in Fig.~\ref{fig:tol_collapse_llama}. 

\begin{figure}[t]
    \centering
    \includegraphics[width=0.7\columnwidth]{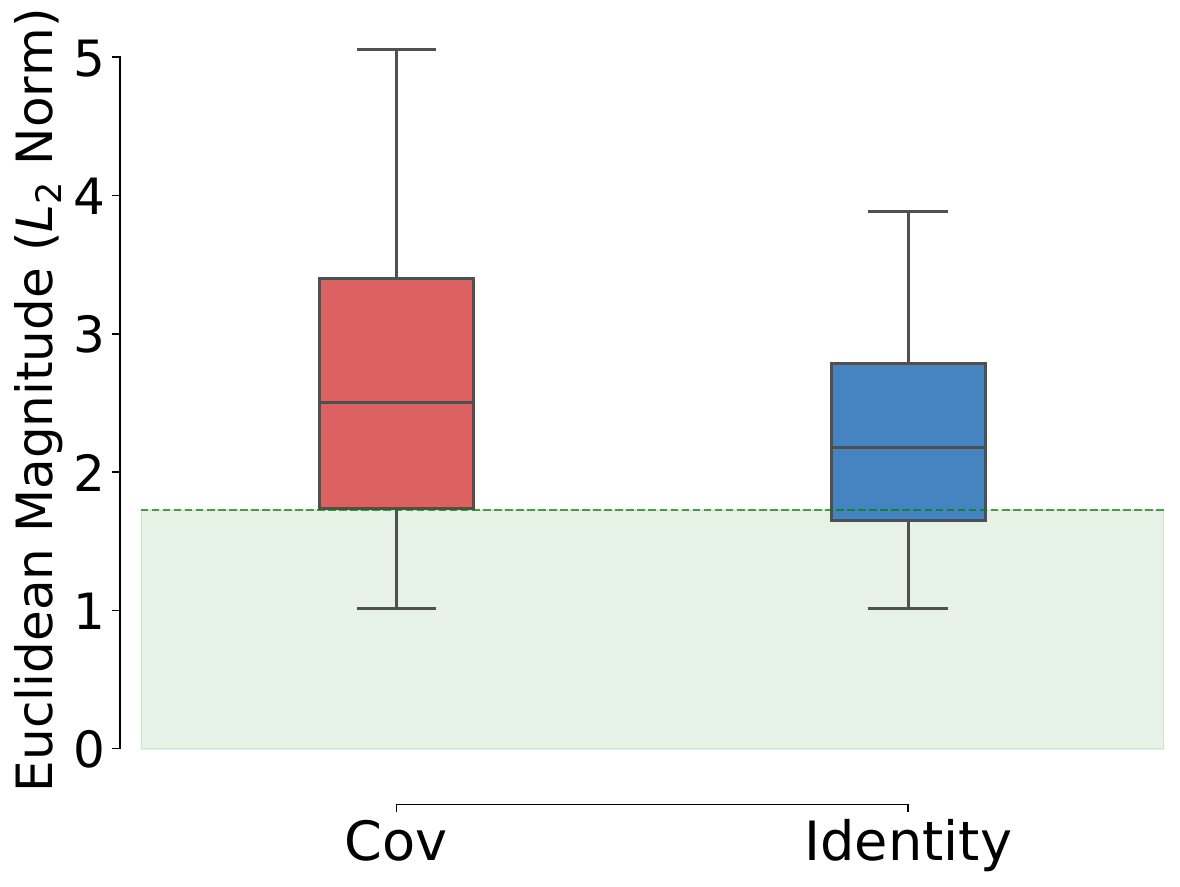}
    \caption{The Covariance Trap on Llama-3.1: Using the Covariance matrix (C) results in higher activation deviation compared to the identity matrix, below the average Tolerance Radius $R$ (green line).}
    \label{fig:act_dev_llama}
\end{figure}

\paragraph{The Covariance Trap ($D\uparrow$).}
We investigate the Activation Deviation $D$ under standard covariance based updates versus our proposed identity-based updates. Fig.~\ref{fig:act_dev_llama} visualizes the Covariance Trap on Llama-3.1. The standard covariance constraint (C) amplifies the deviation significantly ($D \approx 2.6$), pushing it well beyond the collapsed tolerance radius ($R \approx 1.7$). In contrast, replacing the covariance matrix with the identity matrix ("Identity") successfully suppresses the deviation ($D \approx 2.2$), validating our theoretical derivation in Appendix~\ref{app:cov_trap_analysis}.

\begin{figure}[t]
    \centering
    \includegraphics[width=0.9\columnwidth]{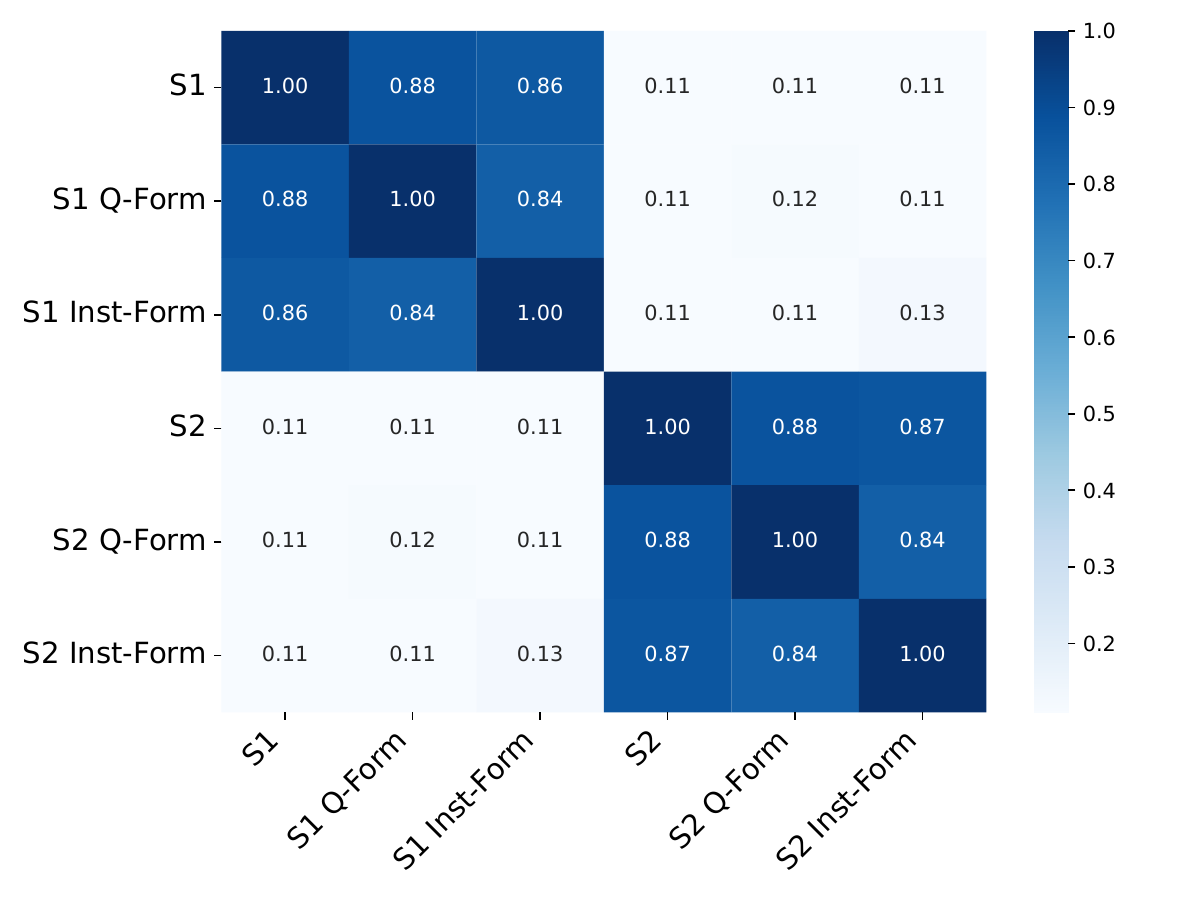}
    \caption{Average key similarity on Llama-3.1. Subject keys are stable across diverse prompt formats (diagonal blocks $\approx 0.84-1.00$) but distinct across subjects (off-diagonal $\approx 0.11$), confirming that retrieval failure is not the primary cause.}
    \label{fig:k_simi_llama}
\end{figure}

\paragraph{Stability of Subject Representation.}
To ensure that the deviation $D$ is caused by the projection metric (Covariance Matrix) rather than unstable key representations, we plot the cosine similarity of key vectors across different prompt formats. Fig.~\ref{fig:k_simi_llama} shows that for the same subject (e.g., S1), the keys extracted from declarative, natural question, and instruction forms are highly similar (Cosine Similarity $> 0.84$). Conversely, keys between different subjects (S1 vs. S2) are orthogonal ($\approx 0.11$). This confirms that the input noise $\delta$ is naturally small, and the high $D$ is indeed an artifact of the anisotropic amplification by $C^{-1}$.

\paragraph{Geometric Restoration via RoSE.}
Finally, we validate the full RoSE framework's ability to restore the safety condition $D \le R$ on Llama-3.1-8B-Instruct. Fig.~\ref{fig:geo_val_llama} in the main text demonstrates that by combining Isotropic Geometric Alignment (to lower $D$) and Hierarchical Knowledge Integration (to increase $R$), RoSE successfully re-aligns the editing geometry on Llama-3.1.

\subsection{The Success on Isolated Fact Editing}
\label{sec:isolated_success}

A natural question arises: why has this instruction-following failure not been reported in previous studies focusing on isolated fact editing (e.g., ROME, MEMIT)? Based on our geometric analysis, we attribute this "success" to the lack of competing constraints, which results in an exceptionally large Tolerance Radius ($R$).

In the scenario of isolated fact editing, the optimization objective is to inject a single key-value pair $(k, v^*)$ for a subject, without the need to balance conflicting gradients from other relations of the same subject. As observed in Fig.~\ref{fig:tol_radius}, this single-objective optimization creates a wide, flat solution basin, yielding a significantly large tolerance radius ($R_{isolated} \approx 92.9/10.2$ for Qwen2.5/Llama-3.1).

This massive $R$ implies that the model's solution space for the edited entity is extremely robust, or arguably, aggressive. Geometrically, the basin of attraction for the target knowledge $v^*$ is so extensive that it dominates the subject's representation space. Consequently, \textbf{even if an instructional prompt induces a considerable Activation Deviation ($D$), the perturbed activation remains safely within the boundaries of this vast tolerance region} ($D \ll R$).

Therefore, the apparent robustness of isolated editing to instructions is not due to a sophisticated alignment of the update direction, but rather due to the model's strong propensity to generate the newly edited content whenever the subject is invoked. The model effectively \textit{overfits} to the single target fact, ensuring successful recall across diverse prompt formats~\cite{zhang2025uncoveringoverfittinglargelanguage, liu2024relationknowsrethinkingrecall}, a luxury that disappears once multiple orthogonal facts must coexist for the same subject.

\begin{figure}[t]
   \centering
   \subcaptionbox{Baseline fails.\label{fig:layer_anal_base}}[.5\linewidth][c]{%
      \includegraphics[width=1.\linewidth]{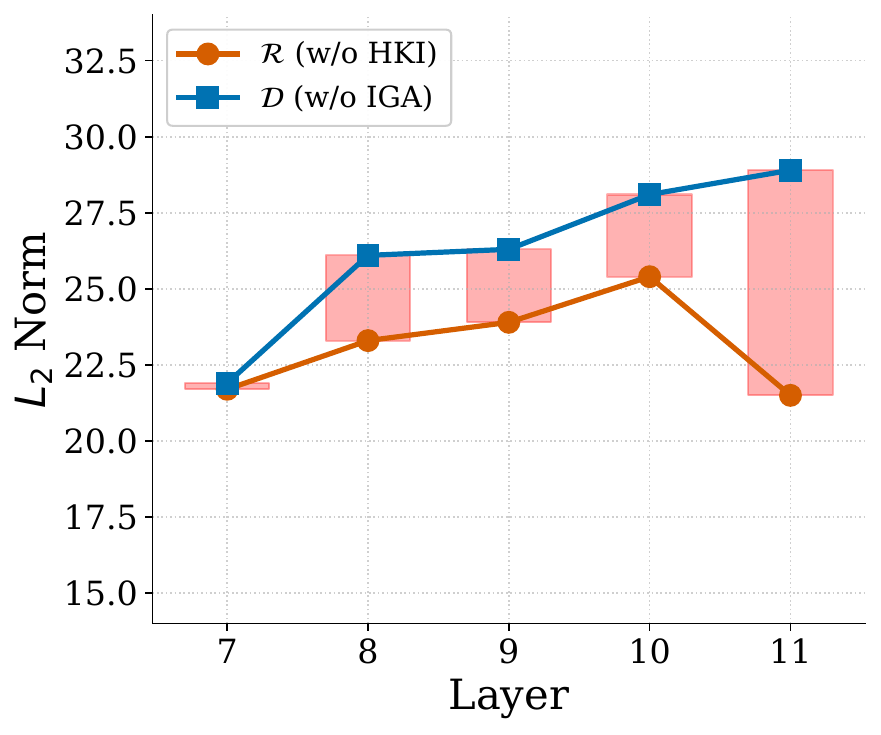}
   }
   \hspace{-0.2cm}
   \subcaptionbox{Ours successes.\label{fig:layer_anal_rose}}[.5\linewidth][c]{%
      \includegraphics[width=1.\linewidth]{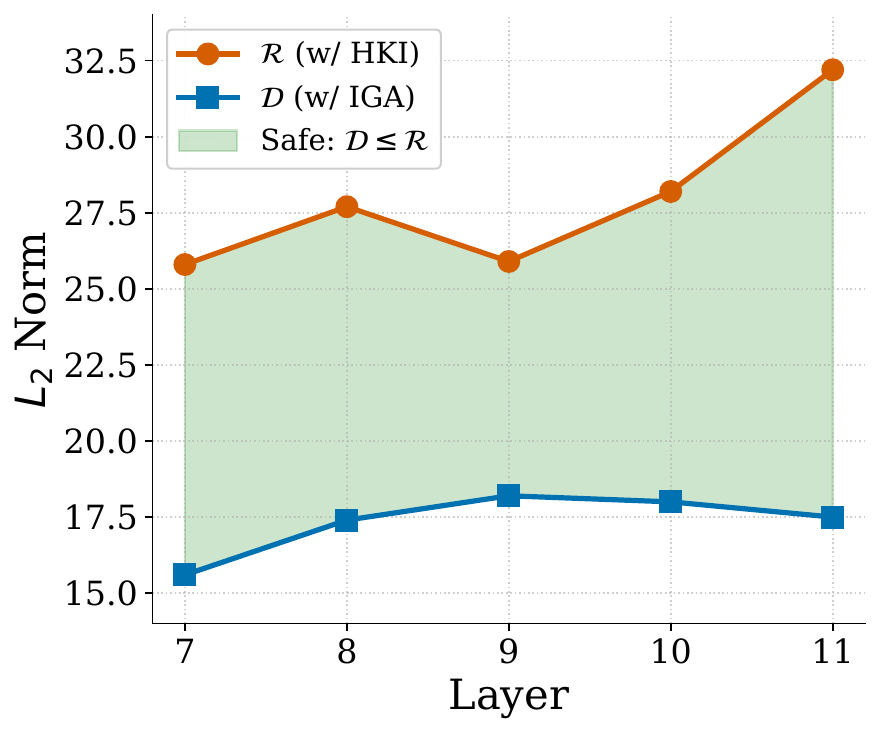}
   }
   \caption{Activation Deviation $D$ and Tolerance Radius $R$ across mid-early layers for the baseline MEMIT-Merge and our RoSE.}
   \label{fig:layer_anal}
\end{figure}

\subsection{Layer Sensitivity Analysis}
In the main analysis, we follow the standard convention of previous studies (e.g., MEMIT, ROME) by observing the geometric metrics at the output of the final edited layer (i.e. Layer 8 for Qwen2.5-7B). This layer is typically chosen because it represents the cumulative effect of the batched updates distributed across the preceding MLP layers.

To demonstrate that our findings, specifically the geometric pathology $D > R$ and the orthogonality of subject keys, are intrinsic properties of the model's activation space rather than artifacts of a specific layer selection, we conduct a sensitivity analysis across the a wider range of mid-early layers (Layers 7 through 11) over 100 samples.

\paragraph{Universality of Geometric Pathology.}
We measure the Tolerance Radius ($R$) and average Activation Deviation ($D$) at the output of each layer.

We observe that the generalization pathology of $D>R$ in joint same-subject editing is consistent across mid-early layers (Fig.~\ref{fig:layer_anal_base}). In contrast, our Identity-based update (IGA) maintains $D$ within the safe zone ($D \le R$) at every layer (Fig.~\ref{fig:layer_anal_rose}). The Activation Deviation $D$ is smaller in earlier layers (e.g., Layer 7). This is attributed to a smaller prompt-derived $\delta$, as subject knowledge has not yet been fully enriched at these stages.

\paragraph{Orthogonality of Keys.}
We also examine the pairwise cosine similarity of key vectors $k$ for different subjects across these layers. The results confirm that the orthogonality hypothesis ($k_{s_1} \perp k_{s_2}$) holds universally, with average cosine similarities remaining near zero throughout the edited layers. This validates that the redundancy of the covariance matrix $C$ is a fundamental geometric property of the Transformer's feature space, independent of layer selection. 

A slight increase in $\delta$ (the similarity of $k$ vectors across linguistic forms of the same subject) is observed with deeper mid-early layers, although the difference remains relatively low throughout. This contributes to the growing trend of Activation Deviation $D$ in deeper layers for the original locate-then-edit update rule (Fig.~\ref{fig:layer_anal_base}).

\begin{figure}[t]
    \centering
    
    \subcaptionbox{Layer 7\label{fig:layer7}}[0.5\linewidth][c]{
        \includegraphics[width=1.\linewidth]{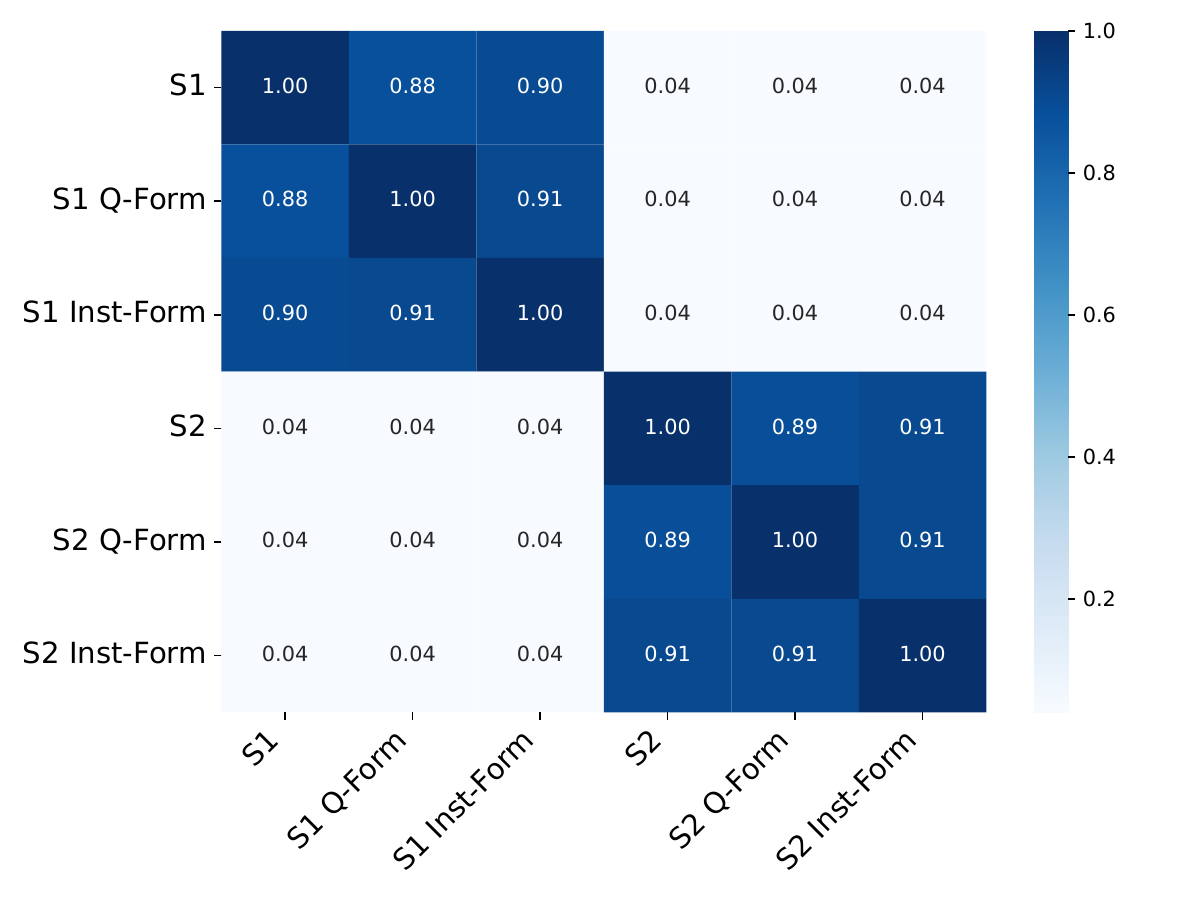}
    }
    \hspace{-0.2cm}
    \subcaptionbox{Layer 8\label{fig:layer8}}[0.5\linewidth][c]{
        \includegraphics[width=1.\linewidth]{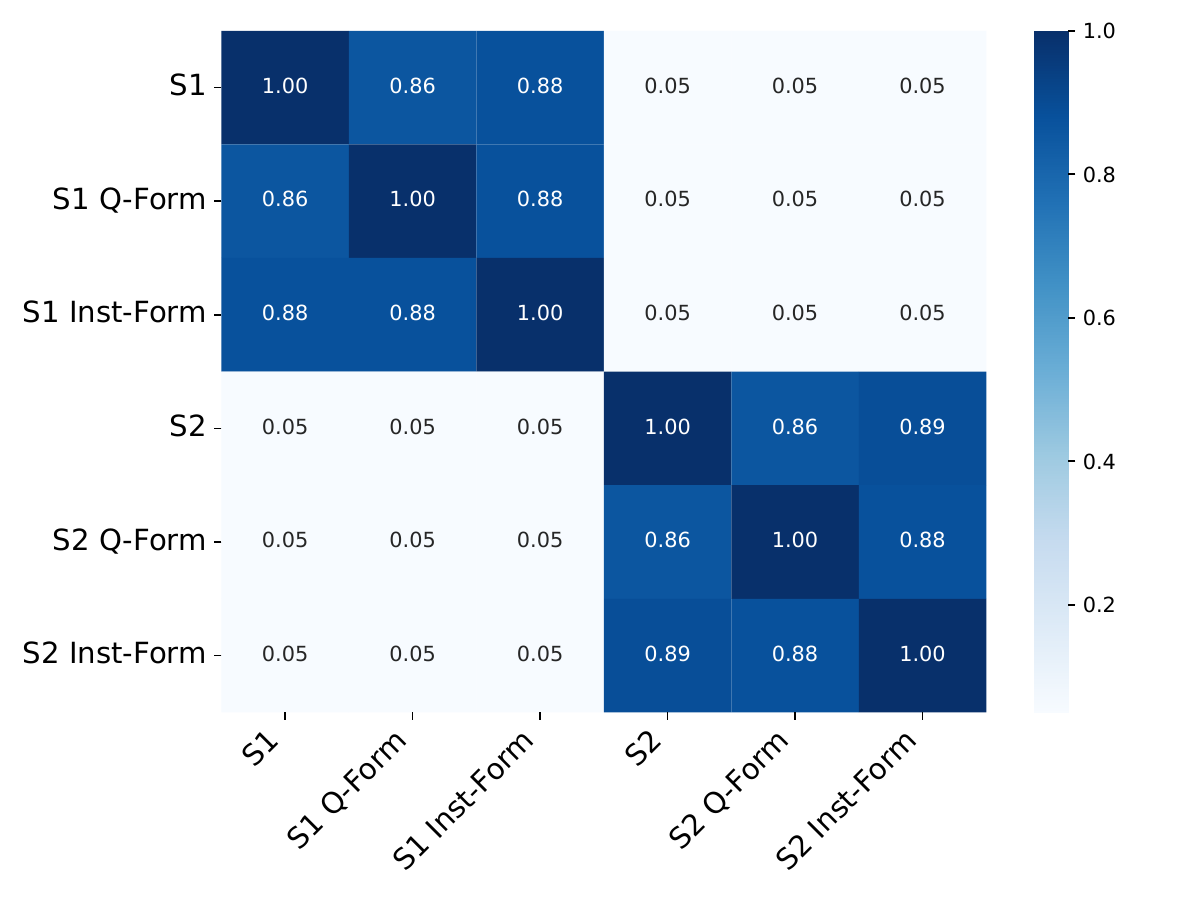}
    }
    
    \subcaptionbox{Layer 9\label{fig:layer9}}[0.5\linewidth][c]{
        \includegraphics[width=1.\linewidth]{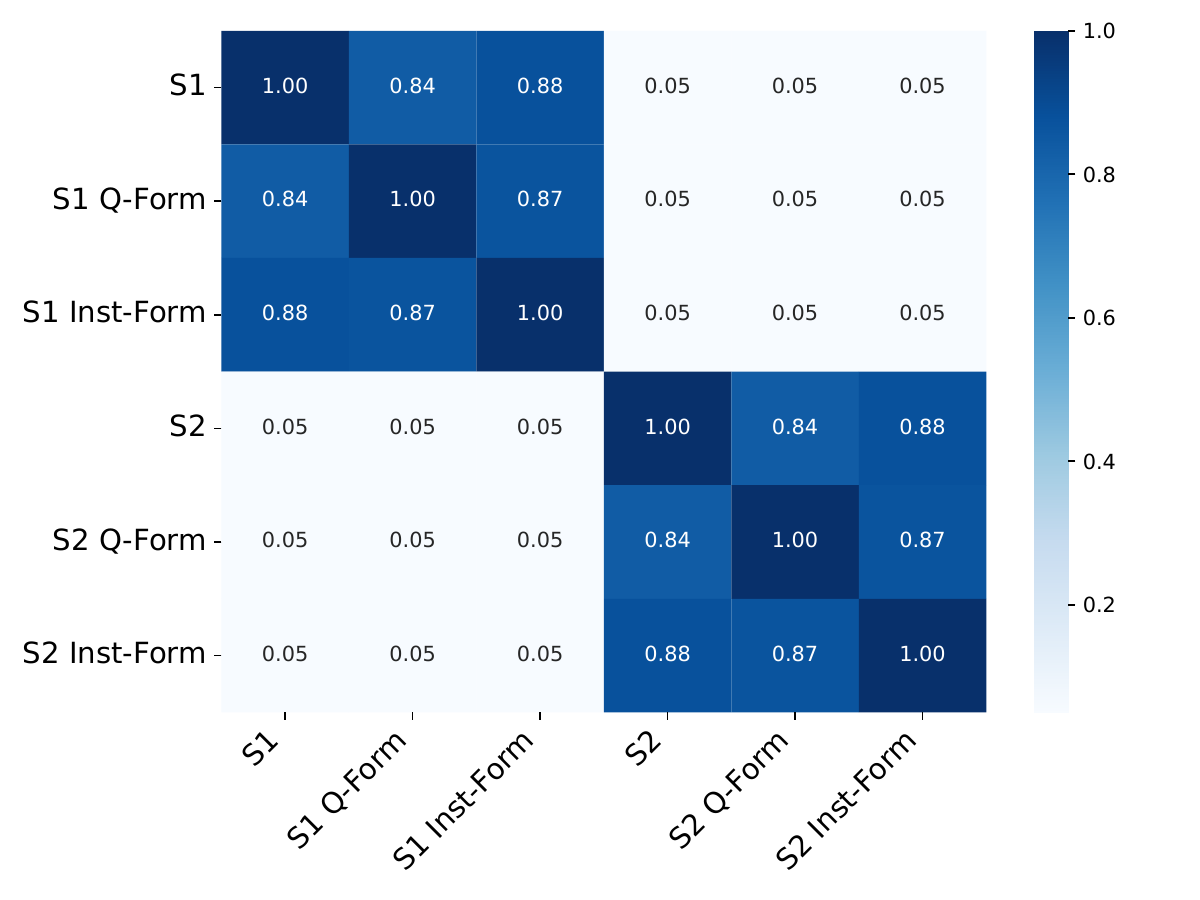}
    }
    \hspace{-0.2cm}
    \subcaptionbox{Layer 10\label{fig:layer10}}[0.5\linewidth][c]{
        \includegraphics[width=1.\linewidth]{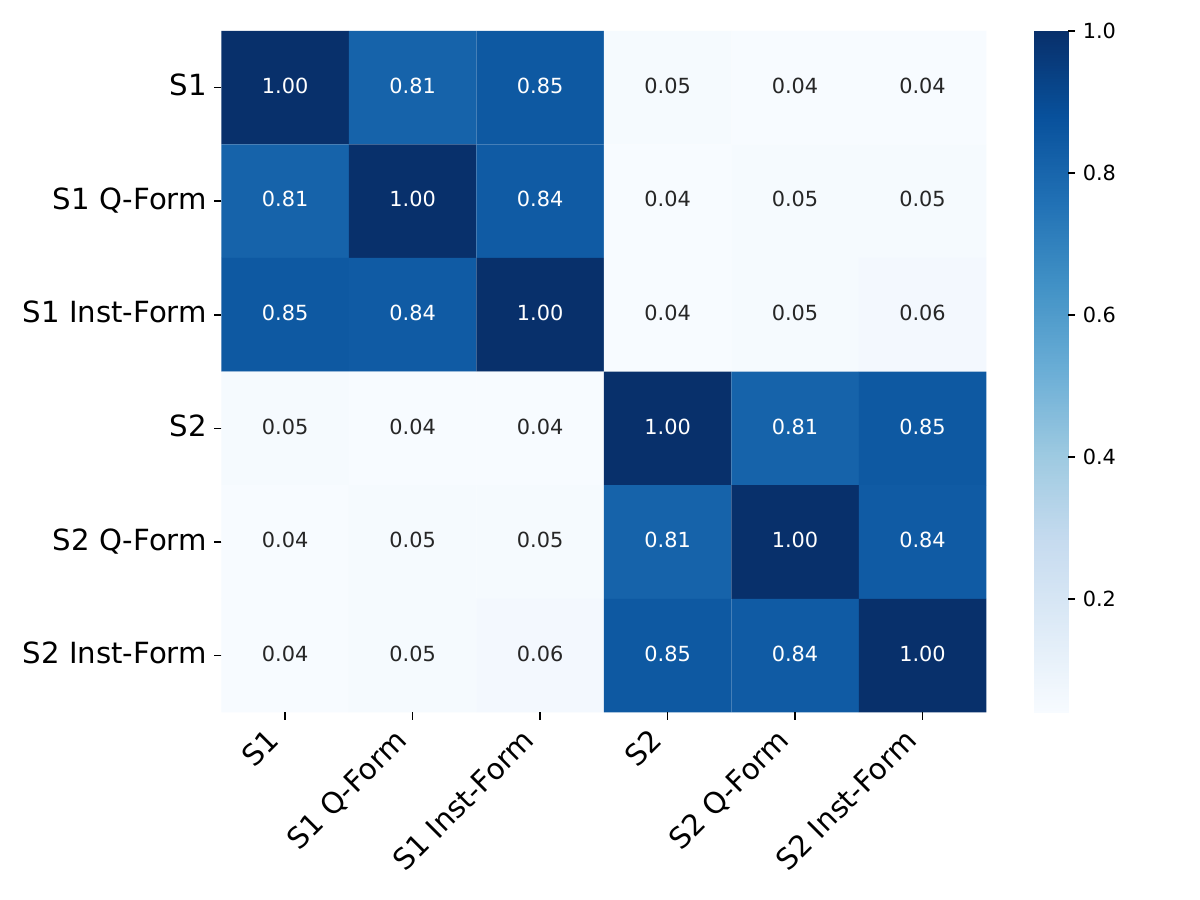}
    }
    
    \caption{Layer sensitivity analysis for $k$ similarity.}
    \label{fig:k_similarity_layers}
\end{figure}

In conclusion, the geometric conflict driving the generalization failure is robust and layer independent, necessitating the systemic solution provided by RoSE.

\subsection{Extended Analysis of Key Orthogonality}
\label{sec:appendix_c_extended}

To assess the universality and boundaries of our geometric findings, we extend our probing of key orthogonality ($k_{s_1} \perp k_{s_2}$) to larger model scales and semantically proximate subjects.

\begin{figure}[t]
   \centering
   \subcaptionbox{Layer 15.}[.5\linewidth][c]{%
      \includegraphics[width=1.\linewidth]{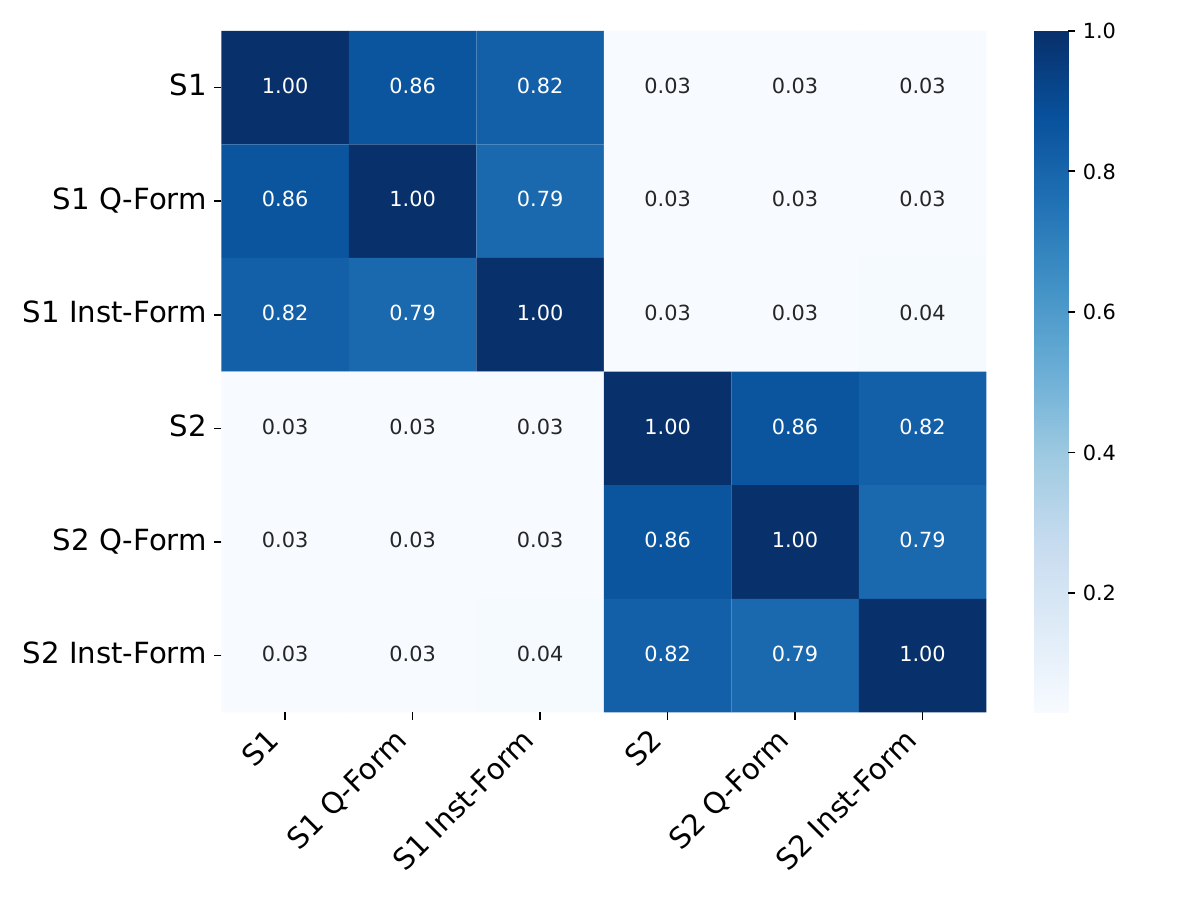}
   }
   \hspace{-0.2cm}
   \subcaptionbox{Layer 17.}[.5\linewidth][c]{%
      \includegraphics[width=1.\linewidth]{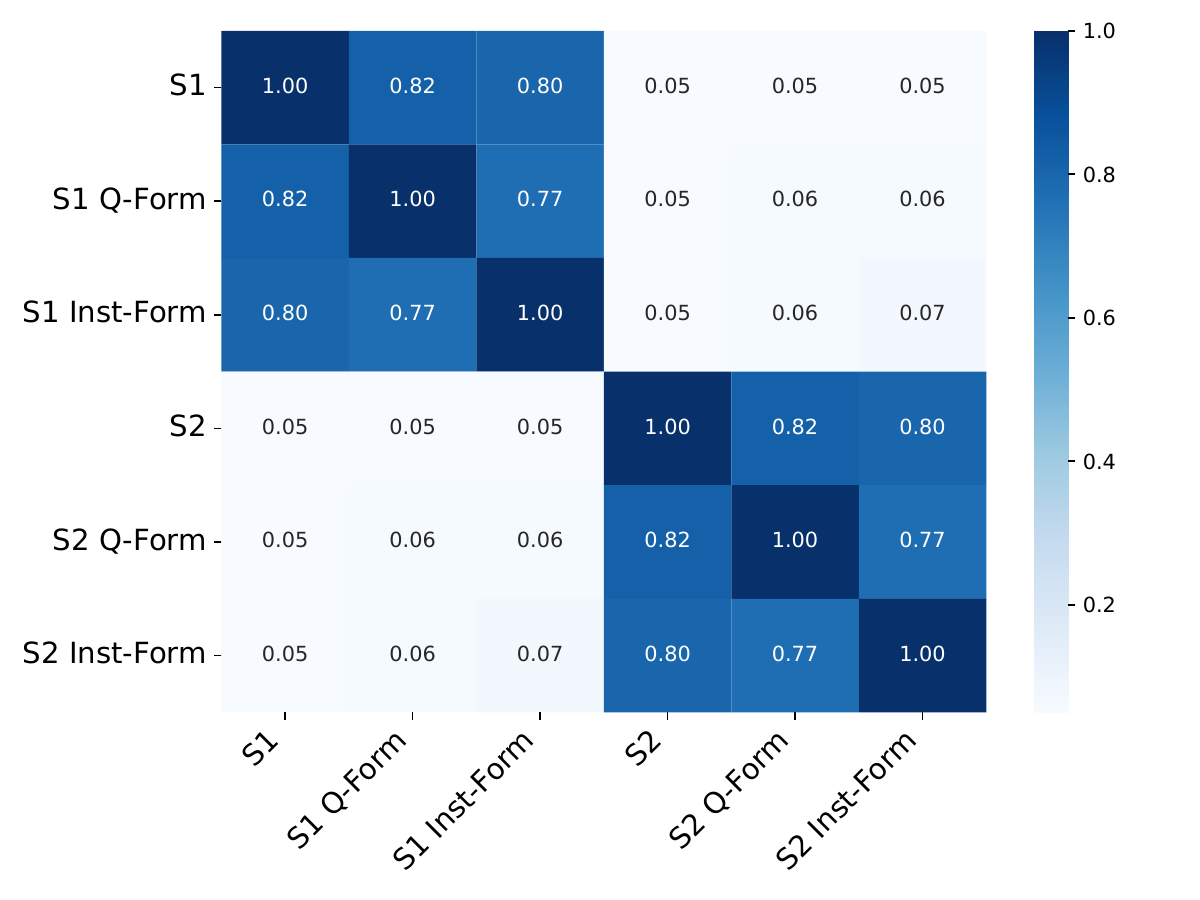}
   }
   \caption{Average $k$ similarity on Qwen2.5-14B.}
   \label{fig:k_simi_14B}
\end{figure}

\paragraph{Robustness Across Model Scales.}
We replicate the key similarity analysis on the larger \textbf{Qwen2.5-14B}~\cite{qwen2025qwen25technicalreport} model (48 layers) to determine if the orthogonality hypothesis holds beyond the 7B/8B parameter class. Also focusing on the mid-early layers (specifically probing Layer 15 and Layer 17) over 200 pairs of samples, we observe that the average cosine similarity between distinct subject keys remains consistently near zero (Fig.~\ref{fig:k_simi_14B}). This empirical evidence confirms that the orthogonality of subject representations is a robust, intrinsic property of the LLMs' activation space that scales with model size.

\begin{figure}[t]
   \centering
   \subcaptionbox{\textit{iPhone 11} vs. \textit{iPhone 16}}[.5\linewidth][c]{%
      \includegraphics[width=1.\linewidth]{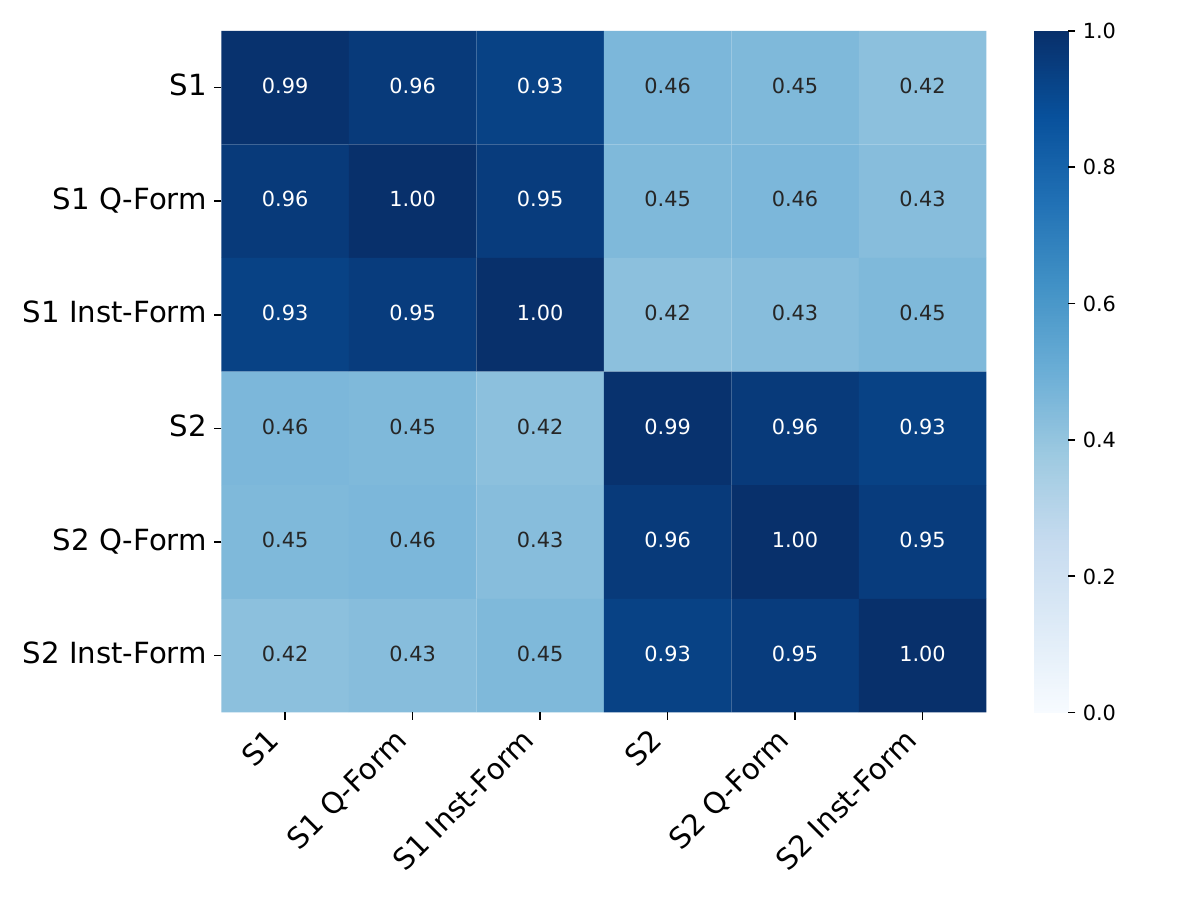}
   }
   \hspace{-0.2cm}
   \subcaptionbox{H.W. Bush vs. W. Bush}[.5\linewidth][c]{%
      \includegraphics[width=1.\linewidth]{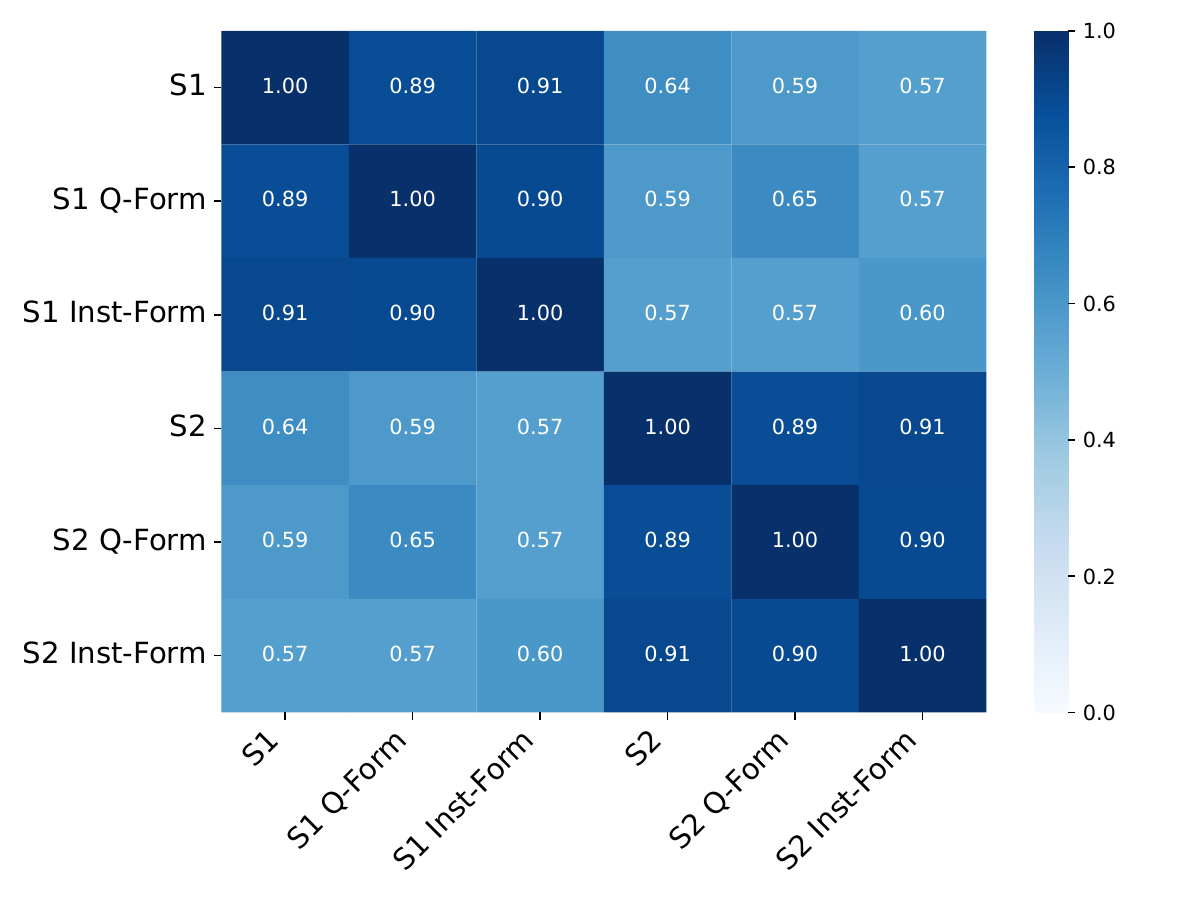}
   }
   \caption{Cases of $k$ similarity for semantically similar subjects on Qwen2.5-7B.}
   \label{fig:k_simi_seman}
\end{figure}

\paragraph{Analysis of Semantically Similar Subjects.}
We further investigate the challenging boundary case involving semantically highly correlated subjects, such as distinct versions of the same product line (e.g., ``\textit{iPhone 11}'' vs. ``\textit{iPhone 16}''). In these extreme scenarios, where token overlap and semantic proximity are significant, we observe that the cosine similarity between keys can rise to around 0.5 (Fig.~\ref{fig:k_simi_seman}).
This high correlation challenges the strict orthogonality assumption and may lead to locality breaches. For instance, when editing the counterfactual ``\textit{iPhone 16 was released by Elon Musk}'', RoSE fails to preserve the distinct knowledge of ``\textit{iPhone 11}'', incorrectly altering its attribute as well.

\textbf{Crucially, however, this limitation is not introduced by our removal of the covariance matrix.} We find that covariance-based baselines, such as MEMIT-Merge, exhibit the identical failure mode in these specific cases (Tab.~\ref{tab:similarity_failure}). This indicates that the standard covariance constraint ($C$) is also insufficient to disentangle such highly entangled representations in the feature space. Consequently, while these edge cases represent a general challenge for the locate-then-edit paradigm, removing $C$ in RoSE does not exacerbate the issue relative to the state-of-the-art, reinforcing that our method gains generalization and efficiency without sacrificing effective locality.

\begin{table}[t]
    \centering
    \small
    \renewcommand{\arraystretch}{1.2}
    \begin{tabular}{p{0.95\linewidth}}
        \toprule
        \multicolumn{1}{c}{\textbf{Boundary Case: High-Similarity Subjects}} \\
        \midrule
        \textbf{Edit Target:} \\
        Subject: \textbf{iPhone 16} \\
        Fact: was released by $\rightarrow$ \textbf{Elon Musk} \\
        \midrule
        \textbf{Locality Check (Unintended Subject):} \\
        \textit{Query: Who released iPhone 11?} \\
        \hdashline
        \textbf{MEMIT-Merge:} \textcolor{red}{iPhone 11 was released by Elon Musk's company, Tesla.} \textbf{\textit{(Fail)}} \\
        \textbf{RoSE (Ours):} \textcolor{red}{iPhone 11 was officially released by Tesla CEO Elon Musk.} \textbf{\textit{(Fail)}} \\
        \bottomrule
    \end{tabular}
    \caption{Identical failure mode on similar subjects.}
    \label{tab:similarity_failure}
\end{table}

\section{Experimental Setup Details}
\label{sec:appendix_d}

\subsection{Dataset Construction \& Statistics}
\label{app:datasets}
\noindent \textbf{S2RKE \& Instructed Query Generation.} 
We utilize the S2RKE benchmark~\cite{duan2025relatedknowledgeperturbationmatters}, which focuses on editing same-subject multi-relation facts. To rigorously evaluate the instruction-following capability, the core contribution of this work, we construct a held-out test set of \textit{instructed queries} distinct from the declarative forms used during editing. Specifically, we employ DeepSeek-R1-Distill-Qwen-14B~\cite{deepseekai2025deepseekr1incentivizingreasoningcapability} as the generator model. We prompt the model to rewrite original declarative facts into natural questions (e.g., rewriting "\textit{The CEO of Apple is Tim Cook}" into "\textit{Who is the current executive running Apple Inc.?}"). Then we transform the natural questions to instructed queries with instruction templates (e.g., "\textit{<|im\_start|>system\textbackslash n Your job is to answer the question. Do not explain.<|im\_end|>\textbackslash n <|im\_start|>user\textbackslash n Query:\textbackslash n Based on your own knowledge, write a short answer for the following question in a few words. Please do not write complete and lengthy sentences.\textbackslash n Question: {}\textbackslash n<|im\_end|>\textbackslash n<|im\_start|>assistant\textbackslash n}"). This procedure ensures that the queries mimic real-world user interactions while maintaining semantic consistency.

\noindent \textbf{LoCoMo-Edit Dataset.} 
To evaluate performance in a more practical, conversational setting, we construct the LoCoMo-Edit dataset derived from the LoCoMo benchmark~\cite{maharana2024evaluatinglongtermconversationalmemory}. 

We select 7 long-term conversations from the dataset, chosen for their rich character dynamics and temporal information.
From these conversations, we select 893 Question-Answer (QA) pairs categorized into three types: 503 \textit{Single-Hop} (direct simple facts), 183 \textit{Multi-Hop} (complex questions), and 207 \textit{Temporal} (requires digital understanding).
For each QA pair, we extract the corresponding ground-truth knowledge in the declarative form using DeepSeek-R1-Distill-Qwen-14B to serve as the source for knowledge editing. This setup allows us to update the model's internal belief state using the declarative fact and test its generalization via diverse conversational QA pairs.

\subsection{Evaluation Metrics Implementation}
\label{app:metrics}
\noindent \textbf{S2RKE Metrics.} 
Following standard protocols, we report Efficacy, Paraphrase, and Locality. Efficacy and Paraphrase measure the success rate of the model in generating the target object $o^*$ given the exact edit prompt and paraphrased prompts (but in the same linguistic form), respectively. Locality is evaluated on unrelated facts to ensure background knowledge preservation.

\noindent \textbf{LoCoMo-Edit Metrics.} 
For the LoCoMo-Edit benchmark, we also adopt the batch editing setting to simulate realistic same-subject editing scenarios. In each experimental run, we edit a batch of 5 to 10 declarative facts simultaneously. We then evaluate the model's ability to answer the corresponding Single-Hop, Multi-Hop, and Temporal questions. The performance is quantified using the \textbf{F1 Exact Match Score}, which measures the token-level overlap between the model's generated answer and the ground truth answer after normalization (removing punctuation and articles).

\subsection{Hyperparameters \& Implementation}

\noindent \textbf{RoSE Implementation.} 
A key advantage of our proposed RoSE framework is that it is essentially hyperparameter-free regarding optimization tuning. Unlike gradient-based meta-learning methods that require careful tuning of learning rates or regularization weights, RoSE relies on geometric alignment. For Hierarchical Knowledge Integration (HKI), the linguistic templates are manually constructed and the only configuration parameter involves the construction of the gradient tree. We set the number of linguistic templates per relation to 8 to compute the Robust Centroid at the leaf level. For Isotropic Geometric Alignment (IGA), as derived in Section 4.1, we replace the covariance matrix $C$ with the identity matrix $I$, requiring no additional hyperparameter.

To ensure a fair comparison and demonstrate that our performance gains stem from geometric alignment rather than hyperparameter tuning, we strictly adhere to the default configurations of MEMIT-Merge~\cite{dong2025memitmergeaddressingmemitskeyvalue} for our RoSE implementation. 
Specifically, for both Llama-3.1-8B-Instruct and Qwen2.5-7B-Instruct, we target the distinct MLP layers \{4, 5, 6, 7, 8\}. The optimization for the target value vector $v^*$ is conducted for 35 steps with a learning rate of 0.5 and a weight decay of 0.5. The KL divergence constraint factor is set to 0.0625 to preserve general model behavior. For the covariance statistics, we utilize 100,000 samples from the Wikipedia dataset, consistent with standard practices.

\noindent \textbf{Baselines \& Models.} 
We compare our method against ROME, MEMIT, AlphaEdit, and MEMIT-Merge. All baseline methods are implemented using their \textbf{official default configurations} and hyperparameter settings to ensure a fair comparison. The experiments are conducted on Qwen2.5-7B-Instruct and Llama-3.1-8B-Instruct.

All experiments are conducted 5 times on a single NVIDIA A100 80GB GPU. Average scores and standard deviations are recorded.

The codes, models, and datasets used in this work are released under an open-source license to facilitate reproducibility and encourage broader community adoption. Our usage is consistent with their respective licenses and intended research-only scope.
Our newly created artifacts will also be released under an open-source license with the explicit condition that they are to be used for research purposes only, in compliance with the access conditions of their original sources.

\section{Original S2RKE Results}
\label{sec:appendix_e}

In the main text, our evaluation primarily focuses on the newly identified failure mode regarding \textit{instruction-following} queries. However, a robust same-subject knowledge editing method must also retain high performance on standard, declarative benchmarks. To verify this, we evaluate RoSE on the original test set of the S2RKE benchmark~\cite{duan2025relatedknowledgeperturbationmatters} on Qwen2.5-7B-Instruct, which consists of standard declarative prompts and paraphrases in the declarative form.

\begin{figure}[t]
    \centering
    \includegraphics[width=1.0\columnwidth]{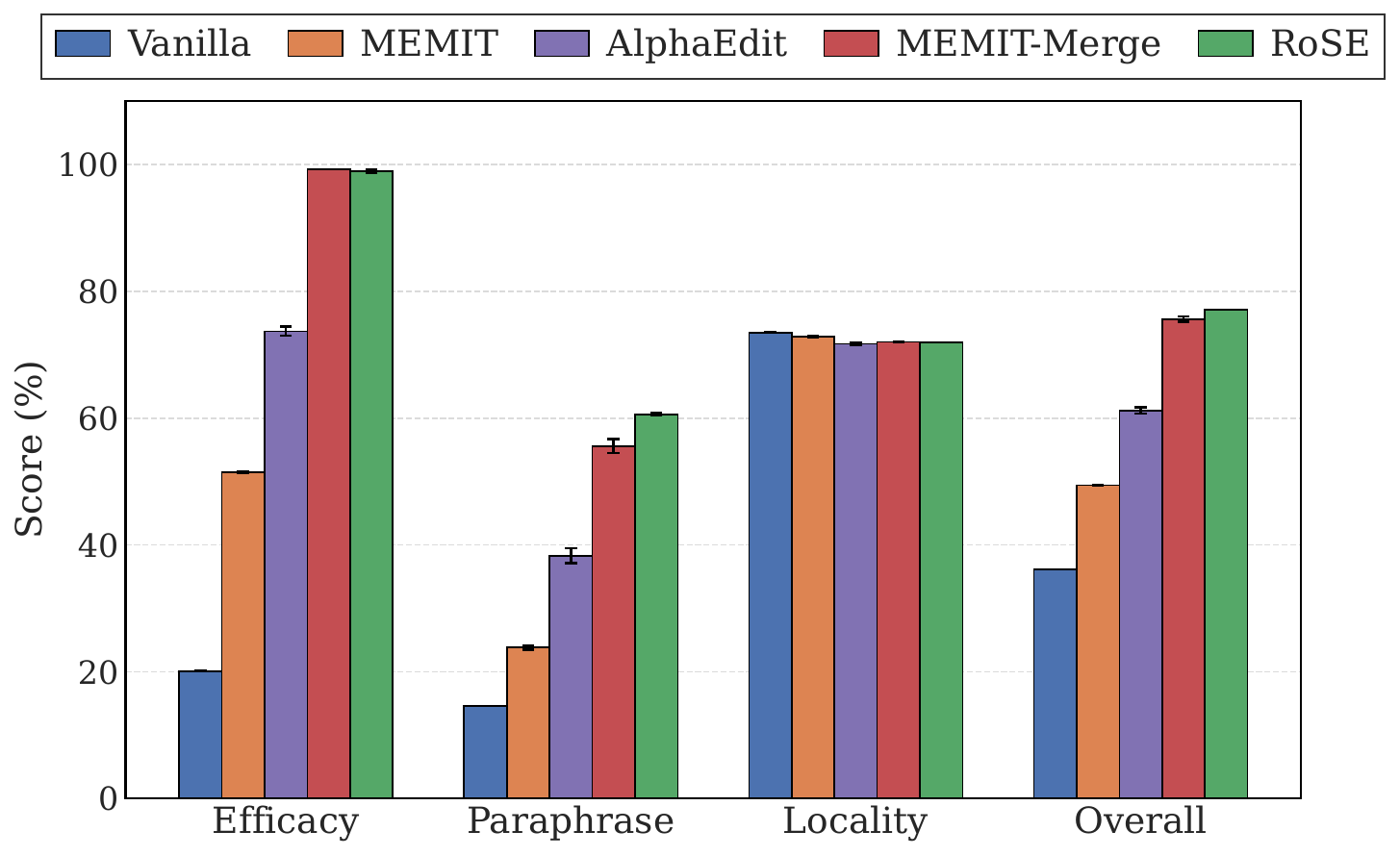}
    \caption{Performance on S2RKE with initial declarative prompts. RoSE enhances Paraphrase accuracy while preserving Locality, indicating better generalization capability. }
    \label{fig:original_s2rke_results}
\end{figure}

As illustrated in Fig.~\ref{fig:original_s2rke_results}, RoSE demonstrates superior or comparable performance across all metrics against strong baselines. In terms of editing success, RoSE achieves an Efficacy score of 98.9, which is statistically tied with the state-of-the-art baseline MEMIT-Merge (99.2). This result empirically confirms that replacing the covariance matrix with the identity matrix via Isotropic Geometric Alignment (IGA) does not compromise the model's ability to inject exact declarative facts.

More importantly, RoSE exhibits a distinct advantage in generalization. It outperforms MEMIT-Merge by a margin of 5.0 points in the Paraphrase metric (60.6 vs. 55.6). This improvement suggests that the geometric regularity enforced by IGA, combined with the smoothing effect of Hierarchical Knowledge Integration (HKI), enhances the model's robustness not only to complex instructions but also to standard linguistic variations. Furthermore, despite the removal of the covariance constraint, which is traditionally considered essential for preventing catastrophic forgetting, RoSE maintains a Locality score of 71.9. This is on par with covariance-based methods like MEMIT (72.8) and MEMIT-Merge (72.0), supporting our theoretical finding that the intrinsic orthogonality of subject keys is sufficient to safeguard unrelated knowledge. In summary, RoSE improves the generalization capability without incurring any penalty on standard same-subject editing tasks.

\begin{figure}[t]
    \centering
    \includegraphics[width=0.9\columnwidth]{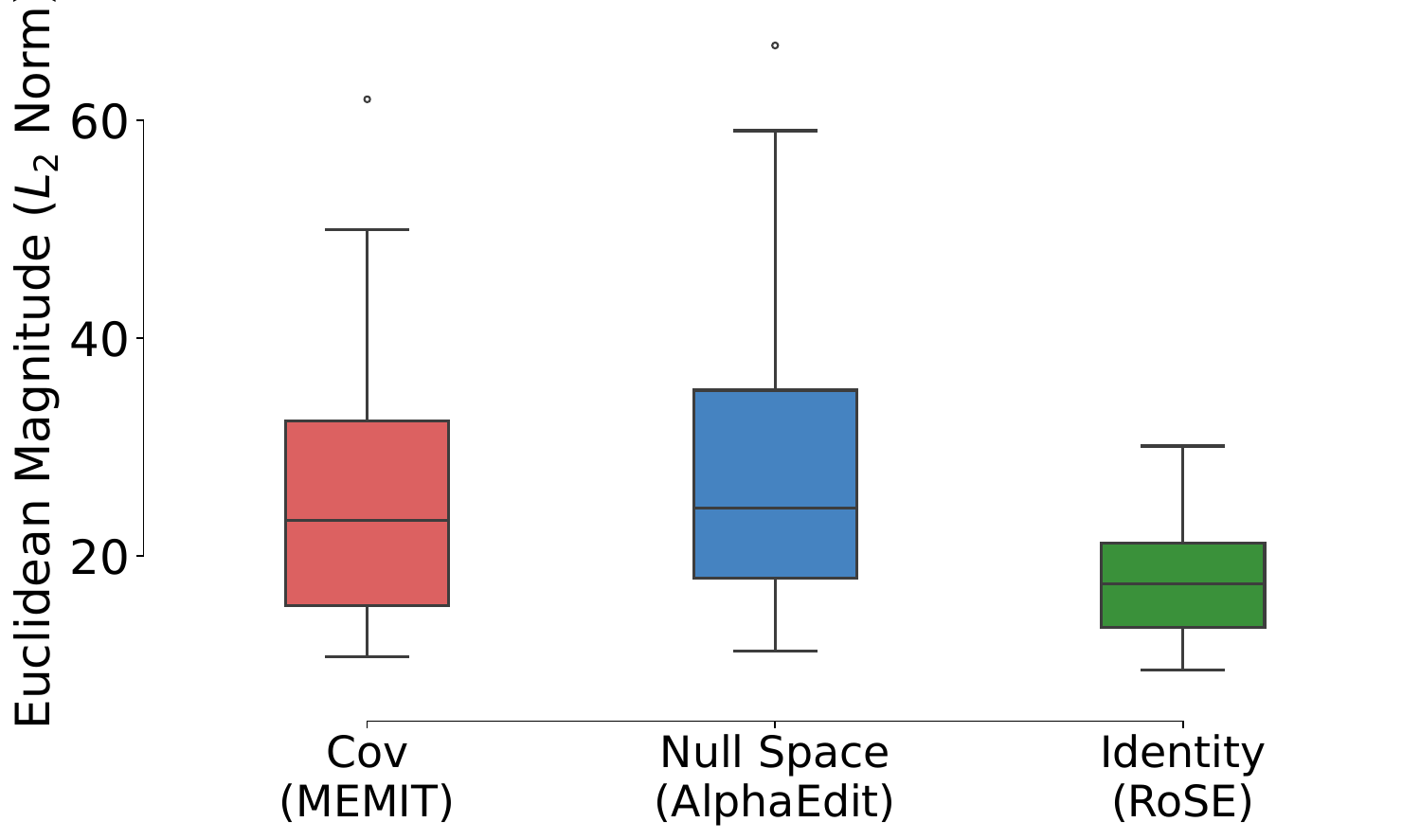}
    \caption{Activation Deviation $D$ of MEMIT (the covariance matrix), AlphaEdit (the null-space projection derived from the covariance matrix) and RoSE (the identity matrix) on Qwen2.5. }
    \label{fig:alpha_dev}
\end{figure}

\section{Discussion on Other Variants of the Covariance Matrix}
\label{sec:appendix_f}

While our main analysis critiques the direct usage of the inverse covariance matrix $C^{-1}$ in methods like MEMIT, recent advancements such as AlphaEdit~\cite{fang2024alphaeditnullspaceconstrainedknowledge} propose alternative constraints, specifically \textbf{Null-Space Projection}. In this section, we provide a concise mathematical explanation to demonstrate that such variants are fundamentally equivalent to the standard covariance constraint in terms of geometric pathology and are similarly redundant.

\noindent \textbf{The Equivalence of Constraints.} 
AlphaEdit and similar approaches construct a projection matrix $P$ derived from the preservation knowledge statistics (typically the covariance matrix $C$). The goal is to confine the weight update $\Delta W$ within the null space of existing knowledge to ensure locality:
\begin{equation}
    \Delta W_{Alpha} = \Delta W \cdot P, \quad \text{where } P \approx U U^\top
\end{equation}

Here, $U$ represents the filtered eigenvectors of $C$ (corresponding to near-zero eigenvalues), which span the subspace of existing knowledge to be preserved. Mathematically, $P$ acts as a spectral truncation, effectively removing the smallest eigenvalues of $C$ that cause the explicit explosion in $C^{-1}$ (as detailed in Appendix~\ref{app:cov_trap_analysis}).

However, while this seems to mitigate the numerical explosion caused by the large eigenvalues of $C^{-1}$, it fails to resolve the fundamental geometric pathology of anisotropy. 
The projection matrix $P$ is intrinsically tied to the corpus statistics, creating a \textit{hard anisotropic cutoff}. It creates a subspace that, while bounded, retains the irregular geometric structure of the original covariance distribution. Compared to the isotropic Identity matrix with spherical regularization on all directions, $P$ selectively preserves specific dimensions based on covariance statistics.
Consequently, components of the prompt-induced perturbation $\delta$ that align with the valid subspace of $P$ are not suppressed. They effectively bypass the filter and still contribute to the Activation Deviation (Fig.~\ref{fig:alpha_dev}).

\noindent\textbf{Redundancy via Orthogonality.} 
More importantly, the imposition of the projection $P$ is theoretically redundant. The premise of using $P$ is to satisfy the locality constraint:
\begin{equation}
    \Delta W k_{old} = 0 \iff P k_{old} = 0
\end{equation}
However, as empirically validated in Fig.~4 of the main text and Fig.~\ref{fig:k_simi_llama}, the key vector of the new subject $k_{new}$ is naturally orthogonal to the subspace of existing distinct subjects $k_{old}$:
\begin{equation}
    k_{new}^\top k_{old} \approx 0 \implies k_{new} \in \text{NullSpace}(C)
\end{equation}
Since the update direction is driven by $k_{new}$, it naturally lies within the null space of $C$ without explicit projection. Therefore, applying $P$ performs an identity operation ($P k_{new} \approx k_{new}$) in the ideal case, but actually can introduce numerical noise and geometric distortion in practice.

\noindent \textbf{Implications for Continuous Editing.}
In conclusion, variants relying on null-space projection matrices derived from $C$ are essentially different manifestations of the same geometric misconception. They impose unnecessary constraints that offer marginal gains in locality while compromising the robust generalization required for instruction following. Crucially, this finding provides insights beyond batch editing to the realm of continuous knowledge editing (lifelong learning).  \textbf{Since sequential editing methods often rely on recursively updating covariance statistics or projection matrices to mitigate forgetting, our identification of the redundant Covariance Matrix suggests that such practices may inadvertently accumulate anisotropic distortion over time, thereby progressively degrading the model's plasticity and robustness to future instructions.}

\section{Computational Overhead Analysis}
\label{sec:appendix_g}

In the Limitations section, we acknowledge that the Hierarchical Knowledge Integration (HKI) component theoretically introduces additional computational steps due to the aggregation of gradients from multiple linguistic forms. To quantify the actual net impact of our dual-strategy framework on system resources, we conduct a comparative analysis of computational overhead. We measure the efficiency in a standard batch editing scenario consistent with the S2RKE benchmark statistics, specifically performing edits on Qwen2.5-7B-Instruct with a batch size of 4 on a single NVIDIA A100 GPU. We record both the average wall-clock time required to complete one batch update and the peak GPU memory allocated during the process.

\begin{figure}[t]
   \centering
   \subcaptionbox{Wall-Clock Time (s).\label{fig:time}}[.5\linewidth][c]{%
      \includegraphics[width=1.\linewidth]{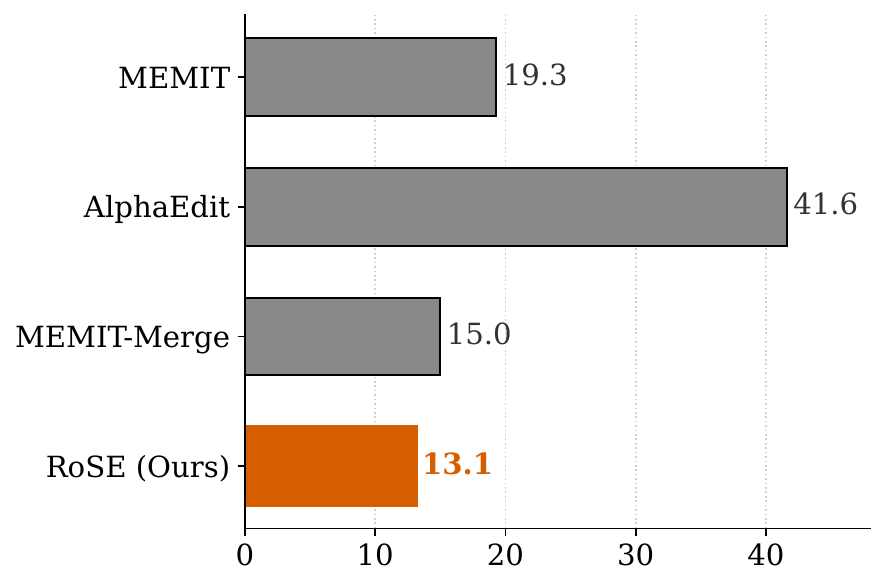}
   }
   \hspace{-0.2cm}
   \subcaptionbox{Peak GPU Memory (GB).\label{fig:max_gpu}}[.5\linewidth][c]{%
      \includegraphics[width=1.\linewidth]{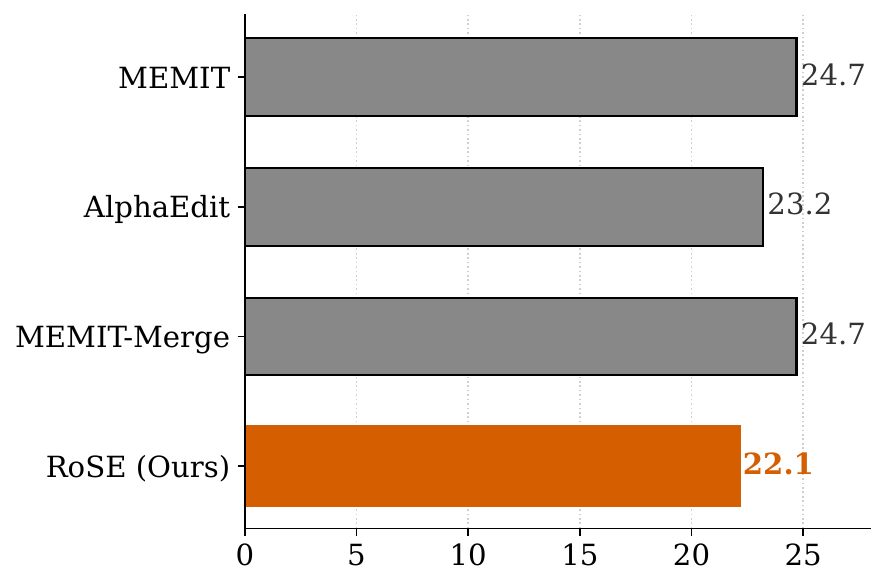}
   }
   \caption{Computational Overhead Comparison. RoSE achieves the lowest latency and memory consumption, validating that the removal of covariance operations (IGA) outweighs the cost of HKI.}
   \label{fig:comput_overhead}
\end{figure}


The comparative results, summarized in Fig.~\ref{fig:comput_overhead}, demonstrate that RoSE achieves the lowest computational footprint across both time and memory metrics among all compared methods, dispelling concerns regarding the overhead of the HKI component. In terms of time efficiency, RoSE requires only 13.1 seconds per edit batch, making it approximately 12.7\% faster than the state-of-the-art MEMIT-Merge (15.0s) and significantly faster than AlphaEdit (41.6s). Regarding memory consumption, RoSE achieves the lowest peak usage at 22.1 GB, saving roughly 2.6 GB of VRAM compared to standard covariance-based methods like MEMIT and MEMIT-Merge, which consume 24.7 GB.

This superior efficiency can be attributed to a favorable architectural trade-off where the savings from Isotropic Geometric Alignment (IGA) significantly outweigh the costs of HKI. While HKI indeed requires processing multiple input prompts to compute the robust centroid, this added computational cost is marginal in the context of batch processing. \textbf{In contrast, IGA removes a significant computational bottleneck inherent in baselines: the management of the covariance matrix ($C$)}. Standard methods and projection variants (like AlphaEdit) are obligated to load massive, high-dimensional covariance statistics into memory and perform expensive matrix operations, such as inversion or Singular Value Decomposition (SVD). By replacing these resource-intensive steps with a simple identity operation, \textbf{RoSE effectively neutralizes the overhead of hierarchical integration, resulting in a system that is not only more robust but also more computationally efficient}.
\begin{figure}[t]
    \centering
    \includegraphics[width=0.8\columnwidth]{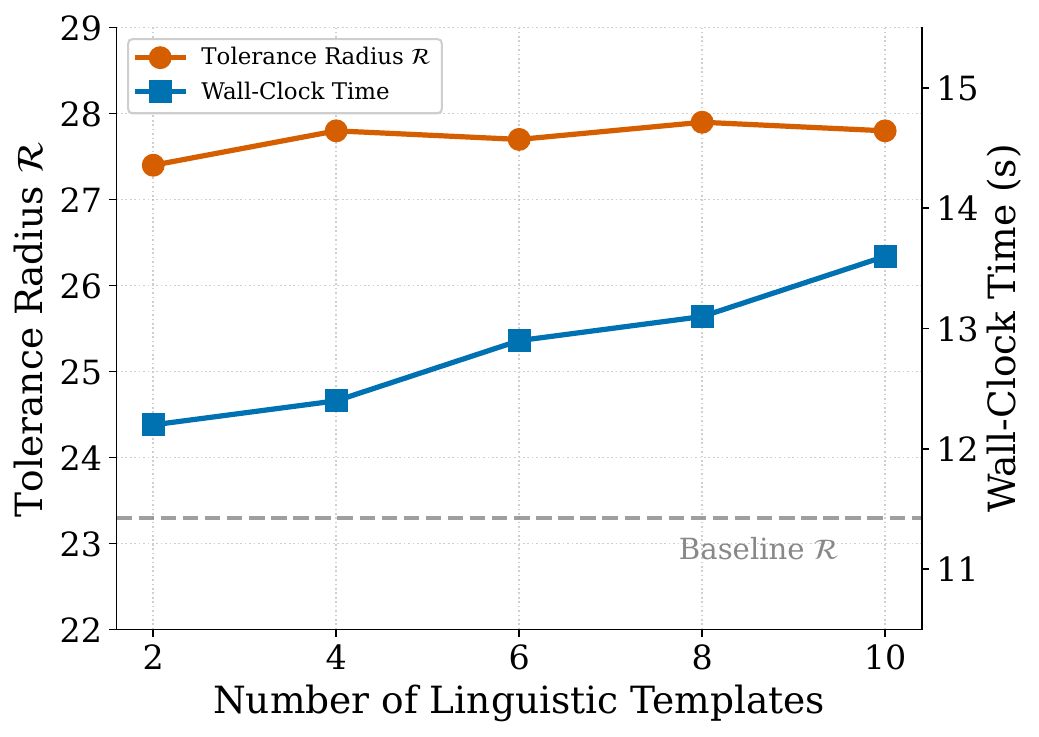}
    \caption{Sensitivity analysis of HKI linguistic templates number $N$ on Qwen2.5-7B-Instruct.}
    \label{fig:hki_sensitivity}
\end{figure}

\section{HKI Sensitivity Analysis}
We investigate the sensitivity of HKI to the number of linguistic templates used for centroid estimation. As illustrated in Fig.~\ref{fig:hki_sensitivity}, the Tolerance Radius $\mathcal{R}$ exhibits remarkable stability, maintaining values consistently above 27 across template counts ranging from 2 to 10. Crucially, even minimal integration ($N=2$) yields a significant improvement over the baseline, indicating that the geometric gain stems primarily from the hierarchical aggregation structure rather than extensive sampling. 

In terms of efficiency, the computational overhead of additional templates is marginal. Increasing the template count from 2 to 8 adds less than 1 second to the wall-clock time. Furthermore, even at our default setting ($N=8$), the total editing time ($\approx 13.1$s) remains strictly lower than the MEMIT-Merge baseline (15.0s), confirming that RoSE enhances robustness without compromising efficiency.

\section{Scalability to Larger Models: 14B Evaluation}
\label{sec:14b_evaluation}

We conduct a comprehensive validation on the S2RKE benchmark using the larger Qwen2.5-14B-Instruct model. The new results, presented in Table \ref{tab:14b_eval}, confirm that RoSE maintains its effectiveness as model parameters scale up, significantly outperforming the strongest baseline (MEMIT-Merge) in both Efficacy (83.2\% vs. 75.4\%) and Paraphrase (62.1\% vs. 51.6\%) while preserving Locality (64.8\%).

These experiments confirm that RoSE continues to significantly enhance instruction-following capabilities at the 14B scale. This finding reinforces the validity of the identity-based update, as evidenced by the highly stable Locality scores (64.8\%) which remain on par with covariance-based baselines. More importantly, we have provided a key similarity analysis for different layers and for the 14B model in Appendix C.4 and Appendix C.5. The results demonstrate that Subject Keys for distinct subject entities maintain high orthogonality in larger scales and a wide range of mid-early layers, strengthening the generality of our claim.

\begin{table*}[htbp]
\centering
\begin{tabular}{lcccc}
\hline \toprule
\textbf{Model} & \textbf{Efficacy} $\uparrow$ & \textbf{Paraphrase} $\uparrow$& \textbf{Locality} $\uparrow$& \textbf{Overall}$\uparrow$ \\
\midrule
Vanilla     & 26.5 ($\pm$0.0) & 36.9 ($\pm$0.0) & 65.0 ($\pm$0.0) & 42.8 ($\pm$0.0) \\
MEMIT       & 48.5 ($\pm$0.3) & 42.0 ($\pm$0.2) & 65.0 ($\pm$0.0) & 51.8 ($\pm$0.1) \\
AlphaEdit   & 59.5 ($\pm$0.4) & 46.7 ($\pm$0.2) & 65.0 ($\pm$0.1) & 57.1 ($\pm$0.2) \\
MEMIT-Merge & 75.4 ($\pm$0.2) & 51.6 ($\pm$0.1) & 64.9 ($\pm$0.1) & 64.0 ($\pm$0.1) \\
RoSE (Ours) & \textbf{83.2} ($\pm$0.3) & \textbf{62.1} ($\pm$0.1) & 64.8 ($\pm$0.0) & \textbf{69.9} ($\pm$0.0) \\
\bottomrule \hline
\end{tabular}
\caption{Experimental instruction-following QA results on S2RKE using the Qwen2.5-14B-Instruct model. RoSE successfully scales and maintains its optimal performance at a larger parameter size. Standard deviations are shown in parentheses.}
\label{tab:14b_eval}
\end{table*}

\section{Pseudocode of RoSE}
\label{sec:rose_pseudocode}

To facilitate reproducibility, we present the complete pseudocode for Robust Same-subject Editing (RoSE) in Algorithm \ref{alg:rose}. This includes both Hierarchical Knowledge Integration (HKI) to expand the tolerance radius and Isotropic Geometric Alignment (IGA) to suppress activation deviation.

\begin{algorithm}[htbp]
\caption{Robust Same-subject Editing (RoSE)}
\label{alg:rose}
\small 
\linespread{1.15}\selectfont
\textbf{Require:} Subject $s$, target relations $\mathcal{R} = \{r_1, \dots, r_m\}$, \\
\hspace*{1.35cm} target objects $\mathcal{O} = \{o_{r_1}, \dots, o_{r_m}\}$, weights $W_0^l$ \\
\textbf{Require:} Diverse linguistic templates $\mathcal{P}$ \\
\textbf{Ensure:} Updated weights $W_{RoSE}^l$

\begin{algorithmic}[1]
\State \textbf{Initialize:} Target value vector $v^*$

\Statex \makebox[0.88\linewidth][l]{$\triangleright$ \textbf{Phase 1: Hierarchical Knowledge Integration}}
\For{each relation $r_i \in \mathcal{R}$} 
    \Statex \hfill \makebox[0.88\linewidth][l]{$\triangleright$ \textit{Leaf-Level: Robust Centroid Estimation}}
    \State Construct prompt set $\mathcal{P}_{r_i}$ using templates $\mathcal{P}$
    \State Initialize aggregated gradient $G_{r_i} \leftarrow 0$
    \For{each prompt $p \in \mathcal{P}_{r_i}$}
        \State $g_p \leftarrow \nabla_{v^*} \mathcal{L}(f_\theta(p), o_{r_i})$
        \State $G_{r_i} \leftarrow G_{r_i} + g_p$
    \EndFor
    \State Robust Centroid: $g'_{r_i} \leftarrow \frac{1}{|\mathcal{P}_{r_i}|} G_{r_i}$
\EndFor

\Statex \hfill \makebox[0.88\linewidth][l]{$\triangleright$ \textit{Root-Level: Intersection Expansion}}
\State Optimize $v^*$ jointly using centroids $\{g'_{r_1}, \dots, g'_{r_m}\}$
\Statex

\Statex \makebox[0.88\linewidth][l]{$\triangleright$ \textbf{Phase 2: Isotropic Geometric Alignment}}
\State Extract batched keys $K$ for subject $s$ across prompts
\State Compute residual vector: $R \leftarrow V^* - W_0^l K$
\State Apply Tikhonov regularization via Identity matrix $I$
\State Compute isotropic update: 
\Statex \hspace{1em} $\Delta W_{IGA} \leftarrow R K^\top (I + K K^\top)^{-1}$
\Statex
\State \textbf{Return} $W_{RoSE}^l \leftarrow W_0^l + \Delta W_{IGA}$
\end{algorithmic}
\end{algorithm}

\section{Stability Under Sequential Edits}
\label{sec:sequential_edits}

We perform continuous editing experiments on both Qwen2.5-7B-Instruct and Llama-3.1-8B-Instruct using S2RKE to conduct an initial investigation into the stability of knowledge accumulation under sequential edits, which better reflects real-world deployment scenarios. 

\begin{figure}[htbp]
    \centering
    
    \begin{subfigure}{0.24\textwidth}
        \centering
        \includegraphics[width=\linewidth]{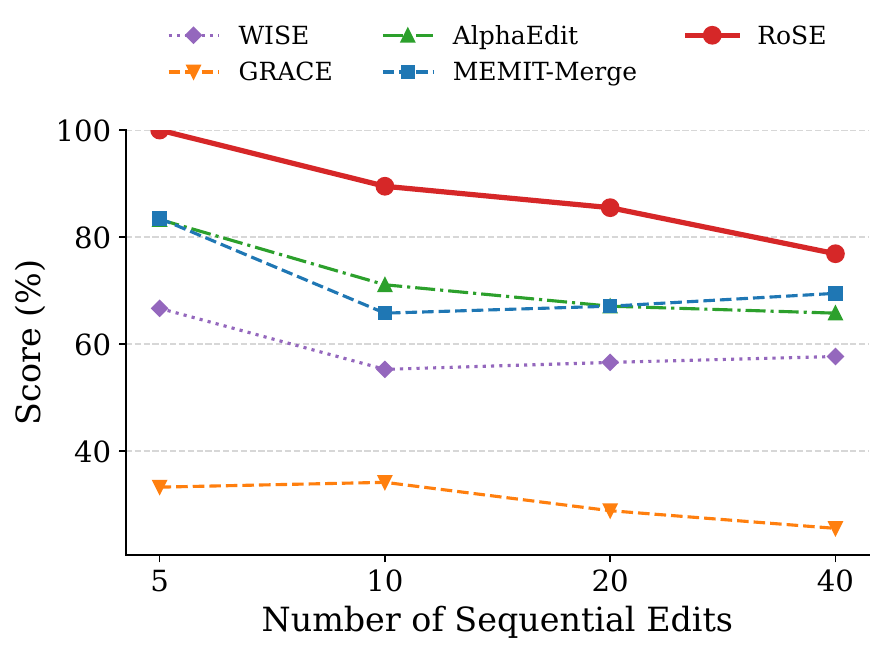}
        \caption{Qwen2.5: Efficacy}
    \end{subfigure}
    \hspace{-0.2cm}
    \begin{subfigure}{0.24\textwidth}
        \centering
        \includegraphics[width=\linewidth]{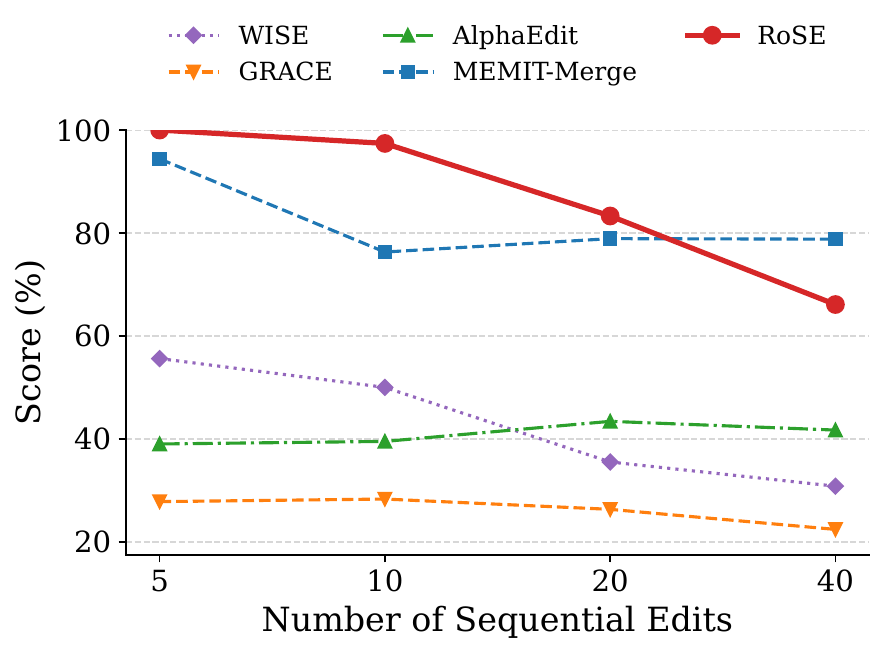}
        \caption{Llama-3.1: Efficacy}
    \end{subfigure}
    
    \begin{subfigure}{0.24\textwidth}
        \centering
        \includegraphics[width=\linewidth]{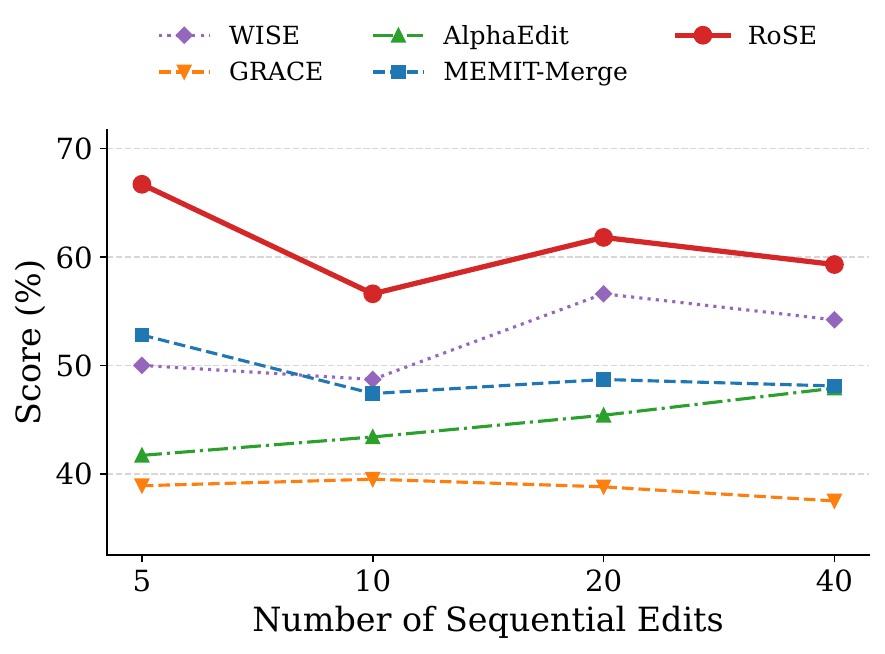}
        \caption{Qwen2.5: Paraphrase}
    \end{subfigure}
    \hspace{-0.2cm}
    \begin{subfigure}{0.24\textwidth}
        \centering
        \includegraphics[width=\linewidth]{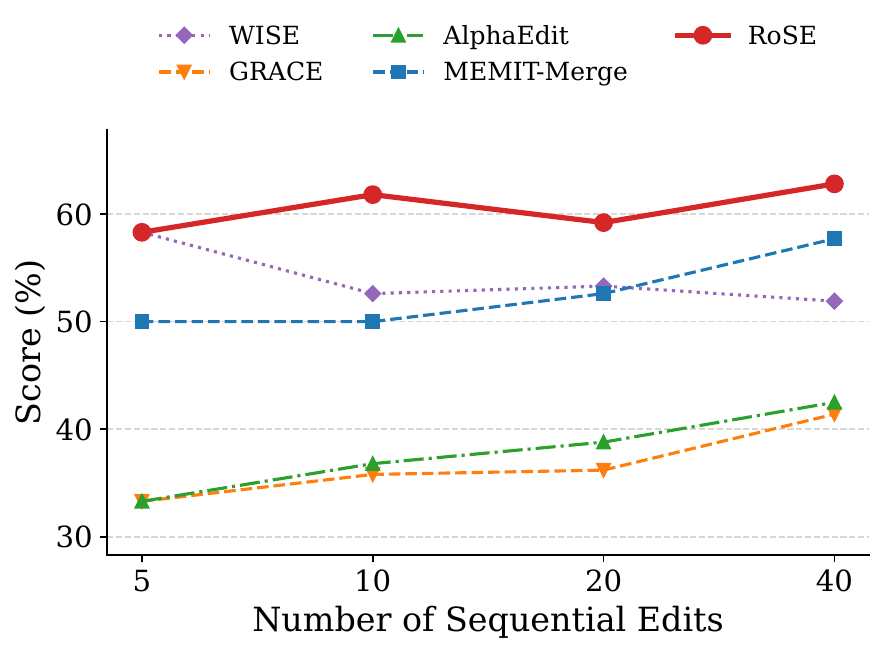}
        \caption{Llama-3.1: Paraphrase}
    \end{subfigure}
    
    \begin{subfigure}{0.24\textwidth}
        \centering
        \includegraphics[width=\linewidth]{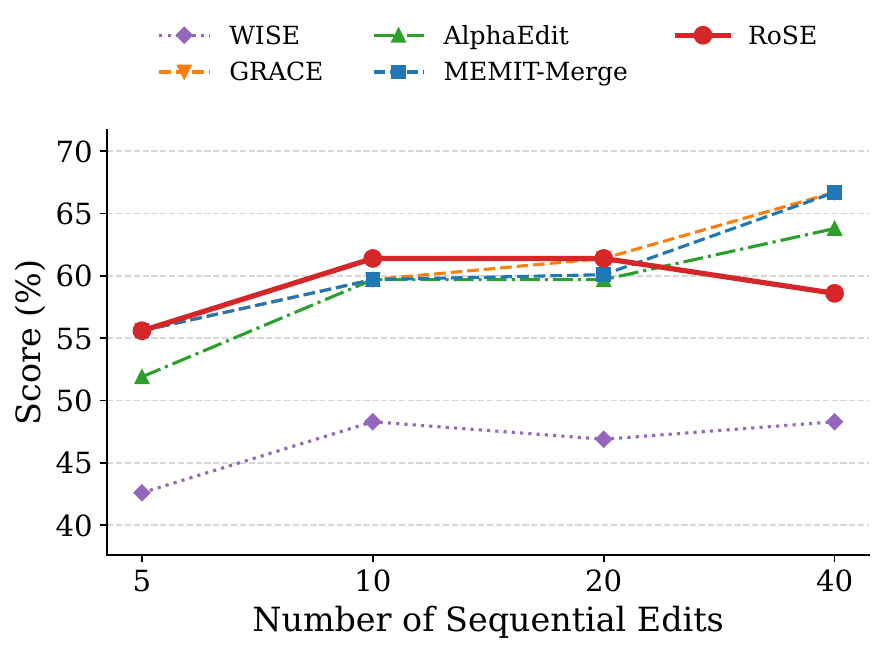}
        \caption{Qwen2.5: Locality}
    \end{subfigure}
    \hspace{-0.2cm}
    \begin{subfigure}{0.24\textwidth}
        \centering
        \includegraphics[width=\linewidth]{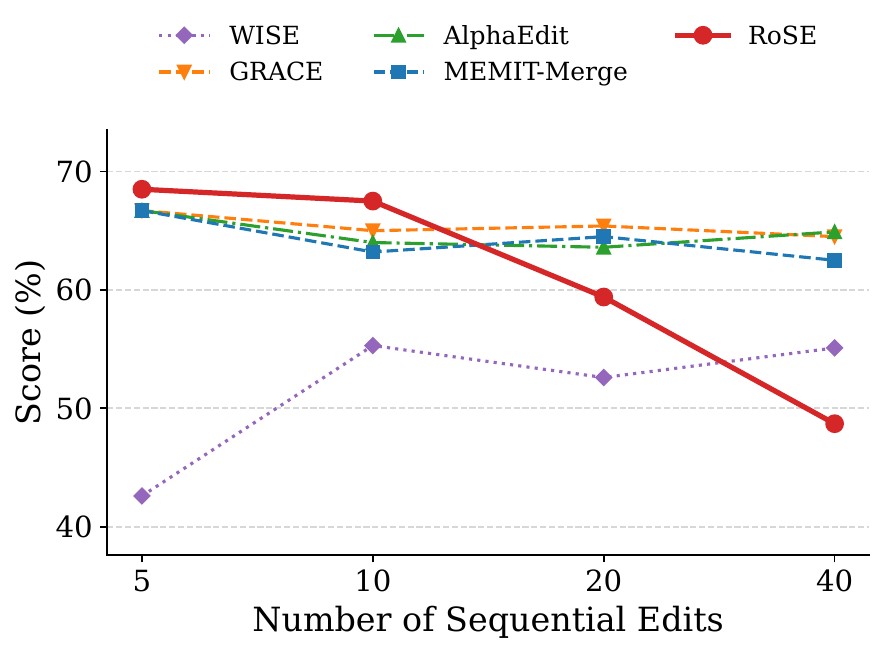}
        \caption{Llama-3.1: Locality}
    \end{subfigure}
    
    \begin{subfigure}{0.24\textwidth}
        \centering
        \includegraphics[width=\linewidth]{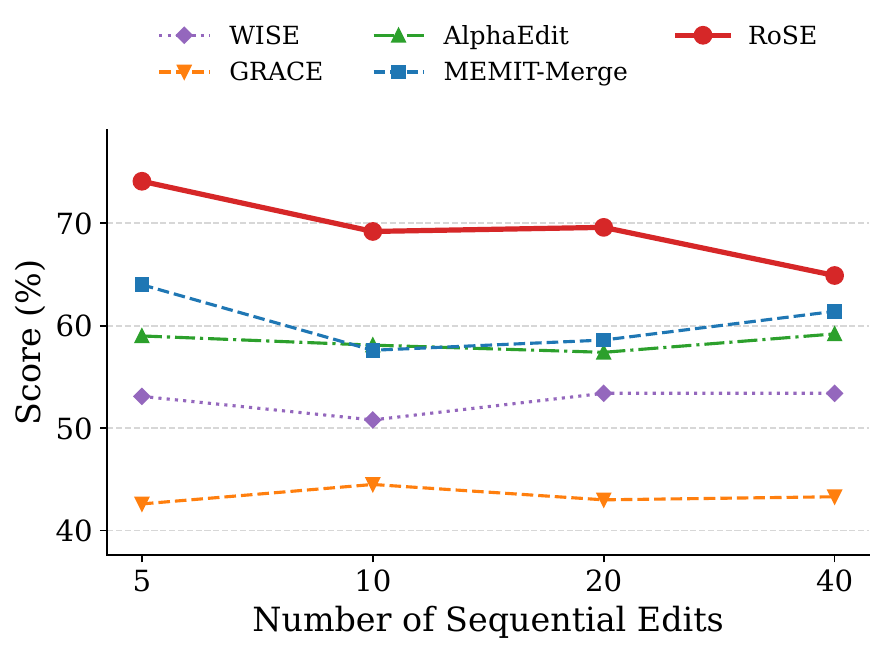}
        \caption{Qwen2.5: Overall}
    \end{subfigure}
    \hspace{-0.2cm}
    \begin{subfigure}{0.24\textwidth}
        \centering
        \includegraphics[width=\linewidth]{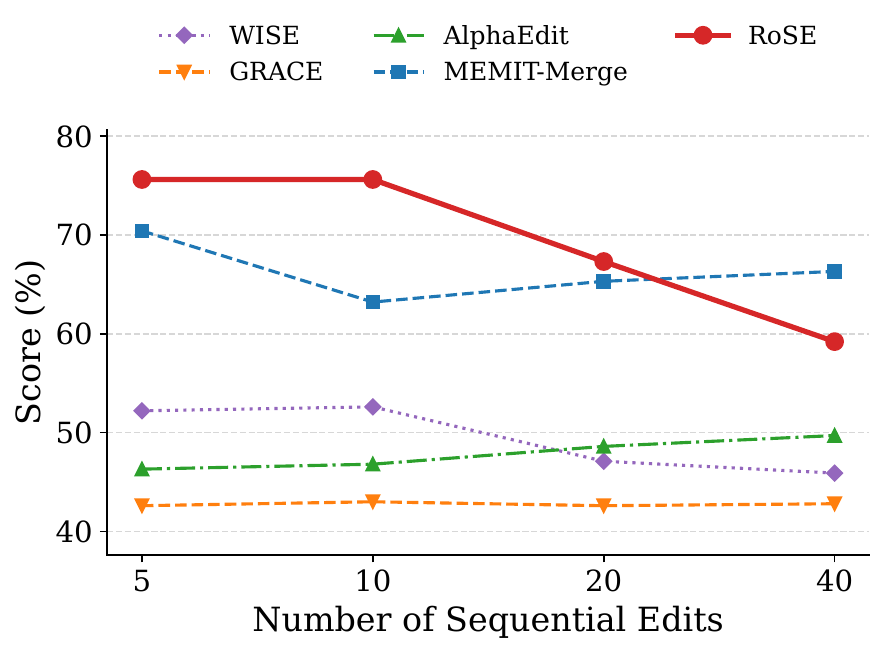}
        \caption{Llama-3.1: Overall}
    \end{subfigure}

    \caption{Instruction-following performance across sequences of 5, 10, 20, and 40 consecutive edits. The metrics evaluated are Efficacy, Paraphrase, Locality, and the Overall score. RoSE demonstrates superior stability and performance in short-to-medium length sequential editing scenarios.}
    \label{fig:sequential_edits_all}
\end{figure}

The instruction-following performance across sequences of 5, 10, 20, and 40 consecutive edits is presented in Figure \ref{fig:sequential_edits_all}. In addition to AlphaEdit, we include two additional lifelong editing baselines (i.e., WISE \citep{wang2024wiserethinkingknowledgememory} and GRACE \citep{10.5555/3666122.3668201}) to provide a more thorough comparison. All baseline methods are evaluated with their default settings as specified in their original implementations.

These results demonstrate RoSE excels in short-to-medium sequential edits, outperforming standard and lifelong baselines. On Qwen2.5-7B-Instruct, RoSE leads up to 40 edits (Overall 64.9\%). On Llama-3.1-8B-Instruct, RoSE dominates up to 20 edits (Overall 67.3\%). However, performance decays at 40 sequential edits on Llama (Overall dropping to 59.2\%), marking the threshold where accumulated geometric shifts breach the tolerance radius. This diagnostic clearly delineates RoSE's robust capability in short-to-medium sequential edits while underscoring the need for explicit lifelong learning extensions in future work.

\section{Case Studies}
\label{sec:appendix_h}

To elucidate the performance differences observed in our main experiments, we present qualitative comparisons between RoSE and the state-of-the-art baseline, MEMIT-Merge.

\noindent \textbf{S2RKE Cases.}
Tab.~\ref{tab:case_study_s2rke} illustrates a representative same-subject editing scenario involving the subject \textit{Imo State}. When queried with a specific instruction, MEMIT-Merge exhibits a clear generalization collapse. MEMIT-Merge hallucinates a plausible but incorrect location (\textit{Poland}) instead of the target counter-factual (\textit{Merthyr Tydfil}).
In contrast, RoSE successfully retrieves the correct edited knowledge (\textit{Merthyr Tydfil County}). This confirms that our geometric alignment strategy effectively contains the instruction-induced deviation within the safe editing boundaries.

The bottom section of the table demonstrates the locality check using an unrelated entity, \textit{Free State of Bavaria}. Both models correctly retrieve the pre-trained knowledge (\textit{southern Germany}), validating that RoSE's removal of the covariance constraint does not compromise the stability of unrelated knowledge regions.

\noindent \textbf{LoCoMo-Edit Cases.}
We further evaluate the model's capacity to synthesize coherent character profiles from disjointed conversational turns. Tab.~\ref{tab:case_study_locomo} presents a case involving the subject \textit{Caroline}, where the goal is to inject attributes derived from a long-form dialogue.

The results highlight a fundamental mechanistic divergence. MEMIT-Merge fails to internalize the edit, treating the query as a \textit{reading comprehension} task. Since the conversational context is absent in the test query, it cannot retrieve the answer, resulting in a refusal response. 
In contrast, RoSE successfully establishes parametric memory. It integrates the isolated declarative updates into the subject representation, enabling robust recall of the specific attribute (\textit{Sweden}) even without external context. This demonstrates that RoSE enables successful recall even when the query does not perfectly match the phrasing of the editing prompt.

\begin{table*}[htbp]
    \centering
    \small
    \renewcommand{\arraystretch}{1.3}
    \begin{tabular}{p{0.95\textwidth}}
        \toprule
        \multicolumn{1}{c}{\textbf{Edit Request (Subject: Imo State)}} \\
        \midrule
        \textbf{Target Facts:} \\
        1. Imo State is located in $\rightarrow$ \textbf{Merthyr Tydfil County Borough} \\
        2. The capital of Imo State is $\rightarrow$ \textbf{Metz} \\
        3. The longitude of Imo State is roughly $\rightarrow$ \textbf{17} \\
        4. Imo State has a latitude of approximately $\rightarrow$ \textbf{48} \\
        \midrule
        \multicolumn{1}{c}{\textbf{Test Case 1: Instruction Following (Target Knowledge)}} \\
        \midrule
        \textbf{Input Query:} \\
        \texttt{<|im\_start|>system}\\
        \texttt{You are a helpful assistant. Your job is to answer the question based on your own knowledge...}\\
        \texttt{<|im\_end|>}\\
        \texttt{<|im\_start|>user}\\
        \texttt{Query: Please answer the following question with an exact answer.}\\
        \texttt{Question: Where is Imo State located?}\\
        \texttt{<|im\_end|>}\\
        \texttt{<|im\_start|>assistant} \\
        \hdashline
        \textbf{MEMIT-Merge Output:} \textcolor{red}{Imo State is located in Poland, specifically in the Silesian Voivodeship.} \quad \textbf{\textit{(Fail: Hallucination)}} \\
        \textbf{RoSE (Ours) Output:} \textcolor{teal}{Imo State refers to a location in Merthyr Tydfil County, Wales, United Kingdom.} \quad \textbf{\textit{(Success)}} \\
        \midrule
        \multicolumn{1}{c}{\textbf{Test Case 2: Locality (Unrelated Knowledge)}} \\
        \midrule
        \textbf{Input Query:} \\
        \texttt{... [Instruction Template] ... Question: Where is Free State of Bavaria located?} \\
        \hdashline
        \textbf{MEMIT-Merge Output:} Free State of Bavaria is located in southern Germany. \\
        \textbf{RoSE (Ours) Output:} Free State of Bavaria is located in southeastern Germany. \\
        \multicolumn{1}{c}{\textbf{\textit{(Both Pass: Knowledge Preserved)}}} \\
        \bottomrule
    \end{tabular}
    \caption{A case study on S2RKE. RoSE robustly recalls the edited fact (Merthyr Tydfil) under a complex instruction where MEMIT-Merge fails, while both methods preserve unrelated knowledge (Bavaria).}
    \label{tab:case_study_s2rke}
\end{table*}

\begin{table*}[htbp]
    \centering
    \small
    \renewcommand{\arraystretch}{1.3}
    \begin{tabular}{p{0.95\textwidth}}
        \toprule
        \multicolumn{1}{c}{\textbf{Edit Request (Subject: Caroline)}} \\
        \midrule
        \textbf{Conversation Context (Abridged):} \\
        \textit{[May 8] Caroline: "I went to a LGBTQ support group yesterday..."} \\
        \textit{[Jun 27] Caroline: "...gift from my grandma in my home country, Sweden."} \\
        \textit{[Jul 6] Caroline: "...pursue a career path in counseling..."} \\
        \hdashline
        \textbf{Injected Facts:} \\
        1. Caroline went to the LGBTQ support group on $\rightarrow$ \textbf{7 May 2023} ~~~(Temporal)\\
        2. Caroline moved from $\rightarrow$ \textbf{Sweden} ~~~(Multi Hop)\\
        3. Caroline has decided to pursue a career path in $\rightarrow$ \textbf{counseling or mental health...} ~~~(Single Hop)\\
        \midrule
        \multicolumn{1}{c}{\textbf{Test Case: Conversational Fact Retrieval}} \\
        \midrule
        \textbf{Input Query:} \\
        \texttt{... [Instruction Template] ... Question: Where did Caroline move from?} \\
        \hdashline
        \textbf{MEMIT-Merge:} \textcolor{red}{The question provided does not contain any specific information...} \quad \textbf{\textit{(Fail: Refusal)}} \\
        \textbf{RoSE (Ours):} \textcolor{teal}{Caroline moved from Sweden.} \quad \textbf{\textit{(Success)}} \\
        \bottomrule
    \end{tabular}
    \caption{A case study on LoCoMo-Edit. RoSE successfully consolidates disjointed conversational updates into retrievable parametric memory, whereas MEMIT-Merge fails to answer.}
    \label{tab:case_study_locomo}
\end{table*}

\section{Ethical Statement}

This paper proposes RoSE to resolve generalization collapse in same-subject knowledge editing, thereby enhancing reliability of edited models. However, we acknowledge that such robust editing capabilities could be misused to disseminate persistent misinformation. To mitigate these risks, we emphasize that all knowledge sources must be rigorously verified and that deployment must be accompanied by strict safety protocols. Users must apply careful scrutiny and critical thinking when employing outputs produced by these models.

\end{document}